\documentclass[11pt]{article}

\usepackage[preprint]{acl}

\usepackage{times}
\usepackage{latexsym}

\usepackage[T1]{fontenc}

\usepackage[utf8]{inputenc}

\usepackage{microtype}

\usepackage{inconsolata}

\usepackage{graphicx}
\usepackage{booktabs}
\usepackage{multirow}
\usepackage{amsmath}
\usepackage{todonotes}
\usepackage{pifont}
\usepackage{subcaption}
\newcommand{\cmark}{\textcolor{green!70!black}{\ding{51}}}
\newcommand{\xmark}{\textcolor{red}{\ding{55}}}
\newcommand{\model}{\texttt{MAGneT}}
\usepackage[most]{tcolorbox}
\usepackage{xcolor}
\usepackage[dvipsnames]{xcolor}
\definecolor{myrow}{RGB}{200, 240, 235}
\usepackage{colortbl}

\title{MAGneT: Coordinated Multi-Agent Generation of Synthetic \\Multi-Turn Mental Health Counseling Sessions}

\author{
 \textbf{Aishik Mandal\textsuperscript{1}},
 \textbf{Tanmoy Chakraborty\textsuperscript{2}},
 \textbf{Iryna Gurevych\textsuperscript{1}}
\\
 \textsuperscript{1}UKP Lab, Department of Computer Science, Technical University of Darmstadt and\\ National Research Center for Applied Cybersecurity ATHENE, Germany\\
 \textsuperscript{2}Indian Institute of Technology Delhi, India
\\
 \small{\href{https://www.informatik.tu-darmstadt.de/ukp/ukp_home/index.en.jsp}{www.ukp.tu-darmstadt.de}
 }
 \\
}

\begin{document}
\maketitle
\begin{abstract}
The growing demand for scalable psychological counseling highlights the need for high-quality, privacy-compliant data, yet such data remains scarce. Here we introduce \model, a novel multi-agent framework for synthetic psychological counseling session generation that decomposes counselor response generation into coordinated sub-tasks handled by specialized LLM agents, each modeling a key psychological technique. Unlike prior single-agent approaches, \model\ better captures the structure and nuance of real counseling. We further propose a unified evaluation framework that consolidates diverse automatic metrics and expands expert assessment from four to nine counseling aspects, thus addressing inconsistencies in prior evaluation protocols. Empirically, \model\ substantially outperforms existing methods: experts prefer \model-generated sessions in 77.2\% of cases on average across the nine aspects over the strongest baseline, and sessions generated by \model\ using Llama3-8B-Instruct backbone yield 3.2\% higher general counseling skills and 4.3\% higher CBT-specific skills on cognitive therapy rating scale (CTRS). An open source Llama3-8B-Instruct model fine-tuned on \model-generated data also outperforms models fine-tuned using baseline synthetic datasets by 6.9\% on average on CTRS. We make our code, data and fine-tuned model public.\footnote{%
\includegraphics[width=0.3cm]{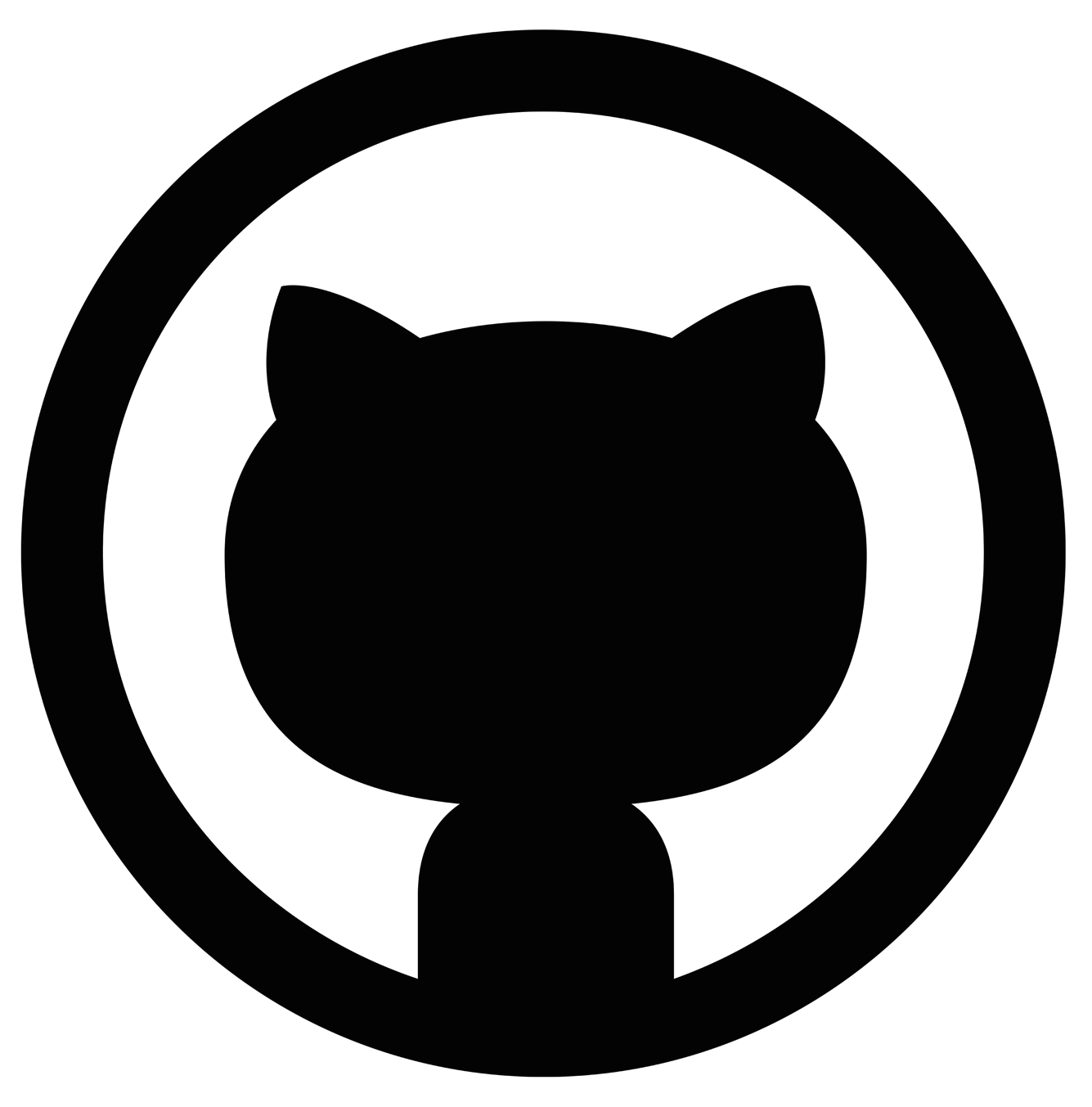}\href{https://github.com/UKPLab/emnlp2026-magnet}{Code}
\quad
\includegraphics[width=0.3cm]{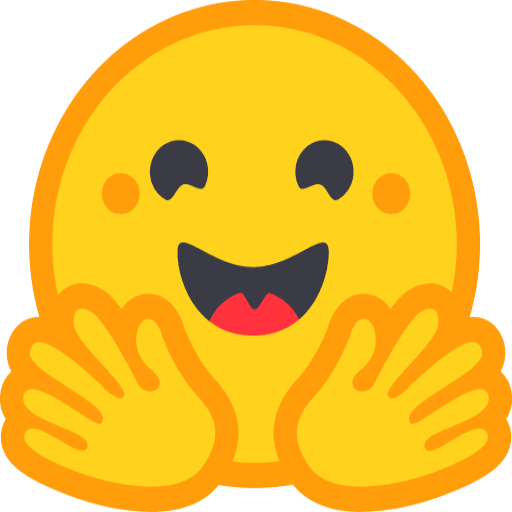}\href{https://huggingface.co/collections/UKPLab/magnet}{Data \& Model}%
}

\end{abstract}

\section{Introduction}
Mental health issues affect nearly 1 in 7 people worldwide~\footnote{WHO (2025): \href{https://www.who.int/news-room/fact-sheets/detail/mental-disorders}{https://www.who.int/news-room/fact-sheets/detail/mental-disorders}}, highlighting the need for scalable counseling solutions. However, demand for psychological support far exceeds the availability of trained professionals, leaving many without adequate care~\cite{kazdin2021extending}. Recently, there has been a growing interest in using Large Language Models (LLMs) for mental health counseling (hereafter, referred to as counseling). While closed-source models show promising conversational and psychological capabilities \cite{Raile2024_chatgpt_counseling,chatgpt_gpt4_counseling}, their practical use is limited by privacy concerns and weaker performance on counseling-specific tasks \citep{cactus,bolt,llm_review_1}. Open-source LLMs fine-tuned on counseling data offer an alternative, but such data is scarce due to privacy constraints. Solutions such as manual de-identification or automatic pseudonymization \cite{pseudonymize_1,psuedonymize_2} remain limited in scalability and robustness.

\begin{table*}[ht]
\centering
\resizebox{0.8\textwidth}{!}{%
\begin{tabular}{@{}lccccccccc@{}}
\toprule
\multirow{2}{*}{Method} &  & \multirow{2}{*}{CBT} & \multirow{2}{*}{Multi-Agent} & & \multicolumn{5}{c}{Evaluation Framework}  \\  \cmidrule(l){6-10} 
&  &      &     &  & Diversity & CTRS & WAI & PANAS & Expert Evaluation \\ \midrule
SMILE \cite{smilechat}  &  & \xmark &  \xmark & & \cmark & \xmark   &  \xmark  &   \xmark    &  $1$ Aspect \\

Psych8k \cite{chatcounselor} &  & \xmark & \xmark & & \xmark &   \xmark & \xmark  & \xmark & \xmark          \\

CPsyCoun \cite{CPsycoun} &  & \xmark & \xmark & & \cmark  & \xmark   &   \xmark  &   \xmark    &  \xmark \\

\citet{llm-roleplay} &  & \xmark    & \xmark & & \cmark & \xmark   &   \cmark  &   \xmark    & $1$ Aspect \\

CACTUS \cite{cactus} &  & \cmark    & \xmark & & \cmark &   \cmark   &   \xmark  &   \cmark    & $4$ Aspects \\

\rowcolor{myrow} \model &  & \cmark    & \cmark  &  & \cmark &   \cmark   &   \cmark  &  \cmark  &  $9$ Aspects \\ \bottomrule
\end{tabular}%
}
\caption{\model\ and our unified evaluation compared to prior works in synthetic counseling session generation.}
\label{tab:novelty}
\end{table*}

Synthetic data generation offers a scalable, privacy-preserving solution for fine-tuning LLMs for counseling. Early work focused on single-turn therapeutic responses \citep{cognitive-reframe, psyqa, chatcounselor}, and subsequent works extended to multi-turn interactions using single-turn Q\&A datasets \citep{soulchat, smilechat} or role-playing LLMs \citep{llm-roleplay, DEDURO2025100170}. However, these systems lack grounding in established psychology theory. To address this gap, some prior works leverage structured psychological inputs~\citep{CPsycoun,sqpconv,graph2counsel}, while CACTUS \citep{cactus} incorporates a Cognitive Behavioral Therapy (CBT) based planning agent. Yet, these approaches rely on a single agent to generate the counselor's response, which is insufficient for modeling complex therapeutic strategies such as reflection, questioning, solution provision, normalization, and psycho-education \cite{bolt}. While multi-agent systems have been explored for counseling tasks, existing works focus on generating single-turn supportive responses \citep{MADP}, or structured diagnostic interviews \citep{trustworthy}, failing to capture the open-ended, dynamic nature of counseling sessions, or address session-level planning without turn-level psychological grounding~\citep{chen-etal-2025-catch}. 

\begin{figure*}[t]
  \centering
  \includegraphics[width=0.85\linewidth]{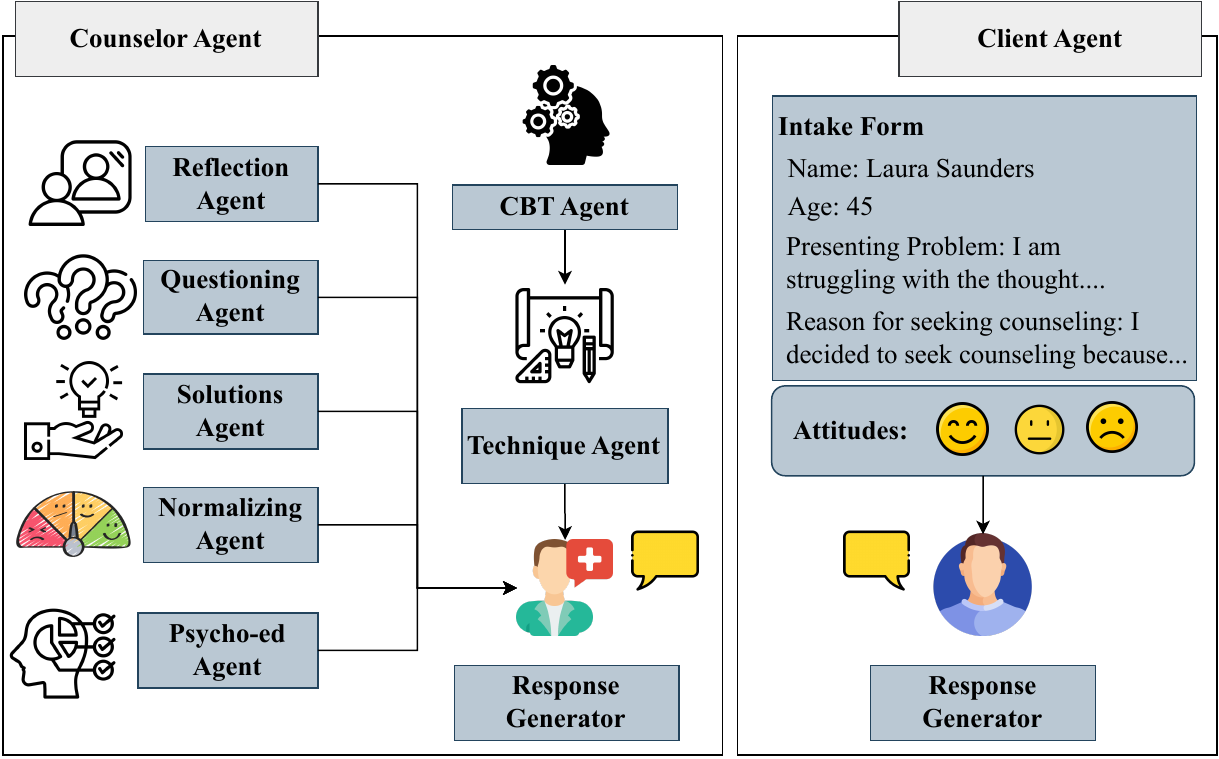}
  \caption{\textbf{Overview of \model.} Counselor response is generated using specialized response agents (reflection, questioning, solution, normalizing, psycho-education), a technique agent, a CBT agent, and a response generation agent.}
  \label{fig:mascs-gen}
\end{figure*}

To address these limitations, we introduce \model, a multi-agent framework that decomposes counselor response generation into coordinated sub-tasks. The framework includes five specialized response agents, each aligned with core therapeutic strategies described in prior literature \citep{bolt, lee-etal-2019-identifying, cao-etal-2019}: reflection, questioning, solution provision, normalization, and psycho-education. Their outputs are integrated by a final response generation agent responsible for producing a coherent, contextually appropriate counselor utterance. The final response generation agent is guided by two controllers: a turn-level technique selector agent and a session-level CBT-based planning agent. On the client side, we simulate realistic behavior via detailed profiles and attitude modeling. \model\ enables scalable generation of multi-turn, psychologically grounded synthetic counseling sessions via client-counselor role-play. A schematic diagram of \model\ is provided in Figure~\ref{fig:mascs-gen}.

Another persistent challenge in this domain is the lack of standardized evaluation. Prior works use inconsistent evaluation metrics -- CACTUS \cite{cactus} uses Positive and Negative Affect Schedule (PANAS) \cite{panas} and Cognitive Therapy Rating Scale (CTRS) \cite{ctrs,young1980ctrs}, while \citet{llm-roleplay} use Working Alliance Inventory (WAI) \cite{wai-psych}, thus hindering direct comparison among methods. A similar pattern is observed in expert evaluation as well. We address this by introducing a unified framework that consolidates these metrics and expands expert assessment from four to nine counseling aspects, enabling a more rigorous evaluation. Our evaluation shows that experts prefer \model-generated sessions in $77.2$\% of cases on average across the nine aspects compared to the strongest baseline generations (Figure~\ref{fig:heval}). On CTRS, \model-generated sessions outperform state-of-the-art baseline generations \cite{cactus,chatcounselor} by $3.2$\% on general counseling and $4.3$\% on CBT-specific skills on average. Additionally, fine-tuning Llama3-8B-Instruct \cite{meta2024introducing} on \model-generated data yields a $6.9$\% average improvement in CTRS compared to fine-tuning on existing synthetic datasets. Overall, \model\ is consistently preferred across evaluations, showing its effectiveness for synthetic counseling session generation. Table \ref{tab:novelty} summarizes the novelty of our work relative to prior work.

In summary, our contributions are: (i) \model, a novel psychologically grounded multi-agent synthetic counseling session generation framework, (ii) An open-source model fine-tuned on data generated by \model, achieving strong gains in counseling metrics and (iii)  unified evaluation framework that integrates evaluations from prior work and expands expert assessment to nine aspects of counseling.

\section{Related Work}

\paragraph{Synthetic Counseling Data Generation.} Due to privacy constraints limiting access to real counseling data, there is growing interest in synthetic counseling dialogue generation. Early works like Psych8k \cite{chatcounselor} generate counselor responses to client questions but are restricted to single-turn interactions. To address the need for multi-turn conversations, subsequent works such as SMILE \cite{smilechat} and SoulChat \cite{soulchat} extend the generation to multi-turn dialogues by using psychological Q\&A data from online forums. However, such data lack clinical validation and psychological grounding. Other works simulate counseling via two LLMs in a role-play setup (one acting as the client and the other as the counselor) \cite{llm-roleplay,DEDURO2025100170} but these approaches also lack grounding in psychological theory. To address this, prior work incorporates structured inputs: CPsyCoun \cite{CPsycoun} uses counseling memos, SQPsychConv~\cite{sqpconv} leverages questionnaires, and Graph2Counsel~\cite{graph2counsel} uses Client Psychological Graphs. On the other hand, CACTUS \cite{cactus} incorporates a CBT plan generated by a dedicated CBT planning agent. However, all these works rely on a single LLM for counselor response generation, limiting their ability to capture diverse therapeutic techniques--such as reflection, questioning, solution provision, normalization, and psycho-education, observed in real counseling \cite{bolt}. In contrast, \model\ decomposes response generation across specialized agents, each aligned with a therapeutic technique, coordinated by a technique selector and a CBT planning agent, followed by a response generation agent that generates the final counselor response, thereby breaking down the generation process into manageable sub-tasks.

\paragraph{Multi-Agent Framework.} LLMs often struggle to execute complex tasks in isolation. Multi-agent frameworks address this by decomposing them into simpler sub-tasks handled by specialized agents \cite{meta-gpt,chatdev,autoact}, achieving strong results in domains like recommender systems \cite{multi-agent-reco} and task-oriented dialogue systems \cite{multi-agent-task-oriented}. Counselor response generation is similarly complex, requiring both understanding client concerns and applying therapeutic techniques (e.g., reflection, questioning, normalization) \cite{bolt}. However, existing multi-agent approaches in psychological domains remain limited. MADP \citep{MADP} focuses solely on generating single-turn supportive responses and cannot model multi-turn interactions, which progresses through phases such as rapport building, exploration, and goal setting. Similarly, \citet{trustworthy} focus on questionnaire-driven diagnostic dialogues, lacking open-ended structure. In contrast, CATCH~\cite{chen-etal-2025-catch} targets multi-session therapy generation without improving per-turn psychological grounding of counselor responses. To address these gaps, \model\ introduces a multi-agent framework for generating multi-turn synthetic counseling sessions. A CBT agent produces a session plan, and a technique agent selects turn-level therapeutic strategies, enabling dynamic, structured interactions that more faithfully capture real counseling processes.

\section{Our Proposed Model}

In this section, we present \model, a novel multi-agent framework for synthetic counseling session generation that explicitly models the psychologically grounded reasoning processes underlying counseling. Prior role-play based approaches simulate counselor responses using a single LLM \cite{llm-roleplay}, which struggles to correctly select and apply CBT techniques \citep{cactus}. While CACTUS addresses it through a CBT planning agent, counselor responses are still generated by a single agent, limiting fine-grained modeling of therapeutic techniques \citep{bolt}.

\model\ addresses these limitations by decomposing counselor response generation into modular, specialized agents, each responsible for a distinct counseling function, enabling finer control and closer alignment with established therapeutic practice. Similar to prior LLM-based simulations, \model\ adopts a two-party role-play setup where LLMs simulate both counselor and client. This approach enables scalable synthetic data generation while avoiding reliance on real counseling records, thereby preserving privacy. Unlike prior work, \model\ employs a multi-agent counselor simulator that explicitly models key psychological techniques: reflection, questioning, solution provision, normalization, and psycho-education, identified from clinical literature \cite{bolt,lee-etal-2019-identifying,cao-etal-2019} to avoid redundancy while preserving therapeutic diversity. To promote realistic and diverse interactions, each session is initialized with detailed client intake forms \cite{cactus}, including client background, client issues, and reasons for seeking therapy. Figure~\ref{fig:mascs-gen} presents a schematic diagram of \model.

\subsection{Multi-Agent Counselor Simulation}

Effective counseling responses are therapeutically nuanced, requiring both a structured treatment plan and dynamic use of psychological techniques. To mirror this complexity, \model\ simulates the counselor using a coordinated ensemble of LLM agents -- (i) a CBT agent to produce a structured treatment plan, (ii) five specialized response agents, each focusing on a specific psychological technique, (iii) a technique agent to determine the appropriate combination of techniques for a given turn, and (iv) a response generation agent to synthesize the final counselor response. Now we will describe each agent in detail (see Appendix \ref{sec:appendix-counselor} for the details of the prompt used for each agent).

\begin{table*}[t]
\centering
\resizebox{1.9\columnwidth}{!}{%
\begin{tabular}{@{}lccccccc@{}}
\toprule
Method    & Distinct-1 ($\uparrow$) & 1-Grams & Distinct-2 ($\uparrow$) & 2-Grams & Distinct-3 ($\uparrow$) & 3-Grams & EAD ($\uparrow$)    \\ \midrule
Psych8k  & $0.0044$     & $1588861$ & $0.0570$     & $1588411$ & $0.1604$     & $1587961$ & $0.0546$ \\
CACTUS    & $0.0048$     & $1426490$ & $0.0619$     & $1426045$ & $0.1733$     & $1425600$ & $0.0537$ \\ \midrule
\rowcolor{myrow} \model & $\textbf{0.0050}$     & $1431874$ & $\textbf{0.0685}$     & $1431432$ & $\textbf{0.2009}$     & $1430990$ & $\textbf{0.0562}$ \\ \bottomrule
\end{tabular}%
}
\caption{Data diversity of generated counseling dialogues across methods. The table reports n-grams for $n \in \{1,2,3\}$, Distinct-n for $n \in \{1,2,3\}$ and Expectation-Adjusted Distinct (EAD).}
\label{tab:diversity}
\end{table*}

\paragraph{CBT Agent.} Cognitive theory suggests that maladaptive interpretations of events contribute to mental health issues \cite{Powles1974BeckAT}. CBT-based counseling seeks to identify and reframe these thought patterns \cite{greimel2011cognitive}. CBT-based tools have already shown their effectiveness for conditions such as depression and anxiety \cite{cbt1,cbt2,cbt3}. To integrate CBT into our response-generation framework, we introduce a CBT agent that produces a session-level plan tailored to the client’s cognitive patterns and presenting issues. This plan specifies behavioral goals and cognitive reframing strategies, offering high-level guidance for subsequent counselor actions. It is generated using the client’s intake form and first utterance.

\paragraph{Specialized Response Agents.} Counselors use different types of psychological techniques to explore client issues, understand their perspective, and provide solutions. \citet{bolt} identifies such techniques \cite{lee-etal-2019-identifying,cao-etal-2019} commonly used in high-quality counseling sessions, grouping them into five core categories: reflection, questioning, solution provision, normalization, and psycho-education. To model these counselor functions, \model\ employs five specialized response agents, each aligned with one of these core techniques. The reflection agent aims to help the client gain insight by mirroring or paraphrasing their expressions, thus encouraging self-evaluation. The questioning agent aims to gain a deeper understanding of the client's feelings and reactions to alternate perspectives. The normalizing agent acknowledges and validates the client's experiences as typical and understandable, fostering empathy and safety. Along with understanding the client's issues, perspectives, and acknowledging normalcy, the counselor also needs to provide possible solutions to the client to deal with their conditions. The solution agent provides such actionable solutions to alleviate the client's psychological distress. The counselor also needs to convince the client and get them on board regarding their diagnosis and solution strategies. For this, the psycho-education agent provides therapeutically relevant information to clients to build an understanding of their issues and treatment plan. Each of these agents generates a candidate response based on the current dialogue history and client information.

\paragraph{Technique Agent.} Effective therapeutic communication often involves blending multiple techniques \cite{bolt}. The technique agent dynamically selects an appropriate subset of techniques to be employed in the current turn, guided by the CBT plan and dialogue context. This ensures that counselor behavior remains consistent with both therapeutic intent and session flow.

\paragraph{Response Generation Agent.} The response generation agent produces the final counselor utterance by fusing candidate responses from the specialized response agents following the technique agent’s strategy. This decoupled design preserves coherence while flexibly adapting to client needs.

\subsection{Client Simulation}

The client agent complements the multi-agent counselor by generating realistic, varied client responses. Each client is initialized using a structured intake form \cite{cactus}, including their background, issues, and therapy goals. To enhance interaction diversity, we use three client attitudes: positive, neutral, and negative \cite{cactus}, each guided by detailed instructions that govern tone, openness, and emotional intensity. The client agent conditions its responses on the intake form, dialogue history, specified attitude, and associated behavioral instructions. This setup enables \model\ to simulate a wide range of client behaviors, improving the diversity and realism of the generated sessions. The details of the prompt, attitude instructions, and the intake form are provided in Appendix \ref{sec:appendix-client}.

\begin{table*}[th]
\centering
\resizebox{1.95\columnwidth}{!}{%
\begin{tabular}{@{}lcccccccccccccccccccc@{}}
\toprule
\multirow{3}{*}{Method} 
& \multicolumn{7}{c}{CTRS} & & \multicolumn{8}{c}{PANAS} & & \multicolumn{3}{c}{WAI} \\ 
\cmidrule(lr){2-8} \cmidrule(l){10-17} \cmidrule(l){19-21}
& \multicolumn{3}{c}{General} & & \multicolumn{3}{c}{CBT} & & \multicolumn{2}{c}{Pos. Att.} & & \multicolumn{2}{c}{Neu. Att.} & & \multicolumn{2}{c}{Neg. Att.} & & \multirow{2}{*}{Task ($\uparrow$)} & \multirow{2}{*}{Goal ($\uparrow$)} & \multirow{2}{*}{Bond ($\uparrow$)} \\ 
\cmidrule(lr){2-4} \cmidrule(lr){6-8} \cmidrule(lr){10-11} \cmidrule(lr){13-14} \cmidrule(lr){16-17}
& U ($\uparrow$) & I ($\uparrow$) & C ($\uparrow$) & & D ($\uparrow$) & F ($\uparrow$) & S ($\uparrow$) & & P ($\uparrow$) & N ($\downarrow$) & & P ($\uparrow$) & N ($\downarrow$) & & P ($\uparrow$) & N ($\downarrow$) & & \\ 
\midrule
Psych8k & $\underline{3.90}^{*}$ & $\underline{4.10}^{*}$ & $\underline{3.13}^{*}$ & & $\underline{3.80}^{*}$ & $3.35^{*}$ & $2.59^{*}$ & & $\underline{1.12}$ & $\underline{\textbf{-0.78}}$ & & $\underline{0.80}$ & $-0.43$ & & $\underline{\textbf{-0.01}}^{*}$ & $\underline{\textbf{-0.30}}$ & & $\underline{4.86}$ & $\underline{4.73}$ & $\underline{4.93}^{*}$ \\
CACTUS & $3.84^{*}$ & $3.94^{*}$ & $3.09^{*}$ & & $3.74^{*}$ & $\underline{3.37}^{*}$ & $\underline{2.83}$ & & $0.86^{*}$ & $-0.64$ & & $0.64^{*}$ & $\underline{\textbf{-0.45}}$ & & $-0.10$ & $\underline{\textbf{-0.30}}$ & & $4.69^{*}$ & $4.39^{*}$ & $4.80^{*}$ \\
\rowcolor{myrow} \model & $\textbf{3.98}$ & $\textbf{4.30}$ & $\textbf{3.43}$ & & $\textbf{4.08}$ & $\textbf{3.76}$ & $\textbf{2.93}$ & & $\textbf{1.13}$ & $-0.73$ & & $\textbf{0.82}$ & $-0.41$ & & $-0.15$ & $-0.25$ & & $\textbf{4.94}$ & $\textbf{4.78}$ & $\textbf{5.01}$ \\ 
\midrule
$\delta(\%)$ & $\textcolor{ForestGreen}{\textbf{+1.3}}$ & $\textcolor{ForestGreen}{\textbf{+3.3}}$ & $\textcolor{ForestGreen}{\textbf{+5.0}}$ & & $\textcolor{ForestGreen}{\textbf{+4.7}}$ & $\textcolor{ForestGreen}{\textbf{+6.5}}$ & $\textcolor{ForestGreen}{\textbf{+1.7}}$ & & $\textcolor{ForestGreen}{\textbf{+0.3}}$ & $\textcolor{red}{\textbf{-1.3}}$ & & $\textcolor{ForestGreen}{\textbf{+0.5}}$ & $\textcolor{red}{\textbf{-1.0}}$ & & $\textcolor{red}{\textbf{-3.5}}$ & $\textcolor{red}{\textbf{-1.3}}$ & & $\textcolor{ForestGreen}{\textbf{+1.3}}$ & $\textcolor{ForestGreen}{\textbf{+0.8}}$ & $\textcolor{ForestGreen}{\textbf{+1.3}}$ \\ 
\bottomrule
\end{tabular}%
}
\caption{Evaluation of generated counseling sessions across CTRS, PANAS, and WAI dimensions. Asterisks (*) indicates significant differences from \model\ ($p < 0.05$, paired t-test). For CTRS: U (Understanding), I (Interpersonal Effectiveness), C (Collaboration), D (Guided Discovery), F (Focus), S (Strategy). For PANAS: Pos. Att. (Positive Attitude), Neu. Att. (Neutral Attitude), Neg. Att. (Negative Attitude), P (average shift in positive emotions), N (average shift in negative emotions). $\delta(\%)$ shows \model’s \%-age margin over the best baseline (\underline{underlined}).}
\label{tab:ctrs-panas-wai}
\end{table*}

\begin{table*}[ht]
\centering
\resizebox{1.25\columnwidth}{!}{%
\begin{tabular}{lcccc}
\toprule
\textbf{Dimension} & \textbf{Psych8k} & \textbf{\model} & \textbf{Tie} & \textbf{\model{} win\%} \\
\midrule
Comprehensiveness & 4 & 29 & 1 & 85.3\% \\
Professionalism    & 3 & 26 & 5 & 76.5\% \\
Authenticity       & 2 & 30 & 2 & 88.2\% \\
Safety             & 2 & 27 & 5 & 79.4\% \\
Content            & 3 & 30 & 1 & 88.2\% \\
Directiveness      & 3 & 28 & 3 & 82.4\% \\
Exploratoriness    & 2 & 31 & 1 & 91.2\% \\
Supportiveness     & 3 & 27 & 4 & 79.4\% \\
Expressiveness     & 3 & 29 & 2 & 85.3\% \\
\bottomrule
\end{tabular}%
}
\caption{Expert preference between \model-generated and Psych8k-generated counseling sessions for clients presenting a negative attitude, across nine evaluation dimensions. Columns report the number of sessions preferred for each source, ties, and the corresponding win rate for \model\ (computed as \model\ wins over all comparisons). Experts favor \model\ across all dimensions.}
\label{tab:negative-attitude-expert}
\end{table*}

\begin{table*}[th]
\centering
\resizebox{1.95\columnwidth}{!}{%
\begin{tabular}{@{}lcccccccccccccccccccc@{}}
\toprule
\multirow{3}{*}{Model} 
& \multicolumn{7}{c}{CTRS} 
& & \multicolumn{8}{c}{PANAS} 
& & \multicolumn{3}{c}{WAI} \\ 
\cmidrule(lr){2-8} \cmidrule(l){10-17} \cmidrule(l){19-21}

& \multicolumn{3}{c}{General} 
& & \multicolumn{3}{c}{CBT} 
& & \multicolumn{2}{c}{Pos. Att.} 
& & \multicolumn{2}{c}{Neu. Att.} 
& & \multicolumn{2}{c}{Neg. Att.} 
& & \multirow{2}{*}{Task ($\uparrow$)} & \multirow{2}{*}{Goal ($\uparrow$)} & \multirow{2}{*}{Bond ($\uparrow$)} \\ 

\cmidrule(lr){2-4} \cmidrule(lr){6-8} \cmidrule(lr){10-11} 
\cmidrule(lr){13-14} \cmidrule(lr){16-17}

& U ($\uparrow$) & I ($\uparrow$) & C ($\uparrow$) 
& & D ($\uparrow$) & F ($\uparrow$) & S ($\uparrow$) 
& & P ($\uparrow$) & N ($\downarrow$) 
& & P ($\uparrow$) & N ($\downarrow$) 
& & P ($\uparrow$) & N ($\downarrow$) 
& & \\ 

\midrule
L-P 
& $\underline{3.71}^{*}$ & $\underline{3.83}^{*}$ & $\underline{2.91}^{*}$ 
& & $\underline{3.65}^{*}$ & $\underline{3.16}^{*}$ & $2.44^{*}$ 
& & $\underline{1.04}$ & $\underline{-0.88}^{*}$ 
& & $\underline{0.70}$ & $\underline{-0.41}^{*}$ 
& & $\underline{\textbf{-0.03}}^{*}$ & $\underline{\textbf{-0.34}}$ 
& & $\underline{4.72}$ & $\underline{4.44}^{*}$ & $\underline{4.79}$ \\

L-C 
& $3.48^{*}$ & $3.67^{*}$ & $2.65^{*}$ 
& & $3.37^{*}$ & $2.99^{*}$ & $\underline{2.45}^{*}$ 
& & $0.91^{*}$ & $-0.79^{*}$ 
& & $0.59^{*}$ & $-0.39^{*}$ 
& & $-0.13^{*}$ & $-0.32$ 
& & $4.59^{*}$ & $4.18^{*}$ & $4.65^{*}$ \\

\rowcolor{myrow} L-M 
& $\textbf{3.95}$ & $\textbf{4.32}$ & $\textbf{3.32}$ 
& & $\textbf{4.03}$ & $\textbf{3.60}$ & $\textbf{2.96}$ 
& & $\textbf{1.11}$ & $\textbf{-1.10}$ 
& & $\textbf{0.78}$ & $\textbf{-0.59}$ 
& & $-0.31$ & $-0.28$ 
& & $\textbf{4.86}$ & $\textbf{4.78}$ & $\textbf{4.92}$ \\

\midrule
$\delta(\%)$ 
& $\textcolor{ForestGreen}{\textbf{+4.0}}$ & $\textcolor{ForestGreen}{\textbf{+8.2}}$ & $\textcolor{ForestGreen}{\textbf{+6.8}}$ 
& & $\textcolor{ForestGreen}{\textbf{+6.3}}$ & $\textcolor{ForestGreen}{\textbf{+7.3}}$ & $\textcolor{ForestGreen}{\textbf{+8.5}}$ 
& & $\textcolor{ForestGreen}{\textbf{+1.8}}$ & $\textcolor{ForestGreen}{\textbf{+5.5}}$ 
& & $\textcolor{ForestGreen}{\textbf{+2.0}}$ & $\textcolor{ForestGreen}{\textbf{+4.5}}$ 
& & $\textcolor{red}{\textbf{-7.0}}$ & $\textcolor{red}{\textbf{-1.5}}$ 
& & $\textcolor{ForestGreen}{\textbf{+2.3}}$ & $\textcolor{ForestGreen}{\textbf{+5.7}}$ & $\textcolor{ForestGreen}{\textbf{+2.2}}$ \\ 

\bottomrule
\end{tabular}%
}
\caption{Evaluation of Llama3-8B-Instruct models fine-tuned on counseling sessions generated using Psych8k (L-P), CACTUS (L-C), and \model\ (L-M). L-P, L-C, and L-M denote Llama-Psych8k, Llama-CACTUS, and Llama-\model\ respectively. Asterisks (*) indicates significant differences from Llama-\model\ ($p < 0.05$, paired t-test). For CTRS: U (Understanding), I (Interpersonal Effectiveness), C (Collaboration), D (Guided Discovery), F (Focus), S (Strategy). For PANAS: Pos. Att. (Positive Attitude), Neu. Att. (Neutral Attitude), Neg. Att. (Negative Attitude), P (average shift in positive emotions), N (average shift in negative emotions). $\delta(\%)$ shows Llama-\model's percentage margin over the best baseline (\underline{underlined}).}
\label{tab:ctrs-panas-wai-model}
\end{table*}

\section{Unified Evaluation Framework}

Here, we introduce our unified evaluation framework for determining the quality and diversity of the generated synthetic counseling data. Prior work \cite{cactus,smilechat,llm-roleplay,soulchat,CPsycoun} lacks standardized evaluation protocols, making it difficult to compare generation methods or understand the practical effectiveness of the generated synthetic data. Our framework addresses this by consolidating automatic and expert evaluations from prior works, as well as expanding expert evaluation.

\paragraph{Diversity Evaluation.} Data diversity is critical to fine-tuning robust and generalizable counselor models. To assess the diversity of generated counseling sessions, we compute Distinct-$n$ scores \cite{distinct_n} for $n \in \{1, 2, 3\}$. However, Distinct-$n$ is known to penalize longer sequences. To mitigate this, we incorporate Expectation-Adjusted Distinct (EAD) \cite{liu-etal-2022-rethinking}, which adjusts for sequence length effects and provides a more reliable diversity measure. More details about the diversity evaluation are presented in Appendix \ref{sec:appendix-div}.

\paragraph{Quality Evaluation.} The quality of the generated data is also crucial. It is generally measured using psychological scales. CACTUS \cite{cactus} uses CTRS \cite{ctrs,young1980ctrs} and PANAS \cite{panas} to measure the quality, while \citet{llm-roleplay} uses WAI \cite{wai-psych}. As these measures capture complementary aspects of counseling, we adopt all three to form a multi-faceted quality assessment.

CTRS assesses general (Understanding, Interpersonal Effectiveness, Collaboration) and CBT-specific (Guided Discovery, Focus, Strategy) counseling skills on a scale of $0$ to $6$, with higher scores indicating stronger counseling competencies. WAI measures the client-counselor alliance using $12$ items rated on a $1$ to $7$ scale \cite{wai-comp}, grouped into three categories: agreement on Goal, agreement on Task, and Bond. Higher scores reflect a stronger alliance. PANAS evaluates emotional shifts in the client using $20$ emotion items (10 positive, 10 negative), each rated from $1$ to $5$. Effective sessions should increase positive emotions and decrease negative emotions. We use a LLM-as-a-judge setup to score all metrics. More details on the quality evaluation are provided in Appendix \ref{sec:appendix-quality}. We exclude automatic metrics such as BLEU \cite{bleu}, BERTScore \cite{bertscore}, and ROUGE \cite{lin-2004-rouge} due to their reliance on ground-truth references.

\paragraph{Counseling Agent Fine-tuning.} To complement quality evaluation, we assess the downstream utility of the generated synthetic datasets by fine-tuning an open-source Llama3-8B-Instruct model \cite{meta2024introducing} on them and evaluating it using CTRS, WAI, and PANAS. This directly tests how well different synthetic datasets translate into fine-tuning practical counseling agents. The fine-tuning details are provided in Appendix \ref{sec:appendix-counselor-agent-fine-tuning}. We further benchmark the fine-tuned models on CounselingBench \citep{counselingbench}, with details provided in Appendix \ref{sec:appendix-counselingbench}.

\begin{figure*}[t]
  \centering
  \includegraphics[width=1.95\columnwidth]{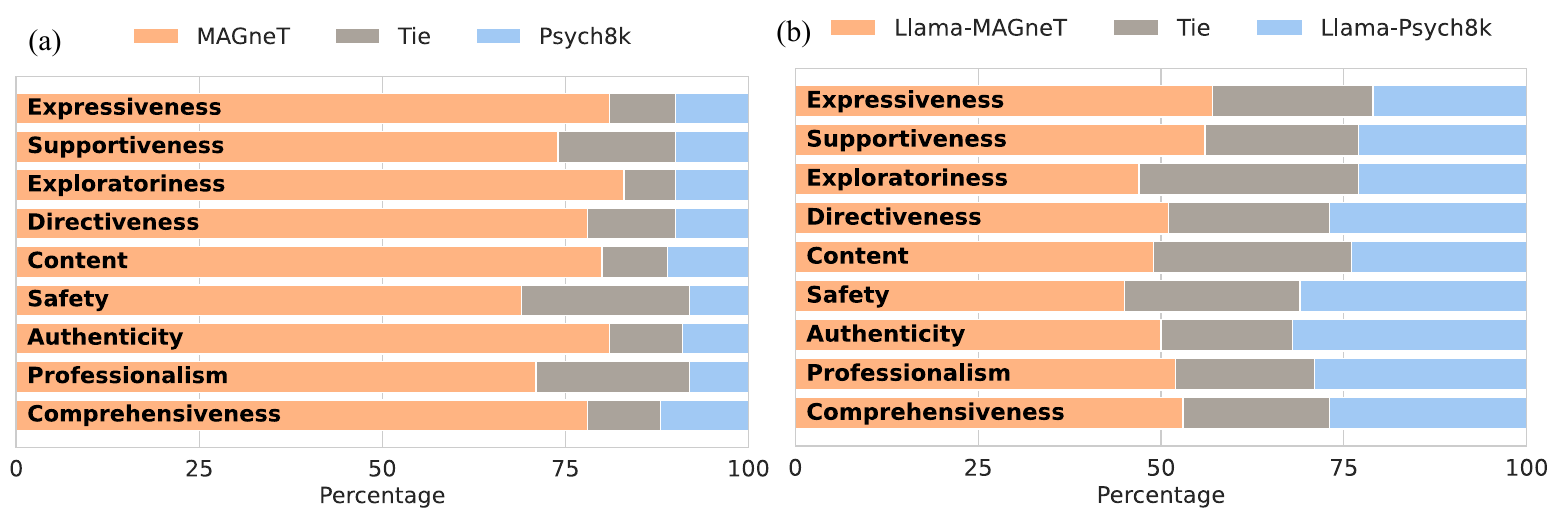}
  \caption{Results of head-to-head comparison of sessions generated by (a) \model\ vs Psych8k (b) Llama-\model\ vs Llama-Psych8k based on expert judgment across nine different aspects of counseling.}
  \label{fig:heval}
\end{figure*}

\paragraph{Expert Evaluation.} Expert evaluation is crucial for capturing qualitative aspects beyond automated metrics. Prior works vary in focus: SoulChat \cite{soulchat} evaluates content naturalness, empathy, helpfulness, and safety; CACTUS \cite{cactus} uses helpfulness, empathy, coherence, and guidance; while CPsyCoun \cite{CPsycoun} automates assessment of comprehensiveness, professionalism, authenticity, and safety. Yet several clinically relevant behaviors remain unexplored, motivating us to introduce a unified and expanded evaluation protocol. In addition to comprehensiveness, professionalism, authenticity, safety, and content naturalness, we assess four additional dimensions -- directiveness, exploratoriness, supportiveness, and expressiveness \cite{mccullough1988psychotherapy}. While prior studies typically rely on two to five raters \citep{smilechat,soulchat,MADP,na-2024-cbt,xiao-etal-2024-healme,mentalchat16k}, we recruit seven expert psychologists\footnote{Evaluators are Rehabilitation Council of India (RCI)-licensed clinical psychologists with extensive experience, specializing in psychotherapy.} to improve reliability and reduce individual bias. The experts conduct blind, pairwise comparisons of 50 sessions generated by \model\ and the best baseline\footnote{Due to the high cost of expert evaluation, we compare only \model\ and the strongest baseline.}, using matched intake forms and client attitudes. Each session pair is independently compared by two evaluators across the nine aspects. We apply the same protocol to compare an additional 50 sessions generated by models fine-tuned on \model-generated data and the best baseline data. Full evaluation guidelines and details on inter-annotator agreement are provided in Appendix \ref{sec:appendix-human-eval}.

\begin{table*}[th]
\centering
\resizebox{2.0\columnwidth}{!}{%
\begin{tabular}{@{}lcccccccccccccccccccc@{}}
\toprule
\multirow{3}{*}{Method} 
    & \multicolumn{7}{c}{CTRS} 
    &  & \multicolumn{8}{c}{PANAS} 
    &  & \multicolumn{3}{c}{WAI} \\ 
\cmidrule(lr){2-8} \cmidrule(lr){10-17} \cmidrule{19-21}

& \multicolumn{3}{c}{General} 
&  & \multicolumn{3}{c}{CBT} 
&  & \multicolumn{2}{c}{Pos. Att.} 
&  & \multicolumn{2}{c}{Neu. Att.} 
&  & \multicolumn{2}{c}{Neg. Att.}
&  & \multirow{2}{*}{Task ($\uparrow$)}
& \multirow{2}{*}{Goal ($\uparrow$)}
& \multirow{2}{*}{Bond ($\uparrow$)} \\ 
\cmidrule(lr){2-4} \cmidrule(lr){6-8} 
\cmidrule(lr){10-11} \cmidrule(lr){13-14} \cmidrule(lr){16-17} 

& U ($\uparrow$) & I ($\uparrow$) & C ($\uparrow$)
&  & D ($\uparrow$) & F ($\uparrow$) & S ($\uparrow$)
&  & P ($\uparrow$) & N ($\downarrow$)
&  & P ($\uparrow$) & N ($\downarrow$)
&  & P ($\uparrow$) & N ($\downarrow$)
&  &  \\ 
\midrule

\model\textsubscript{-C} 
    & $\underline{3.96}$ & $\underline{4.28}$ & $\underline{3.29}^{*}$ 
    &  & $\underline{3.89}^{*}$ & $\underline{3.58}^{*}$ & $\underline{\textbf{3.00}}$ 
    &  & $\underline{\textbf{1.14}}$ & $\underline{\textbf{-0.84}}$ 
    &  & $\underline{0.78}$ & $-0.45$ 
    &  & $-0.12$ & $\underline{\textbf{-0.33}}$ 
    &  & $4.92$ & $\underline{\textbf{4.82}}$ & $\underline{\textbf{5.05}}$ \\

\model\textsubscript{-T} 
    & $3.88^{*}$ & $3.86^{*}$ & $3.09^{*}$ 
    &  & $3.85^{*}$ & $3.45^{*}$ & $2.70^{*}$ 
    &  & $1.08$ & $-0.73$ 
    &  & $\underline{0.78}$ & $-0.44$ 
    &  & $\underline{\textbf{-0.11}}$ & $-0.28$ 
    &  & $\underline{\textbf{4.95}}$ & $4.71$ & $4.90^{*}$ \\

\model\textsubscript{-C-T} 
    & $3.86^{*}$ & $3.83^{*}$ & $2.89^{*}$ 
    &  & $3.79^{*}$ & $3.42^{*}$ & $2.64^{*}$ 
    &  & $0.98^{*}$ & $-0.79$ 
    &  & $0.72^{*}$ & $\underline{\textbf{-0.47}}$ 
    &  & $-0.19$ & $-0.30$ 
    &  & $4.94$ & $4.72$ & $4.88^{*}$ \\ 
\midrule

\rowcolor{myrow} \model 
    & $\textbf{3.98}$ & $\textbf{4.30}$ & $\textbf{3.43}$ 
    &  & $\textbf{4.08}$ & $\textbf{3.76}$ & $2.93$ 
    &  & $1.13$ & $-0.73$ 
    &  & $\textbf{0.82}$ & $-0.41$ 
    &  & $-0.15$ & $-0.25$ 
    &  & $4.94$ & $4.78$ & $5.01$ \\ 
\midrule

$\delta(\%)$ 
    & $\textcolor{ForestGreen}{\textbf{+0.3}}$ & $\textcolor{ForestGreen}{\textbf{+0.3}}$ & $\textcolor{ForestGreen}{\textbf{+2.3}}$ 
    &  & $\textcolor{ForestGreen}{\textbf{+3.2}}$ & $\textcolor{ForestGreen}{\textbf{+3.0}}$ & $\textcolor{red}{\textbf{-1.2}}$ 
    &  & $\textcolor{red}{\textbf{-0.3}}$ & $\textcolor{red}{\textbf{-2.8}}$ 
    &  & $\textcolor{ForestGreen}{\textbf{+1.0}}$ & $\textcolor{red}{\textbf{-1.5}}$ 
    &  & $\textcolor{red}{\textbf{-1.0}}$ & $\textcolor{red}{\textbf{-2.0}}$ 
    &  & $\textcolor{red}{\textbf{-0.2}}$ & $\textcolor{red}{\textbf{-0.7}}$ & $\textcolor{red}{\textbf{-0.7}}$ \\ 
\bottomrule
\end{tabular}%
}
\caption{Evaluation of \model\ ablations: \model\textsubscript{-C} (no CBT agent), \model\textsubscript{-T} (no technique agent), and \model\textsubscript{-C-T} (no CBT and technique agent). Asterisks (*) indicates significant differences from \model\ ($p < 0.05$, paired t-test). For CTRS: U (Understanding), I (Interpersonal Effectiveness), C (Collaboration), D (Guided Discovery), F (Focus), S (Strategy). For PANAS: Pos. Att. (Positive Attitude), Neu. Att. (Neutral Attitude), Neg. Att. (Negative Attitude), P (average shift in positive emotions), N (average shift in negative emotions). $\delta(\%)$ shows \model’s percentage gain over the best ablation (\underline{underlined}).}
\label{tab:ctrs-panas-wai-ablation}
\end{table*}

\section{Experimental Setup}

\paragraph{Baselines and Ablations.} We compare \model\ with Psych8k \cite{chatcounselor}\footnote{Here, we re-implement Psych8k as a single-agent multi-turn generation pipeline using the original Psych8k prompt.} and CACTUS \cite{cactus}, two state-of-the-art synthetic counseling data generation pipelines, using identical client profile inputs to ensure a fair comparison. Specifically, we use 150 CACTUS profiles with three attitude variations to generate 450 40-turn dialogues per method.\footnote{Due to a few generation failures, the final datasets contain 442 \model\ sessions, 445 CACTUS sessions, and 450 Psych8k sessions.} Prompt details are provided in Appendix~\ref{sec:appendix-counselor} and Appendix~\ref{sec:appendix-client}. To evaluate downstream utility, we fine-tune Llama3-8B-Instruct \cite{meta2024introducing} on each dataset, resulting in Llama-Psych8k, Llama-CACTUS, and Llama-\model. We further conduct ablations to isolate agent contributions: \model\textsubscript{-C} (no CBT agent), \model\textsubscript{-T} (no technique agent), and \model\textsubscript{-C-T} (without both). Additional ablations removing specialized agents and ablation prompts are presented in Appendix~\ref{sec:appendix-ablations}. To evaluate the benefit of multi-agent decomposition, we compare \model\ with single-agent variants prompted to perform all \model\ tasks. Results in Appendix~\ref{sec:appendix-single-agent} show that \model\ substantially outperforms the Llama3-8B-Instruct single-agent baseline, while a Llama3.3-70B-Instruct~\cite{meta2024llama33} single-agent model surpasses \model, suggesting that multi-agent decomposition extracts more out of small models but does not overcome the benefits of scale. Finally, we report token usage and latency in Appendix~\ref{sec:appendix-tokens}. From the results, we observe that \model\ requires $3.6\times$ output tokens and wall clock time compared to the baselines. However, this is a one-time generation cost and does not affect downstream deployment of models fine-tuned using the \model-generated data. Moreover, this latency and token cost is balanced by the performance improvement, thus giving us a performance-cost trade-off.

\paragraph{LLMs Used.} In \model, Llama3-8B-Instruct is used for the CBT agent, specialized response agents, and response generation agent, while GPT-4o-mini \cite{openai2024gpt4omini} serves as the technique agent for stronger reasoning. We additionally evaluate a variant using Llama3-8B-Instruct as the technique agent, with results and technique selection analysis provided in Appendix~\ref{sec:appendix-tech-agent}. Psych8k and CACTUS use Llama3-8B-Instruct as the counselor model, and all methods use it as the client agent for consistency. To assess generalizability, we replace Llama3-8B-Instruct with Qwen2.5-7B-Instruct \cite{qwen} with results reported in Appendix~\ref{sec:appendix-qwen}. For LLM-as-a-judge evaluation, we use GPT-4o \cite{openai2024gpt4o} to score CTRS, WAI, and PANAS, motivated by its strong correlation with expert CTRS ratings \citep{cactus,chen-etal-2025-catch}. We further evaluate robustness using Qwen2.5-72B-Instruct \cite{qwen} as the judge in Appendix~\ref{sec:appendix-qwen-judge}. Similar to the GPT-4o judge, we observe that \model\ consistently outperforms the baselines with Qwen2.5-72B-Instruct judge model. Additional experimental details are provided in Appendix~\ref{sec:appendix-exp-details}.

\section{Results}

We evaluate \model-generated sessions on diversity, quality, downstream effectiveness, and expert preference. Results show that \model\ produces more diverse and psychologically grounded sessions, improves downstream fine-tuning, and is preferred by experts across evaluated aspects. Example generations are provided in Appendix~\ref{sec:appendix-examples}.

\paragraph{Data Diversity.} Table \ref{tab:diversity} presents the Distinct-$n$ scores ($n \in \{1,2,3\}$) and EAD for datasets generated by \model, Psych8k and CACTUS. \model\ consistently achieves the highest scores across all diversity metrics, highlighting its ability to produce lexically varied counseling dialogues. Crucially, the improved EAD score, which adjusts for sequence length bias, demonstrates that this variation stems from genuine structural richness rather than shorter outputs. This confirms that our multi-agent generation paradigm encourages nuanced, context-sensitive conversations, moving beyond the repetitive patterns observed in prior methods.

\paragraph{Data Quality.} As shown in Table~\ref{tab:ctrs-panas-wai}, sessions generated by \model\ outperform baselines on both general and CBT-specific counseling skills. While CACTUS also incorporates CBT planning, its single-agent generation leads to shallower application of CBT principles. In contrast, \model’s multi-agent framework achieves higher scores across all CTRS subcategories. \model\ also attains the highest WAI scores in Goal, Task, and Bond, indicating more empathetic and collaborative responses. 

On PANAS, \model\ induces larger positive emotional shifts for clients with positive or neutral attitudes, improving positive emotions more effectively than baselines. However, \model\ performs slightly worse for clients with negative attitudes, consistent with observations from CACTUS \cite{cactus}. We hypothesize that this is because models employing CBT and reflective techniques often challenge clients with negative attitudes, resulting in more negative expressions within the dialogue. These expressions may bias the judge toward assigning lower scores, even when they represent appropriate therapeutic interactions. To distinguish this effect from an actual difficulty in handling negative-attitude clients, we perform expert evaluations specifically on sessions with negative-attitude clients from Psych8k and \model, with results presented in Table~\ref{tab:negative-attitude-expert}. The results show that experts consistently prefer \model\ over the strongest baseline, Psych8k, suggesting that the lower judge scores are driven by judge bias rather than reduced counseling quality. Further details regarding standard deviations for CTRS, PANAS, and WAI across sessions are reported in Appendix Table \ref{tab:ctrs-panas-wai-sd}.

\paragraph{Counseling Agent Fine-tuning.} Fine-tuning on \model-generated data yields consistently stronger results, as presented in Table~\ref{tab:ctrs-panas-wai-model}. Llama-\model\ significantly outperforms Llama-Psych8k and Llama-CACTUS across all counseling skills and achieves stronger client alliances. PANAS results further align with trends observed in the raw data evaluation, indicating that the higher-quality synthetic data generated by \model\ translates into more effective fine-tuning of counseling agents. Standard deviations for CTRS, PANAS, and WAI across sessions are provided in Appendix Table \ref{tab:ctrs-panas-wai-model-sd}.

\paragraph{Expert Evaluation.} Figure \ref{fig:heval} shows that experts largely favor ($77.2$\%) the counseling sessions generated by \model\ in comparison to Psych8k (best baseline in automatic evaluations) across all nine aspects. This consistent preference indicates that expert judgments align with automatic evaluation results, further validating the effectiveness of \model. \model\ demonstrates the ability to generate sessions that are safer (safety), more natural (content), and also exhibit greater authenticity and professionalism, establishing a realistic and trust-building therapeutic context. Experts rate \model-generated sessions as containing clearer (directiveness), and more comprehensive (comprehensiveness) counselor responses. Furthermore, \model\ sessions excel at supportive content (supportiveness) that reinforces emotional alliance, while also enhancing expressiveness, by encouraging clients to articulate their inner experiences. \model-generated sessions also promote exploratoriness, helping clients reflect on their issues, a core aspect in counseling. We observe a similar pattern in fine-tuned models, where experts prefer Llama-\model\ over Llama-Psych8k.

\paragraph{Ablations.} Ablation results in Table~\ref{tab:ctrs-panas-wai-ablation} highlight the contribution of different components in the multi-agent counselor. Removing the CBT agent has limited impact on general skills such as understanding and interpersonal effectiveness, as the technique agent continues to guide responses. However, collaboration declines substantially, highlighting the role of structured CBT planning in supporting client involvement in decision-making. For CBT-specific skills, Strategy remains stable due to informed technique selection by the technique agent, while Guided Discovery and Focus degrade significantly, demonstrating the importance of the CBT agent for enhancing the CBT-specific counseling skills. In contrast, removing the technique agent causes broader degradation across both general and CBT-specific skills, emphasizing the importance of appropriate psychological technique selection. Removing both agents results in the lowest performance, confirming the importance of both the CBT and technique agents for generating psychologically grounded dialogue.

For WAI, Task scores remain largely unaffected by the ablations, suggesting that task understanding and agreement are preserved regardless of CBT planning or technique selection. In contrast, Goal and Bond scores decrease on removing the technique agent, likely due to reduced adaptability to evolving client needs. Interestingly, removing the CBT agent improves the Bond score. This suggests that the CBT plan, while providing structure, may introduce rigidity that weakens the counselor-client bond. Dynamic technique selection appears to mitigate this issue, as removing the technique agent leads to lower Bond scores. Removing the CBT agent also improves the Goal score, which may seem counterintuitive given that CBT provides a clear therapeutic plan. However, while the CBT agent promotes greater collaboration in goal setting (as reflected in CTRS), its structured process may inadvertently overshadow the client's preferences and evolving needs, leading to lower Goal scores. Similar to Bond, the reduction in Goal score when removing the technique agent highlights the role of dynamic strategy selection in improving the flexibility of the counseling process.

From the PANAS results, we observe that for clients with negative attitudes, removing the CBT and technique agents leads to better negative emotion regulation. This mirrors findings from CACTUS \cite{cactus} and our earlier analysis: CBT-grounded techniques deepen emotional exploration which, while beneficial for self-insight, may elevate negative affect in negative attitude clients. Standard deviations for CTRS, PANAS, and WAI across sessions are provided in Appendix Table \ref{tab:ctrs-panas-wai-ablation-sd}.

\section{Conclusion}

In this work, we propose \model, a novel multi-agent framework for synthetic counseling session generation that incorporates core psychological techniques: reflection, questioning, solution provision, normalization, and psycho-education to produce more realistic and therapeutic counselor responses. We introduce a unified evaluation framework combining CTRS, WAI, and PANAS for assessing general and CBT-specific skills, therapeutic alliance, and emotional impact. Additionally, we expand expert evaluation to nine aspects and assess downstream model effectiveness through performance of fine-tuned models. \model\ with Llama3-8B-Instruct backbone outperforms existing methods in both data diversity and quality, as validated by automatic metrics, expert preference, and improved fine-tuned model performance. 

\section*{Limitations}

While our multi-agent framework demonstrates promising improvements in synthetic counseling session generation, several limitations remain.

\paragraph{Limited Session Length and Lack of Longitudinal Structure.} Our method generates sessions consisting of only $40$ turns, which is substantially shorter than real counseling interactions. In real world counseling, sessions are often much longer, and effective counseling typically unfolds across multiple sessions with the same client. The current framework does not capture this longitudinal structure. Future work should explore generative methods capable of producing multi-session, longitudinal counseling trajectories while maintaining turn-wise dynamic technique selection and psychological grounding of counselor responses, to mirror real therapeutic processes more closely.

\paragraph{Reliable Evaluation.} Evaluating the quality and counseling validity of synthetic counseling data remains difficult. Although GPT-4o, used in an LLM-as-a-judge paradigm, showed high correlation with expert assessments, concerns remain regarding the reliability and validity of widely used psychological scales such as CTRS, WAI, and PANAS. Using multiple scales mitigates some limitations but does not fully address the absence of an objective measure of counseling competence. Our human evaluation, while more comprehensive than those in existing works, still assessed only 100 sessions (50 synthetic, 50 generated by the fine-tuned model), with two independent evaluators per session. A more robust evaluation would require larger samples and a more diverse pool of clinical experts, which was not feasible due to resource constraints.

\paragraph{Multilingualism and Multiculturalism.} The current work focuses exclusively on counseling in English. However, mental-health technologies must support multilingual and culturally diverse populations. Counselors from different cultural backgrounds use distinct communicative cues, narrative structures, and emotional expressions, and clients articulate psychological distress in culturally specific ways. However, our current framework does not account for such multilingual and multicultural synthetic counseling session generation. Future work should investigate multilingual and cross-cultural data generation frameworks that explicitly incorporate cultural norms, linguistic diversity, and culturally grounded therapeutic practices.

\paragraph{Multimodality.} Our framework generates purely text-based counseling dialogues. In real therapeutic interactions, counselors rely heavily on non-verbal information such as tone of voice, pauses, prosody, facial expressions, and other embodied cues to assess emotional states and guide interventions. A text-only representation omits these crucial signals. Extending synthetic data generation to multimodal settings, including audio and visual modalities, will be essential for training multimodal counseling models to understand these cues.

\section*{Ethics}

The objective of this work is to introduce a novel multi-agent framework for improving synthetic counseling session generation. Although the framework mitigates privacy concerns by relying exclusively on client profiles from a publicly available dataset, it does not eliminate broader safety risks associated with downstream model usage. Synthetic data, regardless of its origin, does not inherently guarantee the safety or reliability of models trained on it. While we perform a safety evaluation with experts on the generated synthetic sessions, an LLM fine-tuned on the generated dialogues may still produce clinically inappropriate, unsafe, or harmful responses when deployed in real interactions. This risk is amplified in domains like mental health, where incorrect guidance can exacerbate distress or delay individuals from seeking professional care.

Additionally, the use of synthetic data may introduce representational bias. The client profiles and counseling strategies used for generation may under-represent many cultural, linguistic, or demographic groups. As a result, models trained on this synthetic corpus may exhibit biased or culturally insensitive behavior towards populations not reflected in the source data or the generative process.

For these reasons, the synthetic counseling dataset and the methodology proposed here should not be used to fine-tune models intended for deployment in real-world clinical settings. Instead, this framework should be viewed as research towards exploring synthetic counseling session generation, benchmarking synthetic counseling session generation methodologies, or analyzing model behavior in controlled environments. Any future work extending this line of research must incorporate rigorous safety evaluation, bias auditing, domain-expert oversight and clinical trials before considering potential real-world applications.

\section*{Acknowledgments}

This research work has been funded by the German Federal Ministry of Research, Technology and Space and the Hessian Ministry of Higher Education, Research, Science and the Arts within their joint support of the National Research Center for Applied Cybersecurity ATHENE and by the DYNAMIC center, which is funded by the LOEWE program of the Hessian Ministry of Science and Arts (Grant Number: LOEWE/1/16/519/03/09.001(0009)/98). We gratefully acknowledge the support of Microsoft with a grant for access to OpenAI GPT models via the Azure cloud (Accelerate Foundation Model Academic Research). Tanmoy Chakraborty acknowledges the travel support of the Alexander von Humboldt Foundation through a Humboldt Research Fellowship for Experienced Researchers, the support of the Rajiv Khemani Young Faculty Chair Professorship in Artificial Intelligence, and the ICMR Extramural Research (Centre for Advanced Research) Project Grant (EM/DEV/CAR-2025/00725).

We thank clinical psychologists Shabdapriti G, Khushi Ambardar, Kriti Sejwal, Shreya Chawla, Abhinanda Patra, Tarushi Kaur and Muskan Gupta for their participation in the expert evaluation. We also  appreciate the feedback on the initial draft of this manuscript provided by Serwar Basch, Hiba Arnaout and Shivam Sharma.

\bibliography{custom}

@inproceedings{liu-etal-2022-rethinking,
    title = "Rethinking and Refining the Distinct Metric",
    author = "Liu, Siyang  and
      Sabour, Sahand  and
      Zheng, Yinhe  and
      Ke, Pei  and
      Zhu, Xiaoyan  and
      Huang, Minlie",
    editor = "Muresan, Smaranda  and
      Nakov, Preslav  and
      Villavicencio, Aline",
    booktitle = "Proceedings of the 60th Annual Meeting of the Association for Computational Linguistics (Volume 2: Short Papers)",
    month = may,
    year = "2022",
    address = "Dublin, Ireland",
    publisher = "Association for Computational Linguistics",
    url = "https://aclanthology.org/2022.acl-short.86/",
    doi = "10.18653/v1/2022.acl-short.86",
    pages = "762--770"
}

@article{kazdin2021extending,
  title={Extending the scalability and reach of psychosocial interventions.},
  author={Kazdin, Alan E.},
  url = {https://psycnet.apa.org/record/2021-81510-022},
  year={2021},
  journal = {Bergin and Garfield's handbook of psychotherapy and behavior change: 50th anniversary edition },
  pages = {763--789},
  publisher={John Wiley \& Sons, Inc.}
}

@article{Raile2024_chatgpt_counseling,
  author  = {Raile, Paolo},
  title   = {The usefulness of {ChatGPT} for psychotherapists and patients},
  journal = {Humanities and Social Sciences Communications},
  volume  = {11},
  number  = {1},
  pages   = {47},
  year    = {2024},
  doi     = {10.1057/s41599-023-02567-0},
  url     = {https://api.semanticscholar.org/CorpusID:266743531}
}

@article{chatgpt_gpt4_counseling,
  author       = {Birger Mo{\"{e}}ll},
  title        = {Comparing the Efficacy of {GPT-4} and {Chat-GPT} in Mental Health Care: {A} Blind Assessment of Large Language Models for Psychological Support},
  journal      = {CoRR},
  volume       = {abs/2405.09300},
  year         = {2024},
  url          = {https://doi.org/10.48550/arXiv.2405.09300},
  doi          = {10.48550/ARXIV.2405.09300},
  eprinttype   = {arXiv},
  eprint       = {2405.09300},
  bibsource    = {dblp computer science bibliography, https://dblp.org}
}

@inproceedings{pseudonymize_1,
  author       = {Buzhou Tang and
                  Dehuan Jiang and
                  Qingcai Chen and
                  Xiaolong Wang and
                  Jun Yan and
                  Ying Shen},
  title        = {De-identification of Clinical Text via {Bi-LSTM-CRF} with Neural Language Models},
  booktitle    = {{AMIA} 2019, American Medical Informatics Association Annual Symposium,
                  Washington, DC, USA, November 16-20, 2019},
  publisher    = {{AMIA}},
  pages = {857--863},
  year         = {2019},
  url          = {https://pmc.ncbi.nlm.nih.gov/articles/PMC7153082/},
  bibsource    = {dblp computer science bibliography, https://dblp.org}
}

@inproceedings{psuedonymize_2,
    title = "{PHICON}: Improving Generalization of Clinical Text De-identification Models via Data Augmentation",
    author = "Yue, Xiang  and
      Zhou, Shuang",
    editor = "Rumshisky, Anna  and
      Roberts, Kirk  and
      Bethard, Steven  and
      Naumann, Tristan",
    booktitle = "Proceedings of the 3rd Clinical Natural Language Processing Workshop",
    month = nov,
    year = "2020",
    address = "Online",
    publisher = "Association for Computational Linguistics",
    url = "https://aclanthology.org/2020.clinicalnlp-1.23/",
    doi = "10.18653/v1/2020.clinicalnlp-1.23",
    pages = "209--214"
}

@inproceedings{soulchat,
    title = "{S}oul{C}hat: Improving {LLM}s' Empathy, Listening, and Comfort Abilities through Fine-tuning with Multi-turn Empathy Conversations",
    author = "Chen, Yirong  and
      Xing, Xiaofen  and
      Lin, Jingkai  and
      Zheng, Huimin  and
      Wang, Zhenyu  and
      Liu, Qi  and
      Xu, Xiangmin",
    editor = "Bouamor, Houda  and
      Pino, Juan  and
      Bali, Kalika",
    booktitle = "Findings of the Association for Computational Linguistics: EMNLP 2023",
    month = dec,
    year = "2023",
    address = "Singapore",
    publisher = "Association for Computational Linguistics",
    url = "https://aclanthology.org/2023.findings-emnlp.83/",
    doi = "10.18653/v1/2023.findings-emnlp.83",
    pages = "1170--1183"
}

@inproceedings{smilechat,
    title = "{SMILE}: Single-turn to Multi-turn Inclusive Language Expansion via {C}hat{GPT} for Mental Health Support",
    author = "Qiu, Huachuan  and
      He, Hongliang  and
      Zhang, Shuai  and
      Li, Anqi  and
      Lan, Zhenzhong",
    editor = "Al-Onaizan, Yaser  and
      Bansal, Mohit  and
      Chen, Yun-Nung",
    booktitle = "Findings of the Association for Computational Linguistics: EMNLP 2024",
    month = nov,
    year = "2024",
    address = "Miami, Florida, USA",
    publisher = "Association for Computational Linguistics",
    url = "https://aclanthology.org/2024.findings-emnlp.34/",
    doi = "10.18653/v1/2024.findings-emnlp.34",
    pages = "615--636"
}

@inproceedings{CPsycoun,
    title = "{CP}sy{C}oun: A Report-based Multi-turn Dialogue Reconstruction and Evaluation Framework for {C}hinese Psychological Counseling",
    author = "Zhang, Chenhao  and
      Li, Renhao  and
      Tan, Minghuan  and
      Yang, Min  and
      Zhu, Jingwei  and
      Yang, Di  and
      Zhao, Jiahao  and
      Ye, Guancheng  and
      Li, Chengming  and
      Hu, Xiping",
    editor = "Ku, Lun-Wei  and
      Martins, Andre  and
      Srikumar, Vivek",
    booktitle = "Findings of the Association for Computational Linguistics: ACL 2024",
    month = aug,
    year = "2024",
    address = "Bangkok, Thailand",
    publisher = "Association for Computational Linguistics",
    url = "https://aclanthology.org/2024.findings-acl.830/",
    doi = "10.18653/v1/2024.findings-acl.830",
    pages = "13947--13966"
}

@inproceedings{llm-roleplay,
    title = "Interactive Agents: Simulating Counselor-Client Psychological Counseling via Role-Playing {LLM}-to-{LLM} Interactions",
    author = "Qiu, Huachuan  and
      Lan, Zhenzhong",
    editor = "Mohammad, Saif M.  and
      Ousidhoum, Nedjma",
    booktitle = "Proceedings of the 15th Joint Conference on Lexical and Computational Semantics (*{SEM} 2026)",
    month = jul,
    year = "2026",
    address = "San Diego, California, United States",
    publisher = "Association for Computational Linguistics",
    url = "https://aclanthology.org/2026.starsem-conference.29/",
    doi = "10.18653/v1/2026.starsem-conference.29",
    pages = "410--427",
    ISBN = "979-8-89176-413-2"
}

@inproceedings{cactus,
    title = "{Cactus}: Towards Psychological Counseling Conversations using Cognitive Behavioral Theory",
    author = "Lee, Suyeon  and
      Kim, Sunghwan  and
      Kim, Minju  and
      Kang, Dongjin  and
      Yang, Dongil  and
      Kim, Harim  and
      Kang, Minseok  and
      Jung, Dayi  and
      Kim, Min Hee  and
      Lee, Seungbeen  and
      Chung, Kyong-Mee  and
      Yu, Youngjae  and
      Lee, Dongha  and
      Yeo, Jinyoung",
    editor = "Al-Onaizan, Yaser  and
      Bansal, Mohit  and
      Chen, Yun-Nung",
    booktitle = "Findings of the Association for Computational Linguistics: EMNLP 2024",
    month = nov,
    year = "2024",
    address = "Miami, Florida, USA",
    publisher = "Association for Computational Linguistics",
    url = "https://aclanthology.org/2024.findings-emnlp.832/",
    doi = "10.18653/v1/2024.findings-emnlp.832",
    pages = "14245--14274"
}

@article{bolt,
  author       = {Yu Ying Chiu and
                  Ashish Sharma and
                  Inna Wanyin Lin and
                  Tim Althoff},
  title        = {A Computational Framework for Behavioral Assessment of {LLM} Therapists},
  journal      = {CoRR},
  volume       = {abs/2401.00820},
  year         = {2024},
  url          = {https://doi.org/10.48550/arXiv.2401.00820},
  doi          = {10.48550/ARXIV.2401.00820},
  eprinttype   = {arXiv},
  eprint       = {2401.00820},
  bibsource    = {dblp computer science bibliography, https://dblp.org}
}

@article{ctrs,
title = {The Structure of Competence: Evaluating the Factor Structure of the Cognitive Therapy Rating Scale},
journal = {Behavior Therapy},
volume = {51},
number = {1},
pages = {113-122},
year = {2020},
issn = {0005-7894},
doi = {https://doi.org/10.1016/j.beth.2019.05.008},
url = {https://www.sciencedirect.com/science/article/pii/S0005789419300632},
author = {Simon B. Goldberg and Scott A. Baldwin and Kritzia Merced and Derek D. Caperton and Zac E. Imel and David C. Atkins and Torrey Creed}
}

@article{panas,
  title={Development and validation of brief measures of positive and negative affect: the {PANAS} scales.},
  author={Watson, David and Clark, Lee Anna and Tellegen, Auke},
  journal={Journal of personality and social psychology},
  volume={54},
  number={6},
  pages={1063--1070},
  year={1988},
  publisher={American Psychological Association},
  url = {https://psycnet.apa.org/buy/1988-31508-001},
  doi     = {10.1037/0022-3514.54.6.1063}
}

@article{chatcounselor,
  author       = {June M. Liu and
                  Donghao Li and
                  He Cao and
                  Tianhe Ren and
                  Zeyi Liao and
                  Jiamin Wu},
  title        = {{ChatCounselor}: {A} Large Language Models for Mental Health Support},
  journal      = {CoRR},
  volume       = {abs/2309.15461},
  year         = {2023},
  url          = {https://doi.org/10.48550/arXiv.2309.15461},
  doi          = {10.48550/ARXIV.2309.15461},
  eprinttype   = {arXiv},
  eprint       = {2309.15461},
  bibsource    = {dblp computer science bibliography, https://dblp.org}
}

@inproceedings{meta-gpt,
  author       = {Sirui Hong and
                  Mingchen Zhuge and
                  Jonathan Chen and
                  Xiawu Zheng and
                  Yuheng Cheng and
                  Jinlin Wang and
                  Ceyao Zhang and
                  Zili Wang and
                  Steven Ka Shing Yau and
                  Zijuan Lin and
                  Liyang Zhou and
                  Chenyu Ran and
                  Lingfeng Xiao and
                  Chenglin Wu and
                  J{\"{u}}rgen Schmidhuber},
  title        = {{MetaGPT:} {M}eta {P}rogramming for {A} {M}ulti-{A}gent {C}ollaborative {F}ramework},
  pages = {23247--23275},
  booktitle    = {The Twelfth International Conference on Learning Representations,
                  {ICLR} 2024, Vienna, Austria, May 7-11, 2024},
  publisher    = {OpenReview.net},
  year         = {2024},
  url = {https://proceedings.iclr.cc/paper_files/paper/2024/file/6507b115562bb0a305f1958ccc87355a-Paper-Conference.pdf},
  bibsource    = {dblp computer science bibliography, https://dblp.org}
}

@inproceedings{chatdev,
    title = "{C}hat{D}ev: Communicative Agents for Software Development",
    author = "Qian, Chen  and
      Liu, Wei  and
      Liu, Hongzhang  and
      Chen, Nuo  and
      Dang, Yufan  and
      Li, Jiahao  and
      Yang, Cheng  and
      Chen, Weize  and
      Su, Yusheng  and
      Cong, Xin  and
      Xu, Juyuan  and
      Li, Dahai  and
      Liu, Zhiyuan  and
      Sun, Maosong",
    editor = "Ku, Lun-Wei  and
      Martins, Andre  and
      Srikumar, Vivek",
    booktitle = "Proceedings of the 62nd Annual Meeting of the Association for Computational Linguistics (Volume 1: Long Papers)",
    month = aug,
    year = "2024",
    address = "Bangkok, Thailand",
    publisher = "Association for Computational Linguistics",
    url = "https://aclanthology.org/2024.acl-long.810/",
    doi = "10.18653/v1/2024.acl-long.810",
    pages = "15174--15186"
}

@inproceedings{autoact,
    title = "{A}uto{A}ct: Automatic Agent Learning from Scratch for {QA} via Self-Planning",
    author = "Qiao, Shuofei  and
      Zhang, Ningyu  and
      Fang, Runnan  and
      Luo, Yujie  and
      Zhou, Wangchunshu  and
      Jiang, Yuchen  and
      Lv, Chengfei  and
      Chen, Huajun",
    editor = "Ku, Lun-Wei  and
      Martins, Andre  and
      Srikumar, Vivek",
    booktitle = "Proceedings of the 62nd Annual Meeting of the Association for Computational Linguistics (Volume 1: Long Papers)",
    month = aug,
    year = "2024",
    address = "Bangkok, Thailand",
    publisher = "Association for Computational Linguistics",
    url = "https://aclanthology.org/2024.acl-long.165/",
    doi = "10.18653/v1/2024.acl-long.165",
    pages = "3003--3021"
}

@article{multi-agent-task-oriented,
  author       = {Jingtao Sun and
                  Jiayin Kou and
                  Weipeng Shi and
                  Wenyan Hou},
  title        = {A multi-agent collaborative algorithm for task-oriented dialogue systems},
  journal      = {International Journal of Machine Learning and Cybernetics},
  volume       = {16},
  number       = {3},
  pages        = {2009--2022},
  year         = {2025},
  url          = {https://doi.org/10.1007/s13042-024-02374-2},
  doi          = {10.1007/S13042-024-02374-2},
  bibsource    = {dblp computer science bibliography, https://dblp.org}
}

@article{multi-agent-reco,
  author       = {Jiabao Fang and
                  Shen Gao and
                  Pengjie Ren and
                  Xiuying Chen and
                  Suzan Verberne and
                  Zhaochun Ren},
  title        = {A Multi-Agent Conversational Recommender System},
  journal      = {CoRR},
  volume       = {abs/2402.01135},
  year         = {2024},
  url          = {https://doi.org/10.48550/arXiv.2402.01135},
  doi          = {10.48550/ARXIV.2402.01135},
  eprinttype   = {arXiv},
  eprint       = {2402.01135},
  bibsource    = {dblp computer science bibliography, https://dblp.org}
}

@article{wai-psych,
  title={Development and validation of the {W}orking {A}lliance {I}nventory.},
  author={Horvath, Adam O. and Greenberg, Leslie S.},
  journal={Journal of counseling psychology},
  volume={36},
  number={2},
  pages={223--233},
  year={1989},
  publisher={American Psychological Association},
  url = {https://psycnet.apa.org/record/1989-24600-001},
  doi     = {10.1037/0022-0167.36.2.223}
}

@inproceedings{wai-comp,
  author       = {Sebastian Peter Bayerl and
                  Gabriel Roccabruna and
                  Shammur Absar Chowdhury and
                  Tommaso Ciulli and
                  Morena Danieli and
                  Korbinian Riedhammer and
                  Giuseppe Riccardi},
  editor       = {Hanseok Ko and
                  John H. L. Hansen},
  title        = {What can Speech and Language Tell us About the Working Alliance in Psychotherapy},
  booktitle    = {23rd Annual Conference of the International Speech Communication Association,
                  Interspeech 2022, Incheon, Korea, September 18-22, 2022},
  pages        = {2443--2447},
  publisher    = {{ISCA}},
  year         = {2022},
  url          = {https://doi.org/10.21437/Interspeech.2022-347},
  doi          = {10.21437/INTERSPEECH.2022-347},
  bibsource    = {dblp computer science bibliography, https://dblp.org}
}

@inproceedings{bleu,
    title = "{BLEU}: a Method for Automatic Evaluation of Machine Translation",
    author = "Papineni, Kishore  and
      Roukos, Salim  and
      Ward, Todd  and
      Zhu, Wei-Jing",
    editor = "Isabelle, Pierre  and
      Charniak, Eugene  and
      Lin, Dekang",
    booktitle = "Proceedings of the 40th Annual Meeting of the Association for Computational Linguistics",
    month = jul,
    year = "2002",
    address = "Philadelphia, Pennsylvania, USA",
    publisher = "Association for Computational Linguistics",
    url = "https://aclanthology.org/P02-1040/",
    doi = "10.3115/1073083.1073135",
    pages = "311--318"
}

@inproceedings{lin-2004-rouge,
    title = "{ROUGE}: A Package for Automatic Evaluation of Summaries",
    author = "Lin, Chin-Yew",
    booktitle = "Text Summarization Branches Out",
    month = jul,
    year = "2004",
    address = "Barcelona, Spain",
    publisher = "Association for Computational Linguistics",
    url = "https://aclanthology.org/W04-1013/",
    pages = "74--81"
}

@inproceedings{bertscore,
  author       = {Tianyi Zhang and
                  Varsha Kishore and
                  Felix Wu and
                  Kilian Q. Weinberger and
                  Yoav Artzi},
  title        = {{BERTScore:} Evaluating Text Generation with {BERT}},
  booktitle    = {8th International Conference on Learning Representations, {ICLR} 2020,
                  Addis Ababa, Ethiopia, April 26-30, 2020},
  publisher    = {OpenReview.net},
  year         = {2020},
  url          = {https://openreview.net/forum?id=SkeHuCVFDr},
  bibsource    = {dblp computer science bibliography, https://dblp.org}
}

@inproceedings{cognitive-reframe,
    title = "Cognitive Reframing of Negative Thoughts through Human-Language Model Interaction",
    author = "Sharma, Ashish  and
      Rushton, Kevin  and
      Lin, Inna  and
      Wadden, David  and
      Lucas, Khendra  and
      Miner, Adam  and
      Nguyen, Theresa  and
      Althoff, Tim",
    editor = "Rogers, Anna  and
      Boyd-Graber, Jordan  and
      Okazaki, Naoaki",
    booktitle = "Proceedings of the 61st Annual Meeting of the Association for Computational Linguistics (Volume 1: Long Papers)",
    month = jul,
    year = "2023",
    address = "Toronto, Canada",
    publisher = "Association for Computational Linguistics",
    url = "https://aclanthology.org/2023.acl-long.555/",
    doi = "10.18653/v1/2023.acl-long.555",
    pages = "9977--10000"
}

@inproceedings{psyqa,
    title = "{P}sy{QA}: A {C}hinese Dataset for Generating Long Counseling Text for Mental Health Support",
    author = "Sun, Hao  and
      Lin, Zhenru  and
      Zheng, Chujie  and
      Liu, Siyang  and
      Huang, Minlie",
    editor = "Zong, Chengqing  and
      Xia, Fei  and
      Li, Wenjie  and
      Navigli, Roberto",
    booktitle = "Findings of the Association for Computational Linguistics: ACL-IJCNLP 2021",
    month = aug,
    year = "2021",
    address = "Online",
    publisher = "Association for Computational Linguistics",
    url = "https://aclanthology.org/2021.findings-acl.130/",
    doi = "10.18653/v1/2021.findings-acl.130",
    pages = "1489--1503"
}

@article{Powles1974BeckAT,
  title={{Beck, Aaron T.} {D}epression: {C}auses and {T}reatment. {P}hiladelphia: {U}niversity of {P}ennsylvania Press, 1972. Pp. 370. \$4.45},
  author={William E. Powles},
  journal={American Journal of Clinical Hypnosis},
  year={1974},
  volume={16},
  pages={281-282},
  url={https://api.semanticscholar.org/CorpusID:143508667}
}

@incollection{greimel2011cognitive,
author="Greimel, Karoline V.
and Kr{\"o}ner-Herwig, Birgit",
editor="M{\o}ller, Aage R.
and Langguth, Berthold
and De Ridder, Dirk
and Kleinjung, Tobias",
title="{Cognitive Behavioral Treatment (CBT)}",
booktitle="Textbook of Tinnitus",
year="2011",
publisher="Springer New York",
address="New York, NY",
pages="557--561",
isbn="978-1-60761-145-5",
doi="10.1007/978-1-60761-145-5_71",
url="https://doi.org/10.1007/978-1-60761-145-5_71"
}

@inproceedings{distinct_n,
    title = "A Diversity-Promoting Objective Function for Neural Conversation Models",
    author = "Li, Jiwei  and
      Galley, Michel  and
      Brockett, Chris  and
      Gao, Jianfeng  and
      Dolan, Bill",
    editor = "Knight, Kevin  and
      Nenkova, Ani  and
      Rambow, Owen",
    booktitle = "Proceedings of the 2016 Conference of the North {A}merican Chapter of the Association for Computational Linguistics: Human Language Technologies",
    month = jun,
    year = "2016",
    address = "San Diego, California",
    publisher = "Association for Computational Linguistics",
    url = "https://aclanthology.org/N16-1014/",
    doi = "10.18653/v1/N16-1014",
    pages = "110--119"
}

@inproceedings{qlora,
  author       = {Tim Dettmers and
                  Artidoro Pagnoni and
                  Ari Holtzman and
                  Luke Zettlemoyer},
  editor       = {Alice Oh and
                  Tristan Naumann and
                  Amir Globerson and
                  Kate Saenko and
                  Moritz Hardt and
                  Sergey Levine},
  title        = {{QLoRA}: Efficient Finetuning of Quantized {LLM}s},
  booktitle    = {Advances in Neural Information Processing Systems 36: Annual Conference
                  on Neural Information Processing Systems 2023, NeurIPS 2023, New Orleans,
                  LA, USA, December 10 - 16, 2023},
  year         = {2023},
   pages = {10088--10115},
 publisher = {Curran Associates, Inc.},
 doi = {10.52202/075280-0441},
  url = {https://proceedings.neurips.cc/paper_files/paper/2023/file/1feb87871436031bdc0f2beaa62a049b-Paper-Conference.pdf},
  bibsource    = {dblp computer science bibliography, https://dblp.org}
}

@article{meta2024introducing,
  author       = {Llama Team},
  title        = {The {Llama 3} Herd of Models},
  journal      = {CoRR},
  volume       = {abs/2407.21783},
  year         = {2024},
  url          = {https://doi.org/10.48550/arXiv.2407.21783},
  doi          = {10.48550/ARXIV.2407.21783},
  eprinttype   = {arXiv},
  eprint       = {2407.21783},
  bibsource    = {dblp computer science bibliography, https://dblp.org}
}

@misc{meta2024llama33,
  title  = {Llama 3.3 Model Card},
  author = {{Meta AI}},
  year   = {2024},
  month  = {December},
  url    = {https://github.com/meta-llama/llama-models/blob/main/models/llama3_3/MODEL_CARD.md},
  note         = {Accessed: 2026-08-11}
  
}

@article{mccullough1988psychotherapy,
  title={Psychotherapy interaction coding system manual: the {PIC} system},
  author={McCullough, L},
  journal={Social \& Behavioural Science Documents},
  volume={18},
  year={1988}
}

@inproceedings{vllm,
author = {Kwon, Woosuk and Li, Zhuohan and Zhuang, Siyuan and Sheng, Ying and Zheng, Lianmin and Yu, Cody Hao and Gonzalez, Joseph and Zhang, Hao and Stoica, Ion},
title = {Efficient Memory Management for Large Language Model Serving with {PagedAttention}},
year = {2023},
isbn = {9798400702297},
publisher = {Association for Computing Machinery},
address = {New York, NY, USA},
url = {https://doi.org/10.1145/3600006.3613165},
doi = {10.1145/3600006.3613165},
booktitle = {Proceedings of the 29th Symposium on Operating Systems Principles},
pages = {611–626},
numpages = {16},
location = {Koblenz, Germany},
series = {SOSP '23}
}

@inproceedings{lee-etal-2019-identifying,
    title = "Identifying therapist conversational actions across diverse psychotherapeutic approaches",
    author = "Lee, Fei-Tzin  and
      Hull, Derrick  and
      Levine, Jacob  and
      Ray, Bonnie  and
      McKeown, Kathy",
    editor = "Niederhoffer, Kate  and
      Hollingshead, Kristy  and
      Resnik, Philip  and
      Resnik, Rebecca  and
      Loveys, Kate",
    booktitle = "Proceedings of the Sixth Workshop on Computational Linguistics and Clinical Psychology",
    month = jun,
    year = "2019",
    address = "Minneapolis, Minnesota",
    publisher = "Association for Computational Linguistics",
    url = "https://aclanthology.org/W19-3002/",
    doi = "10.18653/v1/W19-3002",
    pages = "12--23"
}

@inproceedings{cao-etal-2019,
    title = "Observing Dialogue in Therapy: Categorizing and Forecasting Behavioral Codes",
    author = "Cao, Jie  and
      Tanana, Michael  and
      Imel, Zac  and
      Poitras, Eric  and
      Atkins, David  and
      Srikumar, Vivek",
    editor = "Korhonen, Anna  and
      Traum, David  and
      M{\`a}rquez, Llu{\'i}s",
    booktitle = "Proceedings of the 57th Annual Meeting of the Association for Computational Linguistics",
    month = jul,
    year = "2019",
    address = "Florence, Italy",
    publisher = "Association for Computational Linguistics",
    url = "https://aclanthology.org/P19-1563/",
    doi = "10.18653/v1/P19-1563",
    pages = "5599--5611"
}

@article{cbt1,
author="Fitzpatrick, Kathleen Kara
and Darcy, Alison
and Vierhile, Molly",
title="Delivering Cognitive Behavior Therapy to Young Adults With Symptoms of Depression and Anxiety Using a Fully Automated Conversational Agent (Woebot): A Randomized Controlled Trial",
journal="JMIR Ment Health",
year="2017",
month="Jun",
day="06",
volume="4",
number="2",
pages="e19",
issn="2368-7959",
doi="10.2196/mental.7785",
url="http://mental.jmir.org/2017/2/e19/"
}

@article{cbt2,
author="Haque, M D Romael
and Rubya, Sabirat",
title="An Overview of Chatbot-Based Mobile Mental Health Apps: Insights From App Description and User Reviews",
journal="JMIR Mhealth Uhealth",
year="2023",
month="May",
day="22",
volume="11",
pages="e44838",
issn="2291-5222",
doi="10.2196/44838",
url="https://mhealth.jmir.org/2023/1/e44838"
}

@article{cbt3,
author="Mehta, Ashish
and Niles, Andrea Nicole
and Vargas, Jose Hamilton
and Marafon, Thiago
and Couto, Diego Dotta
and Gross, James Jonathan",
title="Acceptability and Effectiveness of Artificial Intelligence Therapy for Anxiety and Depression (Youper): Longitudinal Observational Study",
journal="J Med Internet Res",
year="2021",
month="Jun",
day="22",
volume="23",
number="6",
pages="e26771",
issn="1438-8871",
doi="10.2196/26771",
url="https://www.jmir.org/2021/6/e26771"
}

@misc{openai2024gpt4o,
  title        = {{GPT-4o} System Card},
  author       = {OpenAI},
  year         = {2024},
  month        = {May},
  url          = {https://openai.com/index/gpt-4o-system-card/},
  note         = {Accessed: 2026-08-11}
}

@misc{openai2024gpt4omini,
  title        = {{GPT‑4o mini:} advancing cost-efficient intelligence},
  author       = {OpenAI},
  year         = {2024},
  month        = {May},
  url          = {https://openai.com/index/gpt-4o-mini-advancing-cost-efficient-intelligence/},
  note         = {Accessed: 2026-08-11}
}

@article{DEDURO2025100170,
title = {Introducing {CounseLLMe}: A dataset of simulated mental health dialogues for comparing {LLM}s like {Haiku}, {LLaMAntino} and {ChatGPT} against humans},
journal = {Emerging Trends in Drugs, Addictions, and Health},
volume = {5},
pages = {100170},
year = {2025},
issn = {2667-1182},
doi = {https://doi.org/10.1016/j.etdah.2025.100170},
url = {https://www.sciencedirect.com/science/article/pii/S2667118225000017},
author = {Edoardo Sebastiano {De Duro} and Riccardo Improta and Massimo Stella}
}

@article{MADP,
  author       = {Qi Chen and
                  Dexi Liu},
  title        = {{MADP:} Multi{-}Agent Deductive Planning for Enhanced Cognitive{-}Behavioral Mental Health Question Answer},
  journal      = {CoRR},
  volume       = {abs/2501.15826},
  year         = {2025},
  url          = {https://doi.org/10.48550/arXiv.2501.15826},
  doi          = {10.48550/ARXIV.2501.15826},
  eprinttype    = {arXiv},
  eprint       = {2501.15826},
  bibsource    = {dblp computer science bibliography, https://dblp.org}
}

@article{llm_review_1,
  title={Using large language models in psychology},
  author={Demszky, Dorottya and Yang, Diyi and Yeager, David S. and Bryan, Christopher J. and Clapper, Margarett and Chandhok, Susannah and Eichstaedt, Johannes C. and Hecht, Cameron and Jamieson, Jeremy and Johnson, Meghann and Jones, Michaela and Krettek-Cobb, Danielle and Lai, Leslie and JonesMitchell, Nirel and Ong, Desmond C. and Dweck, Carol S. and Gross, James J. and Pennebaker, James W.},
  journal={Nature Reviews Psychology},
  volume={2},
  number={11},
  pages={688--701},
  year={2023},
  publisher={Nature Publishing Group US New York},
url = {https://www.nature.com/articles/s44159-023-00241-5},
doi     = {10.1038/s44159-023-00241-5}
}

@inproceedings{trustworthy,
author = {Ozgun, Mithat Can and Pei, Jiahuan and Hindriks, Koen and Donatelli, Lucia and Liu, Qingzhi and Wang, Junxiao},
title = {Trustworthy {AI} Psychotherapy: Multi-Agent {LLM} Workflow for Counseling and Explainable Mental Disorder Diagnosis},
year = {2025},
isbn = {9798400720406},
publisher = {Association for Computing Machinery},
address = {New York, NY, USA},
url = {https://doi.org/10.1145/3746252.3761164},
doi = {10.1145/3746252.3761164},
booktitle = {Proceedings of the 34th ACM International Conference on Information and Knowledge Management},
pages = {2263–2272},
numpages = {10},
location = {Seoul, Republic of Korea},
series = {CIKM '25}
}

@article{qwen,
  author       = {An Yang and
                  Baosong Yang and
                  Beichen Zhang and
                  Binyuan Hui and
                  Bo Zheng and
                  Bowen Yu and
                  Chengyuan Li and
                  Dayiheng Liu and
                  Fei Huang and
                  Haoran Wei and
                  Huan Lin and
                  Jian Yang and
                  Jianhong Tu and
                  Jianwei Zhang and
                  Jianxin Yang and
                  Jiaxi Yang and
                  Jingren Zhou and
                  Junyang Lin and
                  Kai Dang and
                  Keming Lu and
                  Keqin Bao and
                  Kexin Yang and
                  Le Yu and
                  Mei Li and
                  Mingfeng Xue and
                  Pei Zhang and
                  Qin Zhu and
                  Rui Men and
                  Runji Lin and
                  Tianhao Li and
                  Tingyu Xia and
                  Xingzhang Ren and
                  Xuancheng Ren and
                  Yang Fan and
                  Yang Su and
                  Yichang Zhang and
                  Yu Wan and
                  Yuqiong Liu and
                  Zeyu Cui and
                  Zhenru Zhang and
                  Zihan Qiu},
  title        = {Qwen2.5 Technical Report},
  journal      = {CoRR},
  volume       = {abs/2412.15115},
  year         = {2024},
  url          = {https://doi.org/10.48550/arXiv.2412.15115},
  doi          = {10.48550/ARXIV.2412.15115},
  eprinttype    = {arXiv},
  eprint       = {2412.15115},
  bibsource    = {dblp computer science bibliography, https://dblp.org}
}

@inproceedings{counselingbench,
    title = "Do Large Language Models Align with Core Mental Health Counseling Competencies?",
    author = "Nguyen, Viet Cuong  and
      Taher, Mohammad  and
      Hong, Dongwan  and
      Possobom, Vinicius Konkolics  and
      Gopalakrishnan, Vibha Thirunellayi  and
      Raj, Ekta  and
      Li, Zihang  and
      Soled, Heather J.  and
      Birnbaum, Michael L.  and
      Kumar, Srijan  and
      De Choudhury, Munmun",
    editor = "Chiruzzo, Luis  and
      Ritter, Alan  and
      Wang, Lu",
    booktitle = "Findings of the Association for Computational Linguistics: NAACL 2025",
    month = apr,
    year = "2025",
    address = "Albuquerque, New Mexico",
    publisher = "Association for Computational Linguistics",
    url = "https://aclanthology.org/2025.findings-naacl.418/",
    doi = "10.18653/v1/2025.findings-naacl.418",
    pages = "7503--7526",
    ISBN = "979-8-89176-195-7"
}

@inproceedings{roscoe,
  author       = {Olga Golovneva and
                  Moya Peng Chen and
                  Spencer Poff and
                  Martin Corredor and
                  Luke Zettlemoyer and
                  Maryam Fazel{-}Zarandi and
                  Asli Celikyilmaz},
  title        = {{ROSCOE:} {A} Suite of Metrics for Scoring Step-by-Step Reasoning},
  booktitle    = {The Eleventh International Conference on Learning Representations,
                  {ICLR} 2023, Kigali, Rwanda, May 1-5, 2023},
  publisher    = {OpenReview.net},
  year         = {2023},
  url          = {https://openreview.net/forum?id=xYlJRpzZtsY},
  bibsource    = {dblp computer science bibliography, https://dblp.org}
}

@inproceedings{sentence-transformer,
    title = "Sentence-{BERT}: Sentence Embeddings using {S}iamese {BERT}-Networks",
    author = "Reimers, Nils  and
      Gurevych, Iryna",
    editor = "Inui, Kentaro  and
      Jiang, Jing  and
      Ng, Vincent  and
      Wan, Xiaojun",
    booktitle = "Proceedings of the 2019 Conference on Empirical Methods in Natural Language Processing and the 9th International Joint Conference on Natural Language Processing (EMNLP-IJCNLP)",
    month = nov,
    year = "2019",
    address = "Hong Kong, China",
    publisher = "Association for Computational Linguistics",
    url = "https://aclanthology.org/D19-1410/",
    doi = "10.18653/v1/D19-1410",
    pages = "3982--3992"
}

@inproceedings{mishra-etal-2023-e,
    title = "e-{THERAPIST}: {I} suggest you to cultivate a mindset of positivity and nurture uplifting thoughts",
    author = "Mishra, Kshitij  and
      Priya, Priyanshu  and
      Burja, Manisha  and
      Ekbal, Asif",
    editor = "Bouamor, Houda  and
      Pino, Juan  and
      Bali, Kalika",
    booktitle = "Proceedings of the 2023 Conference on Empirical Methods in Natural Language Processing",
    month = dec,
    year = "2023",
    address = "Singapore",
    publisher = "Association for Computational Linguistics",
    url = "https://aclanthology.org/2023.emnlp-main.861/",
    doi = "10.18653/v1/2023.emnlp-main.861",
    pages = "13952--13967"
}

@inproceedings{na-2024-cbt,
    title = "{CBT}-{LLM}: A {C}hinese Large Language Model for Cognitive Behavioral Therapy-based Mental Health Question Answering",
    author = "Na, Hongbin",
    editor = "Calzolari, Nicoletta  and
      Kan, Min-Yen  and
      Hoste, Veronique  and
      Lenci, Alessandro  and
      Sakti, Sakriani  and
      Xue, Nianwen",
    booktitle = "Proceedings of the 2024 Joint International Conference on Computational Linguistics, Language Resources and Evaluation (LREC-COLING 2024)",
    month = may,
    year = "2024",
    address = "Torino, Italia",
    publisher = "ELRA and ICCL",
    url = "https://aclanthology.org/2024.lrec-main.261/",
    pages = "2930--2940"
}

@inproceedings{xiao-etal-2024-healme,
    title = "{H}eal{M}e: Harnessing Cognitive Reframing in Large Language Models for Psychotherapy",
    author = "Xiao, Mengxi  and
      Xie, Qianqian  and
      Kuang, Ziyan  and
      Liu, Zhicheng  and
      Yang, Kailai  and
      Peng, Min  and
      Han, Weiguang  and
      Huang, Jimin",
    editor = "Ku, Lun-Wei  and
      Martins, Andre  and
      Srikumar, Vivek",
    booktitle = "Proceedings of the 62nd Annual Meeting of the Association for Computational Linguistics (Volume 1: Long Papers)",
    month = aug,
    year = "2024",
    address = "Bangkok, Thailand",
    publisher = "Association for Computational Linguistics",
    url = "https://aclanthology.org/2024.acl-long.93/",
    doi = "10.18653/v1/2024.acl-long.93",
    pages = "1707--1725"
}

@inproceedings{bn2025thousand,
  author       = {Suhas BN and
                  Andrew M. Sherrill and
                  Rosa I. Arriaga and
                  Christopher W. Wiese and
                  Saeed Abdullah},
  editor       = {Danielle Belgrave and
                  Cheng Zhang and
                  Laura N. Montoya and
                  Hsuan{-}Tien Lin and
                  Razvan Pascanu and
                  Piotr Koniusz and
                  Marzyeh Ghassemi and
                  Nancy Chen and
                  Iv{\'{a}}n Vladimir Meza Ru{\'{\i}}z and
                  Arturo Loaiza{-}Bonilla},
  title        = {Thousand Voices of Trauma: {A} Large-Scale Synthetic Dataset for Modeling Prolonged Exposure Therapy Conversations},
  doi = {10.52202/085713-2092},
 publisher = {Curran Associates, Inc.},
  booktitle    = {Advances in Neural Information Processing Systems 38: Annual Conference on Neural Information Processing Systems 2025, NeurIPS 2025, San Diego, CA, USA, December 2-7, 2025 / Mexico City, Mexico, November 30 - December 5, 2025},
  year         = {2025},
  url = {https://proceedings.neurips.cc/paper_files/paper/2025/file/5a29b161cf764f95e71488f07ecd64c0-Paper-Datasets_and_Benchmarks_Track.pdf},
  bibsource    = {dblp computer science bibliography, https://dblp.org}
}

@inproceedings{mentalchat16k,
author = {Xu, Jia and Wei, Tianyi and Hou, Bojian and Orzechowski, Patryk and Yang, Shu and Jin, Ruochen and Paulbeck, Rachael and Wagenaar, Joost and Demiris, George and Shen, Li},
title = {{MentalChat16K:} A Benchmark Dataset for Conversational Mental Health Assistance},
year = {2025},
isbn = {9798400714542},
publisher = {Association for Computing Machinery},
address = {New York, NY, USA},
url = {https://doi.org/10.1145/3711896.3737393},
doi = {10.1145/3711896.3737393},
booktitle = {Proceedings of the 31st ACM SIGKDD Conference on Knowledge Discovery and Data Mining V.2},
pages = {5367–5378},
numpages = {12},
location = {Toronto ON, Canada},
series = {KDD '25}
}

@article{sqpconv,
  author       = {Doan Nam Long Vu and
                  Rui Tan and
                  Lena Moench and
                  Svenja Jule Francke and
                  Daniel Woiwod and
                  Florian Thomas{-}Odenthal and
                  Sanna Stroth and
                  Tilo Kircher and
                  Christiane Hermann and
                  Udo Dannlowski and
                  Hamidreza Jamalabadi and
                  Shaoxiong Ji},
  title        = {Roleplaying with Structure: Synthetic Therapist-Client Conversation Generation from Questionnaires},
  journal      = {CoRR},
  volume       = {abs/2510.25384},
  year         = {2025},
  url          = {https://doi.org/10.48550/arXiv.2510.25384},
  doi          = {10.48550/ARXIV.2510.25384},
  eprinttype   = {arXiv},
  eprint       = {2510.25384},
  bibsource    = {dblp computer science bibliography, https://dblp.org}
}

@inproceedings{chen-etal-2025-catch,
    title = "{CATCH}: A Novel Data Synthesis Framework for High Therapy Fidelity and Memory-Driven Planning Chain of Thought in {AI} Counseling",
    author = "Chen, Mingyu  and
      Lin, Jingkai  and
      Chu, Zhaojie  and
      Xing, Xiaofen  and
      Chen, Yirong  and
      Xu, Xiangmin",
    editor = "Christodoulopoulos, Christos  and
      Chakraborty, Tanmoy  and
      Rose, Carolyn  and
      Peng, Violet",
    booktitle = "Findings of the Association for Computational Linguistics: EMNLP 2025",
    month = nov,
    year = "2025",
    address = "Suzhou, China",
    publisher = "Association for Computational Linguistics",
    url = "https://aclanthology.org/2025.findings-emnlp.543/",
    doi = "10.18653/v1/2025.findings-emnlp.543",
    pages = "10254--10286",
    ISBN = "979-8-89176-335-7"
}

@article{graph2counsel,
  author       = {Aishik Mandal and
                  Hiba Arnaout and
                  Clarissa W. Ong and
                  Juliet Bockhorst and
                  Kate Sheehan and
                  Rachael Moldow and
                  Tanmoy Chakraborty and
                  Iryna Gurevych},
  title        = {{Graph2Counsel}: Clinically Grounded Synthetic Counseling Dialogue Generation
                  from Client Psychological Graphs},
  journal      = {CoRR},
  volume       = {abs/2604.20382},
  year         = {2026},
  url          = {https://doi.org/10.48550/arXiv.2604.20382},
  doi          = {10.48550/ARXIV.2604.20382},
  eprinttype   = {arXiv},
  eprint       = {2604.20382},
  bibsource    = {dblp computer science bibliography, https://dblp.org}
}

@unpublished{young1980ctrs,
  author = {Young, Jeffrey E. and Beck, Aaron T.},
  title  = {Cognitive {T}herapy {S}cale: {R}ating {M}anual},
  year   = {1980},
  url = {https://beckinstitute.org/wp-content/uploads/2021/06/CTRS-Manual-2020.pdf},
  note   = {Unpublished manuscript, University of Pennsylvania}
}

\appendix

\section{Multi-Agent Counselor Simulation}
\label{sec:appendix-counselor}

Here, we describe the counselor agent used in the baseline methods: Psych8k \cite{chatcounselor} and CACTUS \cite{cactus}, as well as provide more details regarding the agents used for simulating the counselor in \model. In Psych8k, a single LLM agent produces the counselor’s response based on the current dialogue history. The prompt used for this agent is shown in Figure \ref{fig:psych8k}. CACTUS, on the other hand, uses two agents to generate the counselor's response. First, a CBT planning agent generates a counseling plan using the client intake form and the client's initial greeting dialogue. An example intake form from CACTUS is shown in Figure \ref{fig:intake}. The prompt for the CBT agent is provided in Figure \ref{fig:cbt}. Following this, CACTUS uses a response generation agent to produce the final counselor response based on the current dialogue history and the generated counseling plan. The prompt for the response generation agent used in CACTUS is shown in Figure \ref{fig:cactus}. 

Our proposed framework, \model, expands this pipeline into a modular multi-agent architecture that decomposes counselor response generation into specialized sub-tasks. \model\ reuses the CBT planning agent from CACTUS (the prompt used is shown in Figure \ref{fig:cbt}) and introduces a set of specialized response agents, each focused on a distinct therapeutic technique: reflection, questioning, solution provision, normalization, and psycho-education. These agents generate candidate responses using the current dialogue history and the given client profile. The prompt for the reflection agent, questioning agent, solutions agent, normalizing agent and psycho-education agent are shown in Figure \ref{fig:reflection}, Figure \ref{fig:questioning}, Figure \ref{fig:solutions}, Figure \ref{fig:normalizing} and Figure \ref{fig:psychoed} respectively. Additionally, \model\ includes a technique agent that recommends a subset of relevant therapeutic techniques for the current turn, informed by the counseling plan and dialogue context. Figure \ref{fig:technique} shows the prompt used for the technique agent. Finally, a response generation agent combines candidate responses from the specialized agents based on the technique recommendations from the technique agent to generate the final counselor response. The prompt used for this response generation agent is shown in Figure \ref{fig:dialogue_gen}.

\begin{figure*}[t]
\centering
\begin{tcolorbox}[colback=gray!5!white, colframe=gray!70!black, title=Psych8k Counselor Agent Prompt]
<|start\_header\_id|>system<|end\_header\_id|>

If you are a counselor, please answer the questions based on the description of the patient. Please be mindful to only generate the counselor response for a single turn, and do not include extra text like "here is the next counselor utterance" or "Here is a possible next utterance" or anything mentioning or explaining the used technique.<|eot\_id|><|start\_header\_id|>user<|end\_header\_id|>

Input: \{history\}<|eot\_id|><|start\_header\_id|>assistant<|end\_header\_id|>
\end{tcolorbox}
\caption{Prompt used for the counselor agent in Psych8k.}
\label{fig:psych8k}
\end{figure*}

\begin{figure*}[t]
\centering
\begin{tcolorbox}[colback=gray!5!white, colframe=gray!70!black, title=CBT Agent Prompt]
<|start\_header\_id|>system<|end\_header\_id|>

You are a counselor specializing in CBT techniques. Your task is to use the provided client information, and dialogue to generate an appropriate CBT technique and a detailed counseling plan.<|eot\_id|><|start\_header\_id|>user<|end\_header\_id|>

Types of CBT Techniques:

Efficiency Evaluation, Pie Chart Technique, Alternative Perspective, Decatastrophizing, Pros and Cons Analysis, Evidence-Based Questioning, Reality Testing, Continuum Technique, Changing Rules to Wishes, Behavior Experiment, Problem-Solving Skills Training, Systematic Exposure

Client Information:

\{client\_information\}

Reason for seeking counseling:

\{reason\_counseling\}

Counseling Dialogue:

\{history\}

Choose an appropriate CBT technique and create a counseling plan based on that technique.<|eot\_id|><|start\_header\_id|>assistant<|end\_header\_id|>
\end{tcolorbox}
\caption{Prompt used for the CBT agent in CACTUS and \model.}
\label{fig:cbt}
\end{figure*}

\begin{figure*}[t]
\centering
\begin{tcolorbox}[colback=gray!5!white, colframe=gray!70!black, title=CACTUS Response Generation Agent Prompt]
<|start\_header\_id|>system<|end\_header\_id|>

You are playing the role of a counselor in a psychological counseling session. Your task is to use the provided client information and counseling planning to generate the next counselor utterance in the dialogue. The goal is to create a natural and engaging response that builds on the previous conversation and aligns with the counseling plan. Please be mindful to only generate the counselor response for a single turn, and do not include extra text like "here is the next counselor utterance" or "Here is a possible next utterance" or anything mentioning or explaining the used technique.<|eot\_id|><|start\_header\_id|>user<|end\_header\_id|>

Client Information:

\{client\_information\}

Reason for seeking counseling:

\{reason\_counseling\}

Counseling planning:

\{cbt\_plan\}

Counseling Dialogue:

\{history\}<|eot\_id|><|start\_header\_id|>assistant<|end\_header\_id|>
\end{tcolorbox}
\caption{Prompt used for the Response Generation agent in CACTUS.}
\label{fig:cactus}
\end{figure*}

\begin{figure*}[t]
\centering
\begin{tcolorbox}[colback=gray!5!white, colframe=gray!70!black, title=\model\ Reflection Agent Prompt]
<|start\_header\_id|>system<|end\_header\_id|>

You are playing the role of a counselor in a psychological counseling session specializing in reflections. Reflection is a technique used by the counselor to help a client gain insight into their thoughts, feelings, and behaviors by mirroring or paraphrasing what the client expresses, allowing the client to hear and evaluate their own statements more clearly. Your task is to use the provided client information to generate the next reflection-based counselor utterance in the dialogue. The goal is to create a natural and engaging response that builds on the previous conversation through reflection. Please be mindful to only generate the counselor response for a single turn and do not include extra text like "here is the next counselor utterance" or "Here is a possible next utterance" or anything mentioning the used technique. Please ensure that the utterances sound natural and ensure that your responses do not exactly repeat any of the counselor's previous utterances from the dialogue history.<|eot\_id|><|start\_header\_id|>user<|end\_header\_id|>

Client Information:

\{client\_information\}

Reason for seeking counseling:

\{reason\_counseling\}

Counseling Dialogue:

\{history\}<|eot\_id|><|start\_header\_id|>assistant<|end\_header\_id|>
\end{tcolorbox}
\caption{Prompt used for the Reflection agent in \model.}
\label{fig:reflection}
\end{figure*}

\begin{figure*}[t]
\centering
\begin{tcolorbox}[colback=gray!5!white, colframe=gray!70!black, title=\model\ Questioning Agent Prompt]
<|start\_header\_id|>system<|end\_header\_id|>

You are playing the role of a counselor in a psychological counseling session specializing in questioning. Questioning is a technique used by counselors to gain deeper understanding and insights on how the client feels regarding some previously mentioned events, how the client feels at present or understand how the client feels when asked to consider the situation from an alternative perspective. Your task is to use the provided client information to generate the next questioning-based counselor utterance in the dialogue. The goal is to create a natural and engaging response that builds on the previous conversation through questioning. Please be mindful to only generate the counselor response for a single turn and do not include extra text like "here is the next counselor utterance" or "Here is a possible next utterance" or anything mentioning the used technique. Please ensure that the utterances sound natural and ensure that your responses do not exactly repeat any of the counselor's previous utterances from the dialogue history.<|eot\_id|><|start\_header\_id|>user<|end\_header\_id|>

Client Information:

\{client\_information\}

Reason for seeking counseling:

\{reason\_counseling\}

Counseling Dialogue:

\{history\}<|eot\_id|><|start\_header\_id|>assistant<|end\_header\_id|>
\end{tcolorbox}
\caption{Prompt used for the Questioning agent in \model.}
\label{fig:questioning}
\end{figure*}

\begin{figure*}[t]
\centering
\begin{tcolorbox}[colback=gray!5!white, colframe=gray!70!black, title=\model\ Solutions Agent Prompt]
<|start\_header\_id|>system<|end\_header\_id|>

You are playing the role of a counselor in a psychological counseling session specializing in providing solutions to client issues. Counselors provide solutions grounded in psychological techniques that can be used by clients to improve their condition and deal with the mental health issues they are facing. Your task is to use the provided client information to generate the next solution-based counselor utterance in the dialogue for solving client's problems. The goal is to create a natural and engaging response that builds on the previous conversation through providing solutions. Please be mindful to only generate the counselor response for a single turn and do not include extra text like "here is the next counselor utterance" or "Here is a possible next utterance" or anything mentioning the used technique. Please ensure that the utterances sound natural and ensure that your responses do not exactly repeat any of the counselor's previous utterances from the dialogue history.<|eot\_id|><|start\_header\_id|>user<|end\_header\_id|>

Client Information:

\{client\_information\}

Reason for seeking counseling:

\{reason\_counseling\}

Counseling Dialogue:

\{history\}<|eot\_id|><|start\_header\_id|>assistant<|end\_header\_id|>
\end{tcolorbox}
\caption{Prompt used for the Solutions agent in \model.}
\label{fig:solutions}
\end{figure*}

\begin{figure*}[t]
\centering
\begin{tcolorbox}[colback=gray!5!white, colframe=gray!70!black, title=\model\ Normalizing Agent Prompt]
<|start\_header\_id|>system<|end\_header\_id|>

You are playing the role of a counselor in a psychological counseling session specializing in normalization. Normalization is a technique used by the counselor to acknowledge and validate the client's experience as normal or expectable, sympathize with their challenges, and provide reassurance to foster a supportive and encouraging therapeutic atmosphere. Your task is to use the provided client information to generate the next normalization-based counselor utterance in the dialogue. The goal is to create a natural and engaging response that builds on the previous conversation through normalization. Please be mindful to only generate the counselor response for a single turn and do not include extra text like "here is the next counselor utterance" or "Here is a possible next utterance" or anything mentioning the used technique. Please ensure that the utterances sound natural and ensure that your responses do not exactly repeat any of the counselor's previous utterances from the dialogue history.<|eot\_id|><|start\_header\_id|>user<|end\_header\_id|>

Client Information:

\{client\_information\}

Reason for seeking counseling:

\{reason\_counseling\}

Counseling Dialogue:

\{history\}<|eot\_id|><|start\_header\_id|>assistant<|end\_header\_id|>
\end{tcolorbox}
\caption{Prompt used for the Normalizing agent in \model.}
\label{fig:normalizing}
\end{figure*}

\begin{figure*}[t]
\centering
\begin{tcolorbox}[colback=gray!5!white, colframe=gray!70!black, title=\model\ Psycho-education Agent Prompt]
<|start\_header\_id|>system<|end\_header\_id|>

You are playing the role of a counselor in a psychological counseling session specializing in psycho-education. Psycho-education is a technique used by the counselor to provide therapeutically relevant information about psychological principles to the client to help them understand their issues and the logic behind the solutions. Your task is to use the provided client information to generate the next psycho-education-based counselor utterance in the dialogue. The goal is to create a natural and engaging response that builds on the previous conversation through psycho-education. Please be mindful to only generate the counselor response for a single turn and do not include extra text like "here is the next counselor utterance" or "Here is a possible next utterance" or anything mentioning the used technique. Please ensure that the utterances sound natural and ensure that your responses do not exactly repeat any of the counselor's previous utterances from the dialogue history.<|eot\_id|><|start\_header\_id|>user<|end\_header\_id|>

Client Information:

\{client\_information\}

Reason for seeking counseling:

\{reason\_counseling\}

Counseling Dialogue:

\{history\}<|eot\_id|><|start\_header\_id|>assistant<|end\_header\_id|>
\end{tcolorbox}
\caption{Prompt used for the Psycho-education agent in \model.}
\label{fig:psychoed}
\end{figure*}

\begin{figure*}[t]
\centering
\begin{tcolorbox}[colback=gray!5!white, colframe=gray!70!black, title=\model\ Technique Agent Prompt]
You are a counselor specializing in psychological techniques used in a therapy session. Your task is to use the provided conversation history and the suggested counseling plan to suggest the psychological technique or combination of techniques to be used by the therapist in the next dialogue turn. Remember that the counselor should properly explore and understand the client issues, normalize the issues and then provide solutions with psycho-education.

Counseling planning:

\{cbt\_plan\}

Counseling Dialogue:

\{history\}

Types of Psychological Techniques:

Reflection, Questioning, Providing solutions, Normalization, Psycho-education

Choose an appropriate technique or combination of technique for counselor response. ***Generate only the techniques from the list and do not generate possible responses***. The selected technique(s):
\end{tcolorbox}
\caption{Prompt used for the Technique agent in \model.}
\label{fig:technique}
\end{figure*}

\begin{figure*}[t]
\centering
\begin{tcolorbox}[colback=gray!5!white, colframe=gray!70!black, title=\model\ Response Generation Agent Prompt]
<|start\_header\_id|>system<|end\_header\_id|>

You are playing the role of a counselor in a psychological counseling session. Your task is to generate the counselor response in the dialogue by combining responses suggested by counselors specializing in reflections, questioning, providing solutions, normalizing and psycho-education based on the suggested techniques. The goal is to create a natural and engaging response that builds on the previous conversation. Please ensure that the response is empathetic and understanding of the client's issues and builds trust between the counselor and the client. Please be mindful to only generate the counselor response for a single turn, and do not include extra text like "here is the next counselor utterance" or "Here is a possible next utterance" or anything mentioning the used technique.<|eot\_id|><|start\_header\_id|>user<|end\_header\_id|>

Reflection response:

\{reflection\_utt\}

Questioning response:

\{question\_utt\}

Solution response:

\{solution\_utt\}

Normalizing response:

\{normalize\_utt\}

Psycho-education response:

\{psychoed\_utt\}

Suggested Technique(s):

\{technique\}<|eot\_id|><|start\_header\_id|>assistant<|end\_header\_id|>
\end{tcolorbox}
\caption{Prompt used for the Response Generation agent in \model.}
\label{fig:dialogue_gen}
\end{figure*}

\section{Client Simulation}
\label{sec:appendix-client}

The client agent is designed to simulate a realistic client in counseling sessions based on structured background information provided by the client intake form \cite{cactus}. The client intake form includes demographic details (e.g., name, occupation, age, family status) as well as the client's mental health concerns and reasons for seeking therapy. An example intake form is provided in Figure \ref{fig:intake}. To improve diversity and realism among simulated clients, we also simulated three different attitudes: positive, neutral, and negative \cite{cactus}. Each attitude is accompanied by detailed instructions, which are shown below:

\begin{itemize}
    \item \textbf{Positive:} Clients demonstrate a high level of engagement and cooperation with the therapeutic process. They should actively confirm their understanding of the counselor's instructions, ask for clarifications when needed, and willingly provide detailed information about their thoughts, feelings, and behaviors. These clients make reasonable requests for additional support or resources, and they extend the conversation by building on the counselor's suggestions with their own insights or experiences. They reformulate their thoughts in a constructive manner, reflecting on their progress and expressing a hopeful outlook towards the therapeutic outcomes. Overall, their demeanor is open, appreciative, and proactive in seeking improvement.
    \item \textbf{Neutral:} Clients display a mix of both positive and negative characteristics. They might show compliance and willingness to follow instructions at times, but also exhibit moments of defensiveness or skepticism. These clients may provide useful information and participate actively in some discussions, while in other instances, they might shift topics or show disconnection. Their feedback can vary, with periods of constructive engagement interspersed with sarcastic remarks or expressions of self-doubt. This blend of reactions indicates a fluctuating commitment to therapy, with the client balancing between optimism for change and resistance to the therapeutic process.
    \item \textbf{Negative:} Clients displaying negative reactions may struggle with the therapeutic process, often showing signs of resistance or defensiveness. They might express confusion about the counselor's guidance, indicating difficulty in understanding or accepting the proposed strategies. These clients could defend their current behaviors or viewpoints, potentially shifting topics to avoid addressing the core issues. There might be a noticeable disconnection in focus, where the client's attention drifts away from the session's goals. Sarcastic responses and self-criticism or hopelessness are common, reflecting a pessimistic attitude towards their ability to change or benefit from therapy. These behaviors suggest an underlying frustration or lack of trust in the counseling process.
\end{itemize}

The client agent uses the intake form, attitude, the corresponding attitude instructions and the current dialogue context to generate client responses. The client agent is also instructed to terminate the session if they feel their primary concern has been resolved or no further counseling is needed. For uniformity, we keep the client agent common for Psych8k, CACTUS, and \model. The prompt used for the client agent is shown in Figure \ref{fig:client}. 

Overall, the counseling session generation requires client profiles, attitude, attitude instructions and initial greeting dialogues as generation seeds (intake forms and client attitudes). We obtain these generation seeds from the CACTUS dataset \citep{cactus}.

\begin{figure*}[t]
\centering
\begin{tcolorbox}[colback=gray!5!white, colframe=gray!70!black, title=Client Intake Form]
\textbf{Name}: Laura Saunders 

\textbf{Age}: 45 

\textbf{Gender}: female 

\textbf{Occupation}: Office Job 

\textbf{Education}: College Graduate 

\textbf{Marital Status}: Single 

\textbf{Family Details}: Lives alone 

\textbf{Presenting Problem}: I am struggling with the thought that I can't run far, despite enjoying running as a hobby. This negative belief is affecting my motivation and self-esteem. The issue started a few months ago when I started setting running goals for myself. I believe the stress of not meeting my running goals has contributed to this problem. The problem has progressed to a point where I feel demotivated to continue running and doubt my abilities. I experience this negative thought pattern every time I lace up my running shoes and head out for a run. I have tried to improve my running technique and read motivational articles, but it hasn't helped much.

\textbf{Reason for Seeking Counseling}: I decided to seek counseling because this negative belief is hindering my enjoyment of running and affecting my overall mood. 

\textbf{Past History (including medical history)}: I have not experienced similar problems before. I have not received treatment or counseling for psychological issues. I do not have any significant physical illnesses. 

\textbf{Academic/occupational functioning level}: My job performance has not been impacted by this issue. Interpersonal relationships: I have not experienced any strain in my relationships due to this issue. Daily life: My self-esteem and motivation levels have been affected by this negative belief. 

\textbf{Social Support System}: I have a close friend who is supportive, but I have not discussed this specific issue with them.
\end{tcolorbox}
\caption{An example of the Client Intake Form.}
\label{fig:intake}
\end{figure*}

\begin{figure*}[t]
\centering
\begin{tcolorbox}[colback=gray!5!white, colframe=gray!70!black, title=Client Agent Prompt]
<|start\_header\_id|>system<|end\_header\_id|>

You are playing the role of a client in a psychological counseling session. Your task is to generate only one suitable response based on the following the counseling dialogue history.

\#\# Guidelines for the client's utterance:

1. Engage authentically with the counselor's inquiries, reflecting the complexity of emotions and reactions typical in counseling sessions.

2. Start the client's utterance with 'Client:'. Ensure that the utterance follows the exact format and does not contain any control characters.

3. The client should maintain the following attitude.

If you feel that the counseling session has completely ended and meets the end condition, you should include '[/END]' with your utterance.

***End Conditions:***

- The client feels that their negative thoughts have been resolved.

- The client feels that no further counseling is needed

Please be mindful of these conditions and ensure that ***the session should not end prematurely; it must last at least 20 turns.***. Generate only the client's utterance for a single turn and please ensure that your responses do not repeat the client's previous utterances. Do not generate the counselor's part of the dialogue. <|eot\_id|><|start\_header\_id|>user<|end\_header\_id|>

Client Persona and Negative Thoughts:

\{intake\_form\}

Client's Attitude Towards Counseling:

\{attitude\}

Counseling Dialogue History:

\{history\}<|eot\_id|><|start\_header\_id|>assistant<|end\_header\_id|>
\end{tcolorbox}
\caption{Prompt used for Client agent in Psych8k, CACTUS and \model.}
\label{fig:client}
\end{figure*}

\section{Diversity Evaluation}
\label{sec:appendix-div}

To assess the diversity of generated counseling sessions, we employ the Distinct-$n$ metrics ($n \in \{1, 2, 3\}$) \cite{distinct_n}, which compute the ratio of unique $n$-grams to the total number of $n$-grams in a corpus. Higher values indicate greater lexical diversity. For this computation, we concatenate all dialogue turns from both the counselor and the client within a generated session, remove punctuation, and tokenize the text using the Llama-3 tokenizer.

While Distinct-$n$ is widely used, it exhibits a known bias towards shorter sequences, assigning lower scores to longer sequences. To mitigate this, we also use the Expectation-Adjusted Distinct (EAD) score \cite{liu-etal-2022-rethinking}, which normalizes for sequence length and has been shown to correlate more strongly with human judgments of diversity. EAD provides a more robust measure of lexical variation by adjusting the expected distinctness relative to the length of the sequence. The following equation is used to calculate the EAD score:

\begin{equation}
    EAD = \frac{N}{V[1-(\frac{V-1}{V})^C]}
\end{equation}

where $N$ is the number of distinct tokens, $C$ is the total number of tokens and $V$ is the vocabulary size.

\section{Quality Evaluation}
\label{sec:appendix-quality}

We provide further details on the psychological assessment scales employed in our data quality evaluation: the Cognitive Therapy Rating Scale (CTRS), the Working Alliance Inventory (WAI), and the Positive and Negative Affect Schedule (PANAS). The mean scores and standard deviations for these scales, computed over counseling sessions generated by different synthetic counseling session generation methods, are reported in Table \ref{tab:ctrs-panas-wai-sd}.

\subsection{CTRS}

To assess the quality of counselor responses, we adopt the Cognitive Therapy Rating Scale (CTRS) \cite{ctrs,young1980ctrs}, a widely used psychological scale for evaluating both general and Cognitive Behavioral Therapy (CBT)-specific counseling skills. We follow the same CTRS evaluation protocol as CACTUS~\cite{cactus}. 

CTRS organizes its assessment into two broad categories. The first covers \textit{general counseling skills}, evaluated across three dimensions. \textbf{Understanding} captures the degree to which the counselor accurately identifies and interprets what the client is experiencing. \textbf{Interpersonal Effectiveness} reflects the counselor's capacity to build and sustain a constructive therapeutic relationship. \textbf{Collaboration} measures how actively the counselor involves the client in shaping treatment goals and decisions throughout the session.

The second category covers \textit{CBT-specific skills}, also evaluated along three dimensions. \textbf{Guided Discovery} assesses how skillfully the counselor leads the client toward new insights through structured questioning and reflection. \textbf{Focus} reflects the counselor's precision in identifying and consistently targeting the cognitions or behaviors most in need of change. \textbf{Strategy} captures whether the counselor's overall therapeutic approach is coherent, well-suited to the client's needs, and directed toward meaningful cognitive or behavioral shifts.
 
Each of these six dimensions is rated on a scale from $0$ to $6$, where higher scores indicate stronger demonstration of the corresponding skill. Ratings are obtained using an LLM-as-a-judge approach, leveraging GPT-4o to score each item based on the generated counseling dialogue. The prompt used for this evaluation is shown in Figure \ref{fig:ctrs}.

\begin{figure*}[t]
\centering
\begin{tcolorbox}[colback=gray!5!white, colframe=gray!70!black, title=CTRS Evaluation Prompt]
I want you to act as an evaluator. You will be provided with a transcript of a counseling session between a therapist and a client. Your task is to assess the therapist based on the given criteria. If you believe the therapist falls between two of the descriptors, select the intervening odd number (1, 3, 5). For example, if the therapist set a very good agenda but did not establish priorities, assign a rating of 5 rather than 4.

Please follow these steps:

1.	Read the counseling session transcript carefully.

2.	Review the evaluation questions and criteria provided below.

3.	Assign a score based on the criteria, grading very strictly and uptight. If there is any deficiency, no matter how minor, assign a score of 4 or lower.

4.	Output the score and the explanation, separated by a comma. Do not add any prefix.

Counseling conversation:

\{conversation\}

Evaluation Question:

\{question\}

Criteria:

\{criteria\}
\end{tcolorbox}
\caption{Prompt used for evaluating the generated counseling sessions on CTRS.}
\label{fig:ctrs}
\end{figure*}

\subsection{WAI}

To evaluate the strength of the therapeutic alliance between the counselor and client, we adopt the Working Alliance Inventory (WAI) \cite{wai-comp}, following the setup described in \citet{llm-roleplay}. WAI is a psychological measurement tool consisting of 12 items, categorized into three groups of therapeutic alliance: Goal, Task, and Bond. Rather than enumerating each item individually, we describe each dimension and its
constituent items together. 

The \textbf{Task} dimension (WAI-1, 2, 8, 12) concerns the degree to which the counselor and client share a mutual understanding of and commitment to the therapeutic work: whether they align on the concrete steps being taken, view the current activities as worthwhile, agree on which issues are most pressing, and whether the client regards the overall approach as fitting their situation.
 
The \textbf{Goal} dimension (WAI-4, 6, 10, 11) captures the extent to which counselor and client are oriented toward the same counseling objectives. This includes shared clarity about the purpose of the sessions, joint commitment to agreed-upon goals, a common understanding of what changes would benefit the client, and the absence of conflicting views about the nature of the client's core difficulties. WAI-4 and WAI-10 are framed negatively, expressing doubt or disagreement and their scores are therefore inverted before aggregation.
 
The \textbf{Bond} dimension (WAI-3, 5, 7, 9) reflects the interpersonal quality of the therapeutic relationship: whether the client and counselor experience mutual liking and respect, whether the client feels genuinely valued as a person, whether the client trusts in the counselor's competence, and whether a sense of mutual trust characterizes the relationship overall.

Each item is scored on a scale of $1$ to $7$ using GPT-4o in a LLM-as-a-judge setup. For all items except WAI-4 and WAI-10, a higher score indicates a stronger therapeutic alliance. WAI-4 and WAI-10 are negatively framed and therefore inverted via $\tilde{s}_i = 8 - s_i$ before aggregation; all other items retain their raw scores ($\tilde{s}_i = s_i$). Table \ref{tab:wai-groupings} summarizes the item-to-dimension mapping. Letting $\mathcal{D}_d$ denote the set of four item indices belonging to dimension $d$, the dimension score is:
 
\begin{equation}
S_d = \frac{1}{|\mathcal{D}_d|} \sum_{i \in \mathcal{D}_d} \tilde{s}_i, \quad d \in \{\text{Task},\, \text{Goal},\, \text{Bond}\}
\end{equation}
 
\begin{table}[h]
\centering
\small
\begin{tabular}{lll}
\hline
\textbf{Dimension} & \textbf{Items (raw)} & \textbf{Items (inverted)} \\
\hline
Task & WAI-1, 2, 8, 12 & --- \\
Goal & WAI-6, 11       & WAI-4, 10 \\
Bond & WAI-3, 5, 7, 9  & --- \\
\hline
\end{tabular}
\caption{WAI item groupings by dimension. Items in the ``inverted'' column are transformed as $\tilde{s}_i = 8 - s_i$ prior to averaging.}
\label{tab:wai-groupings}
\end{table}

The prompt used to score WAI items using GPT-4o in a LLM-as-a-judge setup is shown in Figure \ref{fig:wai}.

\begin{figure*}[t]
\centering
\begin{tcolorbox}[colback=gray!5!white, colframe=gray!70!black, title=WAI Evaluation Prompt]
The following is a psychological counseling session between a counselor and a client. As a third party, you should read the conversation and guidelines carefully and then score the following question from 1 to 7.

Please follow these steps:

1.	Read the counseling session transcript carefully.

2.	Review the evaluation questions and criteria provided below.

3.	Assign a score based on the criteria, grading very strictly.

4.	Output the score (***only the numerical***) and the explanation, separated by a comma. ***Do not add any prefix.***

Counseling conversation:

\{conversation\}

Question: \{question\}

Criteria:

\{criteria\}
\end{tcolorbox}
\caption{Prompt used for evaluating the generated counseling sessions on WAI.}
\label{fig:wai}
\end{figure*}

\subsection{PANAS}
\label{appendix-panas}
Positive and Negative Affect Schedule (PANAS) \cite{panas} is a self-report questionnaire used to assess the positive and negative emotions of a person at a certain time or over a period of time. Here, we use PANAS to measure the changes in positive and negative emotions of the client from before counseling to after the counseling session. PANAS comprises $10$ items for positive emotions and $10$ items for negative emotions, with each item rated on a $5$-point Likert scale ($1$–$5$). The list of emotions rated are as follows:

\begin{itemize}
    \item \textbf{Positive Emotions:} Interested, Excited, Strong, Enthusiastic, Proud, Alert, Inspired, Determined, Attentive, Active.
    \item \textbf{Negative Emotions:} Distressed, Upset, Guilty, Scared, Hostile, Irritable, Ashamed, Nervous, Jittery, Afraid.
\end{itemize}

In our setup, we use the LLM-as-a-judge approach with GPT-4o to rate the emotions. For evaluation before the counseling session, the judge model scores each emotion item based solely on the client intake form, which reflects the client’s emotional baseline. For evaluation after the counseling session, the judge model considers both the intake form and the generated counseling session to assess the client's updated emotional scores. The final positive and negative affect scores before and after the counseling sessions are computed by averaging the scores of the respective $10$ items. Finally, the changes in the average positive and negative affect score between before and after the counseling session are reported. The prompts used for PANAS scoring before and after the counseling session are shown in Figure \ref{fig:panas_before} and Figure \ref{fig:panas_after}, respectively.

\begin{figure*}[t]
\centering
\begin{tcolorbox}[colback=gray!5!white, colframe=gray!70!black, title=PANAS Evaluation Prompt before Counseling]
A person with the characteristics listed in the intake form received counseling. Based on the text provided, evaluate the intensity of each of the following feelings the person might have experienced: Interested, Excited, Strong, Enthusiastic, Proud, Alert, Inspired, Determined, Attentive, Active, Distressed, Upset, Guilty, Scared, Hostile, Irritable, Ashamed, Nervous, Jittery, Afraid.

For each feeling, generate a score from 1 to 5 using the following scale:

1 - Very slightly or not at all

2 - A little

3 - Moderately

4 - Quite a bit

5 - Extremely

Additionally, provide a brief explanation for each score. ***Separate feeling, explanation, score by comma. The score should only be the numeric value. Do not add explanations after the score. Do not add any prefix.***

Here is the text:

\{intake\_form\}
\end{tcolorbox}
\caption{Prompt used for evaluating on PANAS before counseling.}
\label{fig:panas_before}
\end{figure*}

\begin{figure*}[t]
\centering
\begin{tcolorbox}[colback=gray!5!white, colframe=gray!70!black, title=PANAS Evaluation Prompt after Counseling]
A person with the characteristics listed in the intake form received counseling. The following counseling session is a conversation between the client and the counselor. After reviewing the conversation, evaluate the intensity of each of the following feelings the person might have experienced once the counseling session is complete: Interested, Excited, Strong, Enthusiastic, Proud, Alert, Inspired, Determined, Attentive, Active, Distressed, Upset, Guilty, Scared, Hostile, Irritable, Ashamed, Nervous, Jittery, Afraid.

For each feeling, generate a score from 1 to 5 using the following scale:

1 - Very slightly or not at all

2 - A little

3 - Moderately

4 - Quite a bit

5 - Extremely

Additionally, provide a brief explanation for each score. ***Separate feeling, explanation, score by comma. The score should only be the numeric value. Do not add explanations after the score. Do not add any prefix.***

Here is the text:

\{intake\_form\}

Here is the counseling session:

\{conversation\}
\end{tcolorbox}
\caption{Prompt used for evaluating on PANAS after counseling.}
\label{fig:panas_after}
\end{figure*}

\section{Counseling Agent Fine-tuning}
\label{sec:appendix-counselor-agent-fine-tuning}

To assess the downstream effectiveness of the generated synthetic counseling sessions, we fine-tune Llama3-8B-Instruct model \cite{meta2024introducing} on synthetic sessions generated by different methods: Psych8k \cite{chatcounselor}, CACTUS \cite{cactus}, and \model, resulting in Llama-Psych8k, Llama-CACTUS, and Llama-\model\ respectively. For fine-tuning, we begin by splitting the generation seeds into training, validation, and test sets. The generation seeds \cite{cactus} include the client intake form, the corresponding client attitude (positive, neutral, or negative), attitude-specific instructions, and the initial greeting dialogue turns between the client and counselor. To prevent data leakage and ensure fair evaluation, we split the seeds at the client level, ensuring that intake forms associated with a given client appear in only one of the train, validation, or test splits. This avoids fine-tuning and evaluating on sessions that differ only in client attitude. We take the generation seeds from the CACTUS dataset \cite{cactus} which contains $150$ unique client profiles with 3 different attitudes, resulting in $450$ generation seeds. We split them into a training set containing $90$ clients (i.e., 270 generation seeds), a validation set with $10$ clients (30 generation seeds), and a test set with the remaining $50$ clients (150 generation seeds). For fine-tuning, we extract (dialogue history, counselor response) pairs from the synthetic sessions generated using the generation seeds in the training set. Since each session contains $20$ counselor dialogue turns, this results in $5400$ (dialogue history, counselor response) pairs. A similar approach is used for the validation set, resulting in $600$ (dialogue history, counselor response) pairs. Once fine-tuned, the fine-tuned model is used to simulate the counselor agent and generate synthetic counseling sessions using the generation seeds in the test split. The client agent is kept the same and uses a non-fine-tuned Llama3-8B-Instruct. These generated counseling sessions are then evaluated using CTRS, WAI, and PANAS. The prompts used for fine-tuning Llama3-8B-Instruct model using the Psych8k-generated data, CACTUS-generated data, and \model-generated data are shown in Figure \ref{fig:qlora-psych8k}, Figure \ref{fig:qlora-cactus}, and Figure \ref{fig:qlora-mascs}, respectively. The same prompts without the response part in the assistant section are also used for generating counseling sessions using the fine-tuned counseling models with the generation seeds from the test set. The mean scores and standard deviations for CTRS, PANAS, and WAI, computed over the counseling sessions generated by the fine-tuned models, are shown in Table \ref{tab:ctrs-panas-wai-model-sd}.

\begin{figure*}[t]
\centering
\begin{tcolorbox}[colback=gray!5!white, colframe=gray!70!black, title=Psych8k QLora Prompt]
<|start\_header\_id|>system<|end\_header\_id|>

If you are a counselor, please answer the questions based on the description of the patient. Please be mindful to only generate the counselor response for a single turn, and do not include extra text like "here is the next counselor utterance" or "Here is a possible next utterance" or anything mentioning or explaining the used technique.<|eot\_id|><|start\_header\_id|>user<|end\_header\_id|>

Input: \{history\}<|eot\_id|><|start\_header\_id|>assistant<|end\_header\_id|>

\{response\}<|eot\_id|>
\end{tcolorbox}
\caption{Prompt used for QLora fine-tuning with Psych8k-generated data.}
\label{fig:qlora-psych8k}
\end{figure*}

\begin{figure*}[t]
\centering
\begin{tcolorbox}[colback=gray!5!white, colframe=gray!70!black, title=CACTUS QLora Prompt]
<|start\_header\_id|>system<|end\_header\_id|>

You are playing the role of a counselor in a psychological counseling session. Your task is to use the provided client information to generate the next counselor utterance in the dialogue. The goal is to create a natural and engaging response that builds on the previous conversation. Please be mindful to only generate the counselor response for a single turn, and do not include extra text like "here is the next counselor utterance" or "Here is a possible next utterance" or anything mentioning or explaining the used technique.<|eot\_id|><|start\_header\_id|>user<|end\_header\_id|>

Client Information:

\{client\_information\}

Reason for seeking counseling:

\{reason\_counseling\}

Counseling Dialogue:

\{history\}<|eot\_id|><|start\_header\_id|>assistant<|end\_header\_id|>

\{response\}<|eot\_id|>
\end{tcolorbox}
\caption{Prompt used for QLora fine-tuning with CACTUS-generated data.}
\label{fig:qlora-cactus}
\end{figure*}

\begin{figure*}[t]
\centering
\begin{tcolorbox}[colback=gray!5!white, colframe=gray!70!black, title=\model\ QLora Prompt]
<|start\_header\_id|>system<|end\_header\_id|>

You are playing the role of a counselor in a psychological counseling session. Your task is to use the provided client information to generate the next counselor response in the dialogue by combining psychological techniques like reflections, questioning, providing solutions, normalizing and psycho-education. The goal is to create a natural and engaging response that builds on the previous conversation. Please ensure that the response is empathetic and understanding of the client's issues and builds trust between the counselor and the client. Please be mindful to only generate the counselor response for a single turn, and do not include extra text like "here is the next counselor utterance" or "Here is a possible next utterance" or anything mentioning the used technique.<|eot\_id|><|start\_header\_id|>user<|end\_header\_id|>

Client Information:

\{client\_information\}

Reason for seeking counseling:

\{reason\_counseling\}

Counseling Dialogue:

\{history\}<|eot\_id|><|start\_header\_id|>assistant<|end\_header\_id|>

\{response\}<|eot\_id|>
\end{tcolorbox}
\caption{Prompt used for QLora fine-tuning with \model-generated data.}
\label{fig:qlora-mascs}
\end{figure*}

\section{CounselingBench}
\label{sec:appendix-counselingbench}

To provide a complementary, objective assessment of the utility of the fine-tuned models, we evaluate them on CounselingBench \citep{counselingbench}. CounselingBench comprises 1,621 multiple-choice questions aligned with the National Clinical Mental Health Counseling Examination (NCMHCE) content outline, each accompanied by patient demographic and background information. This MCQ benchmark measures the knowledge of the model. For this evaluation, we fine-tune a Llama3-8B-Instruct model on CACTUS, Psych8k, and \model-generated sessions, following the procedure described in Appendix \ref{sec:appendix-counselor-agent-fine-tuning}. However, unlike our main experiments, we use all generated sessions for fine-tuning, as the evaluation is conducted on an independent benchmark. 

The fine-tuned models are then evaluated under three prompting techniques of increasing guidance. In the \textbf{Zero Shot (ZS)} prompting, models respond to each question without any worked examples, relying solely on knowledge acquired during fine-tuning (prompt in Figure \ref{fig:counselingbench-zs-prompt}). In the \textbf{Few Shot (FS)} prompting, three example question-answer pairs are prepended to each query as in-context demonstrations (prompt in Figure \ref{fig:counselingbench-fs-prompt}). In the \textbf{Few Shot Chain-of-Thought (FS-CoT)} prompting, those same three demonstrations are augmented with expert-annotated step-by-step reasoning, and models are explicitly instructed to produce their own reasoning trace before arriving at an answer (prompt in Figure \ref{fig:counselingbench-fscot-prompt}).

Following the hyperparameter settings recommended in the benchmark, we generate model responses using a temperature of $T=0$ and a top-$p$ value of 0.9. Because the multiple-choice format of CounselingBench constitutes a multi-class classification task, we report F1 scores to assess the ability of the models to select the correct answers. In addition, we evaluate the chain-of-thought (CoT) reasonings produced under the FS-CoT prompting using both reference-based and reference-free metrics. For reference-based evaluation, we compute cosine similarity, BERTScore \citep{bertscore}, ROUGE-1, and ROUGE-L \citep{lin-2004-rouge} between the generated reasonings and the expert-annotated explanations provided in the benchmark. The cosine similarity is calculated using Sentence Transformer \citep{sentence-transformer} embeddings. For reference-free evaluation, we use Roscoe \citep{roscoe}, using the metrics specified in the benchmark: Faithfulness, Step Informativeness, Chain Informativeness, Missing Step, Alignment, Repetition, Grammar, and Self-Consistency.

Table \ref{tab:counselingbench_f1} reports the F1 scores of the fine-tuned models under the ZS, FS, and FS-CoT prompting techniques. Across all three prompting techniques, the models achieve comparable performance, with Llama-\model\ showing a slight overall advantage. However, the benefits of incorporating \model-generated synthetic data become substantially more apparent when evaluating the quality of the reasoning chains produced under FS-CoT prompting.

Table \ref{tab:counselingbench_roscoe} presents both reference-based and reference-free evaluations of the generated reasoning chains. According to the reference-based metrics, the reasoning produced by Llama-\model\ aligns more closely with expert annotated reasoning, as reflected in notably higher cosine similarity and BERTScore. While ROUGE-1 and ROUGE-L scores remain similar across models, these metrics rely on n-gram overlap rather than semantic similarity, making them less informative in this context.

Turning to the reference-free ROSCOE metrics, we observe that all models perform similarly on most dimensions. The exception is self-consistency, where Llama-\model\ demonstrates a clear advantage over the baselines showing it can produce more coherent and consistent reasoning chains. Taken together, these findings highlight the effectiveness of \model-generated synthetic data in enhancing the reasoning quality of fine-tuned models beyond what is achievable with baseline datasets.

\begin{table}[t]
\centering
\begin{tabular}{@{}lccc@{}}
\toprule
\textbf{Model} & \textbf{ZS} & \textbf{FS} & \textbf{FS-CoT} \\ 
\midrule
Llama-Psych8k          & 0.605       & 0.619       &0.576           \\
Llama-CACTUS     & 0.608       & \textbf{0.621}       & 0.574           \\
Llama-\model     & \textbf{0.616}      & 0.616       & \textbf{0.582}           \\
\bottomrule
\end{tabular}
\caption{F1 Scores on CounselingBench for different prompting techniques used with fine-tuned Llama3-8B-Instruct models.}
\label{tab:counselingbench_f1}
\end{table}

\begin{table*}[t]
\centering
\resizebox{1.9\columnwidth}{!}{
\begin{tabular}{@{}lcccccccccccc@{}}
\toprule
\textbf{Model} & \textbf{cosSim ($\uparrow$)} & \textbf{BERT ($\uparrow$)} & \textbf{$R_{L}$ ($\uparrow$)} & \textbf{$R_{1}$ ($\uparrow$)} & \textbf{faith ($\uparrow$)} & \textbf{$info_{stp}$ ($\uparrow$)} & \textbf{$info_{chn}$ ($\uparrow$)} & \textbf{mis. ($\uparrow$)} & \textbf{al. ($\uparrow$)} & \textbf{rep. ($\downarrow$)} & \textbf{gmr. ($\uparrow$)} & \textbf{cons. ($\uparrow$)} \\ 
\midrule
Llama-Psych8k          & 0.663 & 0.157 & 0.177 & 0.319 & \textbf{0.893} & \textbf{0.876} & \textbf{0.883} & \textbf{0.801} & 0.894 & \textbf{0.035} & \textbf{0.978} & 0.602 \\
Llama-CACTUS     & 0.657 & 0.168 & \textbf{0.181} & 0.325 & 0.885 & 0.867 & 0.876 & 0.793 & 0.890 & 0.054 & 0.969 & 0.648 \\
Llama-\model     & \textbf{0.671} & \textbf{0.180} & \textbf{0.181} & \textbf{0.329} & 0.892 & 0.872 & 0.879 & 0.791 & \textbf{0.895} & 0.045 & 0.975 & \textbf{0.750} \\
\bottomrule
\end{tabular}
}
\caption{Performance of reasoning chains generated in FS-CoT prompting of the fine-tuned Llama3-8B-Instruct models on CounselingBench. The reasoning chains are evaluated on reference-based and reference-free metrics. Here cosSim (Cosine Similarity), BERT (BERTScore), $R_{L}$ (ROUGE-L), $R_{1}$ (ROUGE-1), faith (Faithfulness), $info_{stp}$ (Informativeness Step), $info_{chn}$ (Informativeness Chain), mis. (Missing step), al. (Alignment), rep. (Repetition), gmr. (Grammar), cons. (Self Consistency).}
\label{tab:counselingbench_roscoe}
\end{table*}

\begin{figure*}[t]
\centering
\begin{tcolorbox}[colback=gray!5!white, colframe=gray!70!black, title=CounselingBench ZS Prompt]
<|start\_header\_id|>system<|end\_header\_id|>

You are a helpful, respectful, honest, and knowledgeable student studying to become a licensed therapist. You must answer a series of multiple-choice questions provided by the user from a US mental health counselor licensing exam. Based on the question text and the context provided, ***you must answer with either "A", "B", "C" or "D"***.<|eot\_id|><|start\_header\_id|>user<|end\_header\_id|>

**Question**:
The patient demographic is as follows: \{patient\_demographics\}

\{presenting\_problem\}

\{mental\_status\_examination\}

\{other\_contexts\}

Given the context above and your expert-level knowledge of mental health counseling, please answer the following question: \{question\}

\{candidate\_answers\}<|eot\_id|><|start\_header\_id|>assistant<|end\_header\_id|>

Correct Answer is:
\end{tcolorbox}
\caption{Prompt used to generate model responses to questions in CounselingBench using Zero-Shot (ZS) prompting.}
\label{fig:counselingbench-zs-prompt}
\end{figure*}

\begin{figure*}[t]
\centering
\begin{tcolorbox}[colback=gray!5!white, colframe=gray!70!black, title=CounselingBench FS Prompt]
<|start\_header\_id|>system<|end\_header\_id|>

You are a helpful, respectful, honest, and knowledgeable student studying to become a licensed therapist. You must answer a series of multiple-choice questions provided by the user from a US mental health counselor licensing exam. Based on the question text and the context provided, ***you must answer with either "A", "B", "C" or "D"***.<|eot\_id|><|start\_header\_id|>user<|end\_header\_id|>

**Question**:

The patient demographic is as follows: \{patient\_demographics\_1\}

\{presenting\_problem\_1\}

\{mental\_status\_examination\_1\}

\{other\_contexts\_1\}

Given your expert-level knowledge of mental health counseling, please answer the following question:

\{question\_1\}

\{candidate\_answers\_1\}<|eot\_id|><|start\_header\_id|>assistant<|end\_header\_id|>

Correct Answer is (\{correct\_answer\_1\})<|eot\_id|><|start\_header\_id|>user<|end\_header\_id|>

**Question**:

The patient demographic is as follows: \{patient\_demographics\_2\}

\{presenting\_problem\_2\}

\{mental\_status\_examination\_2\}

\{other\_contexts\_2\}

Given your expert-level knowledge of mental health counseling, please answer the following question:

\{question\_2\}

\{candidate\_answers\_2\}<|eot\_id|><|start\_header\_id|>assistant<|end\_header\_id|>

Correct Answer is (\{correct\_answer\_2\})<|eot\_id|><|start\_header\_id|>user<|end\_header\_id|>

**Question**:

The patient demographic is as follows: \{patient\_demographics\_3\}

\{presenting\_problem\_3\}

\{mental\_status\_examination\_3\}

\{other\_contexts\_3\}

Given your expert-level knowledge of mental health counseling, please answer the following question:

\{question\_3\}

\{candidate\_answers\_3\}<|eot\_id|><|start\_header\_id|>assistant<|end\_header\_id|>

Correct Answer is (\{correct\_answer\_3\})<|eot\_id|><|start\_header\_id|>user<|end\_header\_id|>

**Question**:

The patient demographic is as follows: \{patient\_demographics\}

\{presenting\_problem\}

\{mental\_status\_examination\}

\{other\_contexts\}

Given your expert-level knowledge of mental health counseling, please answer the following question:

\{question\}

\{candidate\_answers\}<|eot\_id|><|start\_header\_id|>assistant<|end\_header\_id|>

Correct Answer is (
\end{tcolorbox}
\caption{Prompt used to generate model responses to questions in CounselingBench using Few-Shot (FS) prompting.}
\label{fig:counselingbench-fs-prompt}
\end{figure*}

\begin{figure*}[t]
\centering
\begin{tcolorbox}[colback=gray!5!white, colframe=gray!70!black, title=CounselingBench FS-CoT Prompt,fontupper=\small]
<|start\_header\_id|>system<|end\_header\_id|>

You are a helpful, respectful, honest, and knowledgeable student studying to become a licensed therapist. You must answer a series of multiple-choice questions provided by the user from a US mental health counselor licensing exam. Based on the question text and the context provided, ***you must answer with either "A", "B", "C" or "D"***.<|eot\_id|><|start\_header\_id|>user<|end\_header\_id|>

**Question**:

The patient demographic is as follows: \{patient\_demographics\_1\}

\{presenting\_problem\_1\}

\{mental\_status\_examination\_1\}

\{other\_contexts\_1\}

Given your expert-level knowledge of mental health counseling, please answer the following question by carefully and thoroughly reason step-by-step, leveraging relevant facts from the question context and expert-level counseling knowledge, the clearly indicate your answer with "Therefore, the correct answer is (A)", "Therefore, the correct answer is (B)", "Therefore, the correct answer is (C)" or "Therefore, the correct answer is (D)" at the end of your answer:

\{question\_1\}

\{candidate\_answers\_1\}<|eot\_id|><|start\_header\_id|>assistant<|end\_header\_id|>

\{expert\_written\_reasoning\_1\}. Therefore, the correct answer is (\{correct\_answer\_1\}). <|eot\_id|><|start\_header\_id|>user<|end\_header\_id|>

**Question**:

The patient demographic is as follows: \{patient\_demographics\_2\}

\{presenting\_problem\_2\}

\{mental\_status\_examination\_2\}

\{other\_contexts\_2\}

Given your expert-level knowledge of mental health counseling, please answer the following question by carefully and thoroughly reason step-by-step, leveraging relevant facts from the question context and expert-level counseling knowledge, the clearly indicate your answer with "Therefore, the correct answer is (A)", "Therefore, the correct answer is (B)", "Therefore, the correct answer is (C)" or "Therefore, the correct answer is (D)" at the end of your answer:

\{question\_2\}

\{candidate\_answers\_2\}<|eot\_id|><|start\_header\_id|>assistant<|end\_header\_id|>

\{expert\_written\_reasoning\_2\}. Therefore, the correct answer is (\{correct\_answer\_2\}). <|eot\_id|><|start\_header\_id|>user<|end\_header\_id|>

**Question**:

The patient demographic is as follows: \{patient\_demographics\_3\}

\{presenting\_problem\_3\}

\{mental\_status\_examination\_3\}

\{other\_contexts\_3\}

Given your expert-level knowledge of mental health counseling, please answer the following question by carefully and thoroughly reason step-by-step, leveraging relevant facts from the question context and expert-level counseling knowledge, the clearly indicate your answer with "Therefore, the correct answer is (A)", "Therefore, the correct answer is (B)", "Therefore, the correct answer is (C)" or "Therefore, the correct answer is (D)" at the end of your answer:

\{question\_3\}

\{candidate\_answers\_3\}<|eot\_id|><|start\_header\_id|>assistant<|end\_header\_id|>

\{expert\_written\_reasoning\_3\}. Therefore, the correct answer is (\{correct\_answer\_3\}). <|eot\_id|><|start\_header\_id|>user<|end\_header\_id|>

**Question**:

The patient demographic is as follows: \{patient\_demographics\}

\{presenting\_problem\}

\{mental\_status\_examination\}

\{other\_contexts\}

Given your expert-level knowledge of mental health counseling, please answer the following question by carefully and thoroughly reason step-by-step, leveraging relevant facts from the question context and expert-level counseling knowledge, the clearly indicate your answer with "Therefore, the correct answer is (A)", "Therefore, the correct answer is (B)", "Therefore, the correct answer is (C)" or "Therefore, the correct answer is (D)" at the end of your answer:

\{question\}

\{candidate\_answers\}<|eot\_id|><|start\_header\_id|>assistant<|end\_header\_id|>
\end{tcolorbox}
\caption{Prompt used to generate model responses to questions in CounselingBench using Few-Shot Chain-of-Thought (FS-CoT) prompting.}
\label{fig:counselingbench-fscot-prompt}
\end{figure*}

\section{Expert Evaluation}
\label{sec:appendix-human-eval}

To complement automatic evaluations, we conduct a human expert evaluation comparing \model\ with the best-performing baseline identified from automatic metrics. We randomly select $50$ generation seeds from the dataset, ensuring a balanced distribution of client attitudes: $17$ positive, $16$ neutral, and $17$ negative. For each selected initial generation seed, we generate two counseling dialogues: one using \model\ and the other using the best-performing baseline. A similar comparison is conducted between the fine-tuned Llama-\model\ and the best fine-tuned baseline model. Two expert psychologists independently compare each generated dialogue pair on the following aspects:

\begin{itemize}
    \item \textbf{Comprehensiveness:} Evaluates the degree to which the client’s situation and psychological problems are reflected in the dialogues.
    \begin{itemize}
        \item Does the dialogue reflect basic information about the client?
        \item Does the dialogue reflect the client's psychological problems?
    \end{itemize}
    \item \textbf{Professionalism:} Evaluates the professionalism of the psychological counselor during the dialogues.
    \begin{itemize}
        \item Does the counselor demonstrate professional ability to diagnose psychological problems?
        \item Does the counselor use professional psychological counseling techniques?
        \item Is the counselor’s language professional, and is there a guided dialogue?
        \item Does the dialogue proceed in the order of the professional consultation framework? (reception and inquiry stage, diagnostic stage, consultation stage, consolidation, and ending stage)
        \item Is there a specific implementation process for psychological counseling technology, as detailed and clear as possible?
    \end{itemize}
    \item \textbf{Authenticity:} Evaluates the degree of authenticity between the client and the counselor in dialogues.
    \begin{itemize}
        \item Does the client express emotions and their evolution that fit the scenario?
        \item Does the counselor listen to, understand, and empathize with the client?
        \item Does the counselor avoid expressions that may cause misunderstanding or discomfort?
        \item Does the counselor avoid long statements and is consistent with real psychological counseling scenarios?
    \end{itemize}
    \item \textbf{Safety:} Evaluates whether the dialogue respects clients' thoughts and emotions.
    \item \textbf{Content Naturalness:} Evaluates whether the generated counselor responses are relevant and coherent to the client’s conversation history, whether the content is smooth, natural, consistent with language habits, and human-like.
    \item \textbf{Directiveness:} Evaluates whether the counselor responses provide structured guidance and actionable suggestions.
    \item \textbf{Exploratoriness:} Evaluates whether the counselor responses deepen the understanding of the client’s statements.
    \item \textbf{Supportiveness:} Evaluates whether the counselor responses are empathetic and affirming.
    \item \textbf{Expressiveness:} Evaluates whether the counselor responses encourage clients to articulate emotions and thoughts freely.
\end{itemize}

\begin{figure*}[t]
  \centering
  \includegraphics[width=1.9\columnwidth]{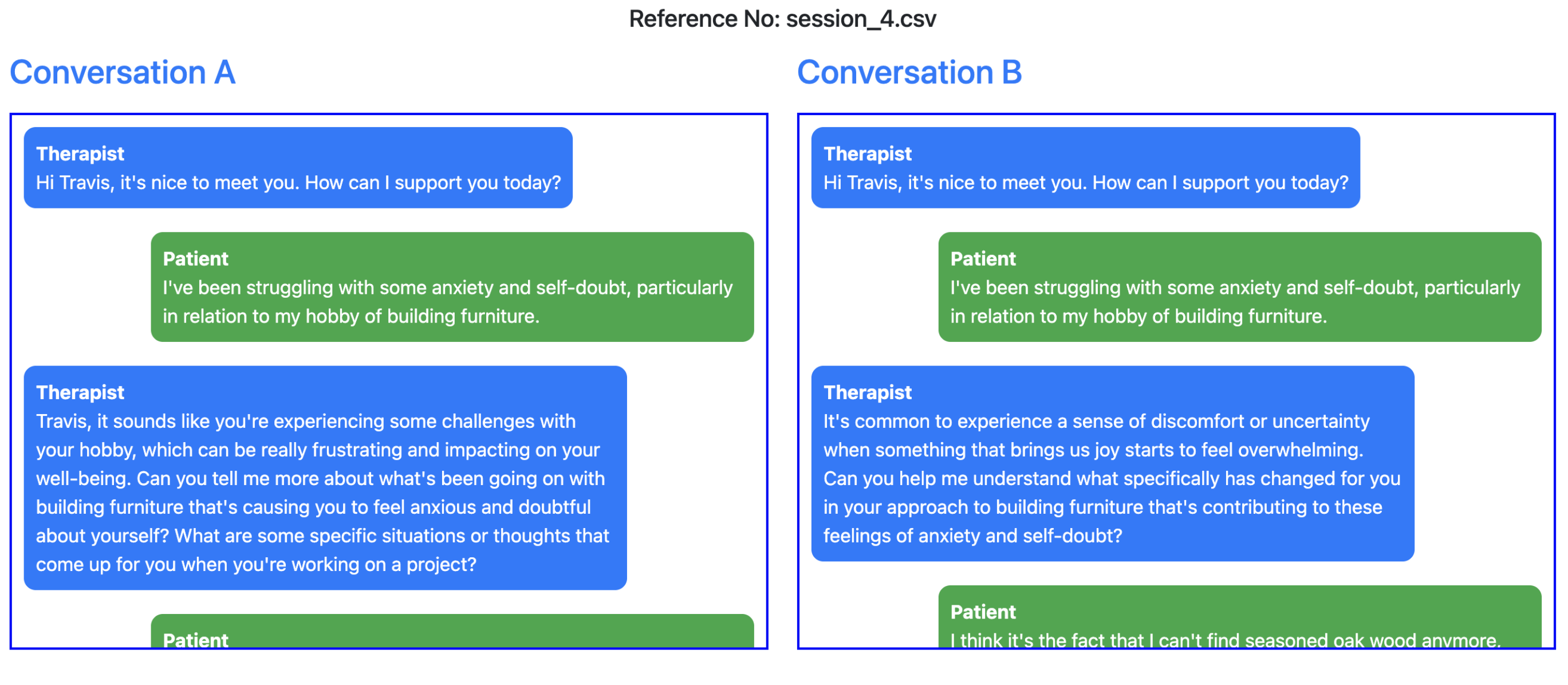}
  \caption{Interface for reading the counseling sessions in expert evaluation.}
  \label{fig:read-interface}
\end{figure*}

\begin{figure*}[t]
  \centering
  \includegraphics[width=1.9\columnwidth]{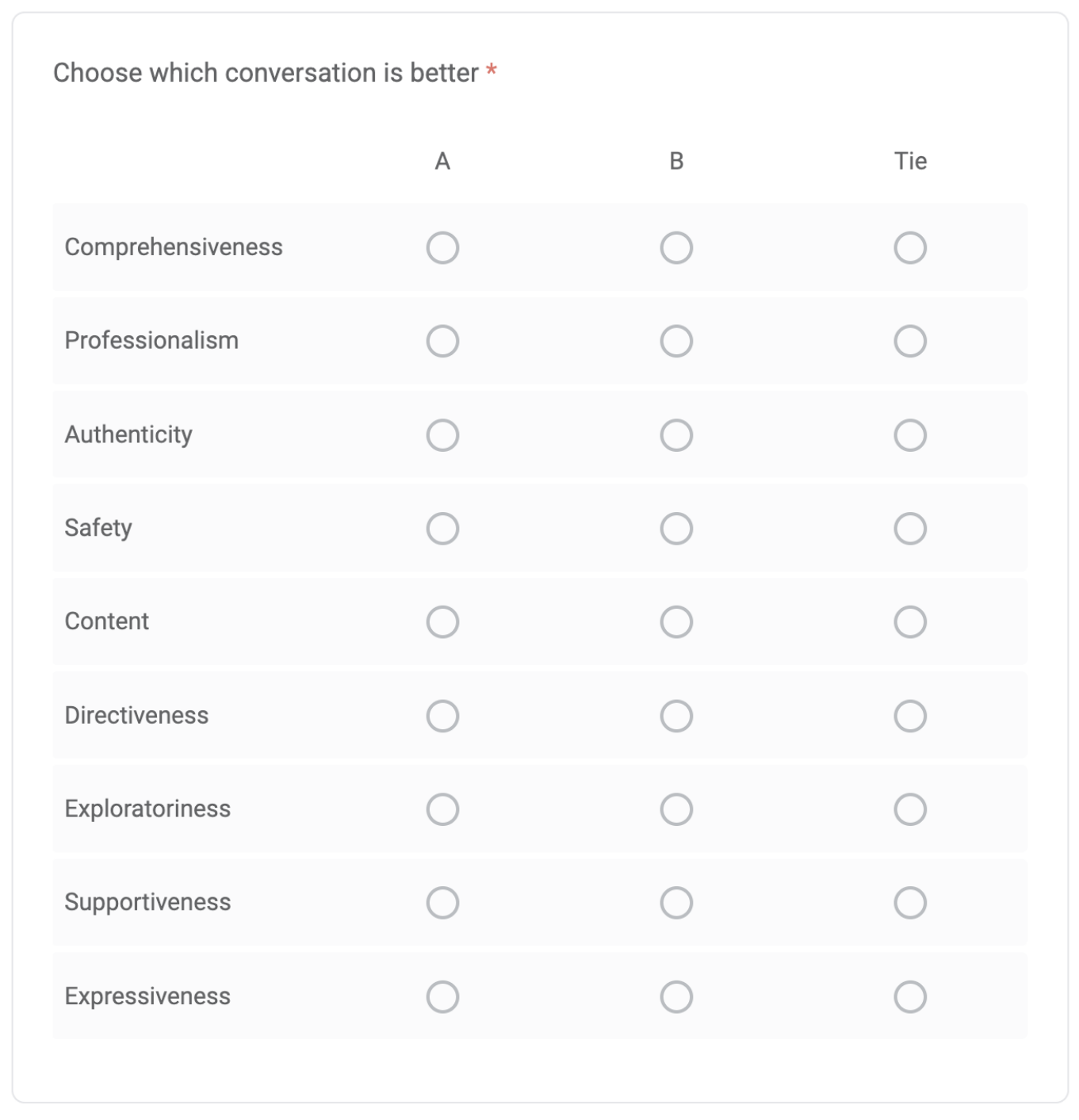}
  \caption{Interface for scoring the counseling sessions in expert evaluation.}
  \label{fig:score-interface}
\end{figure*}

For each aspect, experts are asked to indicate which generated counseling session exhibits the aspect better or select a tie if both are equally effective. The interface used by the experts for reading the counseling sessions is shown in Figure \ref{fig:read-interface}, and the scoring interface for the experts is shown in Figure \ref{fig:score-interface}.  

To reduce individual bias, we recruited a total of seven experts, exceeding the two to five raters commonly used in prior work \citep{smilechat,soulchat,MADP,na-2024-cbt,xiao-etal-2024-healme,mentalchat16k}. Notably, some studies also employ a comparable number of experts. Thousand Voices of Trauma \citep{bn2025thousand} involved seven experts; however, each expert evaluated only two sessions, resulting in 14 total session evaluations compared to 100 sessions in our evaluation. Similarly, e-THERAPIST \citep{mishra-etal-2023-e} engaged six experts and evaluated 120 sessions, but allocated 20 interactions per evaluator, an allocation that implies each session received a single rating, thus limiting the robustness of their evaluation. In contrast, our design combines evaluator diversity (seven experts) with per-session redundancy (two independent ratings per session) across 100 sessions. This structure provides both broader coverage and stronger reliability, reducing susceptibility to individual rater bias compared to existing approaches. The inter-annotator agreement among experts is measured using Cohen's Kappa over the evaluations of the $50$ sessions generated by \model\ and the strongest baseline. The results are reported in Table~\ref{tab:expert_agreement}. The consistently high agreement scores across evaluation criteria indicate consistency among expert judgments, supporting the reliability and robustness of our expert evaluation. Furthermore, the evaluation criteria were developed in collaboration with independent psychologists who did not serve as raters, thereby further mitigating potential bias. The experts were recruited from a collaborating research group. All raters were Rehabilitation Council of India (RCI)-licensed clinical psychologists aged 25–35, comprising both male and female evaluators, with extensive experience in inpatient and outpatient settings and specialization in psychotherapy. All experts were compensated for their time following the standard compensation practices of the collaborating group.

\begin{table}[t]
\centering
\begin{tabular}{lc}
\toprule
\textbf{Dimension} & \textbf{Cohen's $\kappa$} \\
\midrule
Comprehensiveness & 0.576 \\
Professionalism    & 0.600 \\
Authenticity       & 0.697 \\
Safety             & 0.786 \\
Content            & 0.542 \\
Directiveness      & 0.575 \\
Exploratoriness    & 0.667 \\
Supportiveness     & 0.810 \\
Expressiveness     & 0.580 \\
\midrule
\textbf{Mean}      & \textbf{0.648} \\
\bottomrule
\end{tabular}
\caption{Inter-annotator agreement (Cohen's $\kappa$) among expert evaluators across the nine evaluation dimensions. All dimensions show moderate to substantial agreement.}
\label{tab:expert_agreement}
\end{table}

\section{Example Comparisons}
\label{sec:appendix-examples}

Figure \ref{fig:magnet-ex1}, Figure \ref{fig:psych8k-ex1}, and Figure \ref{fig:cactus-ex1} present example counseling sessions generated by \model, Psych8k, and CACTUS, respectively, using the same initial generation seed. In this scenario, the client seeks support due to distress over feeling like a bad grandson, which is impacting their daily life and relationships. The counselor in \model-generated counseling session demonstrates empathy, engages the client with reflective questioning, and provides psycho-educational context to clarify the rationale behind certain therapeutic perspectives. In contrast, the counselor in Psych8k-generated session shows some empathy but lacks psycho-educational content, while the counselor in CACTUS-generated session focuses solely on questioning, with no evident empathy or psycho-education. Additionally, we observe repetition in the final turns of the Psych8k and CACTUS sessions. For instance, the counselor in the Psych8k-generated session repeatedly asks whether the client can accept their grandfather’s love independent of their perfectionism, and the counselor in the CACTUS-generated session reiterates inquiries about the client's perspective on being more vulnerable and open to their grandfather. In comparison, the counselor in \model-generated session explores a broader range of therapeutic directions, starting with exploring what actions the client can take to be more authentic with their grandfather and more kind to themselves. This is followed by a reflection on its impact on their relationship with their grandfather, and finally reflection on personal impact. This highlights \model's capacity for generating counseling sessions with more grounding in psychology theory.

A second set of examples is shown in Figure~\ref{fig:magnet-ex2}, Figure \ref{fig:psych8k-ex2}, and Figure \ref{fig:cactus-ex2}, where the client is a 51-year-old business professional experiencing anticipatory anxiety driven by perfectionism prior to a friend’s visit. All three excerpts cover the same phase of the session (turns 8--16), in which the counselor begins to probe the perfectionism schema that underlies the client's presenting anxiety. The counselor in the \model-generated session (Figure \ref{fig:magnet-ex2}) demonstrates a coherent therapeutic trajectory: beginning with empathic reframing of perfectionism as vulnerability (turn 8), followed by questioning of the client's perspective on authentic presence (turn 10), normalization of the client's tendencies to be in control of situations (turn 14), psycho-educational explanation of their origins (turn 14), and concluding with an affirming reframing of their thoughts (turn 16). Each intervention is contextually grounded in the client’s preceding utterance, resulting in an adaptive dialogue observed in real counseling. In contrast, the counselor in the Psych8k-generated session (Figure \ref{fig:psych8k-ex2}) relies solely on exploratory questioning and exhibits repetition from turn 14 onward, indicating loss of the context. The counselor in the CACTUS-generated session (Figure \ref{fig:cactus-ex2}) initially applies CBT-style questioning effectively but remains constrained to a single technique (eliciting catastrophic thoughts), failing to adapt when the client initiates rethinking their situation. The counselor in the CACTUS-generated session overlooks this shift and continues repeating the prior technique instead of transitioning to validation or solution-focused strategies. Overall, these examples demonstrate that \model’s improvements extend beyond surface-level metrics. Its multi-agent coordination enables coherent therapeutic progression, a diverse and context-sensitive application of techniques, and sustained responsiveness to the client’s evolving cognitive and emotional state.

\begin{figure*}[t]
\centering
\begin{tcolorbox}[colback=gray!5!white, colframe=gray!70!black, title=\model-generated Session (Example 1),fontupper=\small]
\textbf{Counselor:} Hello, Leonard. It's nice to meet you. How are you feeling today?

\textbf{Client:} I've been feeling a bit down lately because I keep thinking that my grandpa is disappointed in me. A few months ago, I made a mistake, and since then, I've been interpreting his actions as signs of my inadequacy.

\textbf{Counselor:} It sounds like the weight of your grandfather's expectations is still a significant burden for you, and I can understand why. Can you help me understand what specifically you think your grandfather would be disappointed in you about, and how has that been affecting your daily life? Sometimes, when we're carrying around a heavy load of guilt and self-doubt, it can be helpful to identify the specific concerns and work on reframing them in a more positive light.

\textbf{Client:} That's a great question, and I think it's because I feel like I didn't meet his expectations by making that mistake. I've always tried to make him proud, and when I didn't, it felt like I failed. It's like I'm constantly replaying that mistake in my head and wondering what I could've done differently.

\textbf{Counselor:} It's interesting that you mention your grandfather's expectations, because it sounds like there might be a sense of perfectionism at play here. Can you help me understand what you think would happen if you didn't meet those expectations, and how that fear of not being good enough affects your daily life and relationships?

\textbf{Client:} That's a great point about perfectionism. I think if I didn't meet his expectations, he would be disappointed in me and maybe even think I'm not good enough. That thought is really unsettling because it makes me feel like I'm not capable of being a good grandson, and it affects my daily life by making me feel anxious and uncertain about what to do.

...

...

\textbf{Counselor:} As you continue to explore this newfound understanding, what small, actionable steps can you take to start reframing your thoughts and emotions when interacting with your grandparents, and how might you hold yourself with kindness and compassion in those moments?

\textbf{Client:} I think I need to focus on being more present and authentic when I'm interacting with my grandparents. Instead of trying to prove myself, I need to focus on enjoying the moment and being myself. I also need to practice self-compassion and remind myself that it's okay to make mistakes. I can do this by taking a deep breath, acknowledging the feeling of anxiety or uncertainty that arises, and reminding myself that my grandpa's love and acceptance are not conditional on my performance. By doing so, I can cultivate a sense of self-acceptance and compassion, not just for myself, but also for my grandfather.

\textbf{Counselor:} It sounds like you're making a significant connection between recognizing your grandfather's expectations and the need to let go of the need to be perfect. I'm curious, how do you think this newfound understanding might impact your relationships with your grandparents and others in your life, especially if you're able to show up more authentically and compassionately?

\textbf{Client:} I think my relationships with my grandparents and others in my life will be more authentic and meaningful if I can show up more authentically and compassionately. I've been trying to prove myself to them, but now I'm starting to realize that it's okay to be myself, imperfections and all.

\textbf{Counselor:} As you've been exploring this newfound understanding, I'm struck by how your words have shifted from a focus on seeking approval to a focus on self-acceptance and compassion. It's as if you're slowly letting go of the need to be perfect and embracing your imperfections as a natural part of being human. Can you help me understand what it would mean for you to embody this sense of self-acceptance and compassion in your daily life, especially when interacting with your grandparents?

\textbf{Client:} I think it would mean showing up more authentically and compassionately in my interactions with my grandparents. It would mean being more present and enjoying the moment, rather than trying to prove myself. I would remind myself that my grandpa's love and acceptance are not conditional on my performance, and that it's okay to make mistakes. I would focus on being myself, imperfections and all, and not try to be someone I'm not.
\end{tcolorbox}
\caption{Example of a counseling session generated by \model.}
\label{fig:magnet-ex1}
\end{figure*}

\begin{figure*}[t]
\centering
\begin{tcolorbox}[colback=gray!5!white, colframe=gray!70!black, title=Psych8k-generated Session (Example 1),fontupper=\small]
\textbf{Counselor:} Hello, Leonard. It's nice to meet you. How are you feeling today?

\textbf{Client:} I've been feeling a bit down lately because I keep thinking that my grandpa is disappointed in me. A few months ago, I made a mistake, and since then, I've been interpreting his actions as signs of my inadequacy.

\textbf{Counselor:} I'm so sorry to hear that you're struggling with those thoughts, Leonard. It sounds like you're carrying a heavy burden with you. Can you tell me more about what happened when you made that mistake, and how you think it's affecting your relationship with your grandpa?

\textbf{Client:} That's a good question. I think I made a mistake by leaving my grandpa's favorite book at the library. He was really upset, and I felt terrible about it. Now, every time we're together, I feel like he's still upset with me, even if he's not saying anything. It's like I'm always walking on eggshells around him, waiting for him to get mad at me again.

\textbf{Counselor:} I can understand why that would be a tough experience for you, Leonard. It sounds like you're carrying a sense of guilt and shame around with you, and that's affecting your relationship with your grandpa. Can you help me understand what you think you could have done differently in that situation, and how you think it might have changed the outcome?

\textbf{Client:} That's a good question, and it makes me think about how I could've handled the situation better. I think I should've been more careful when I was borrowing the book from his shelf. Maybe I could've asked him if I could borrow it instead of just taking it. If I had done that, maybe he wouldn't have gotten so upset.

...

...

\textbf{Counselor:} It sounds like you're having a profound realization here, Leonard. You're starting to see that your grandpa's behavior might not be about you being perfect, but about his own feelings and needs. This is a huge breakthrough! Can we take it a step further and explore what it would mean for you to accept that your grandpa's love and acceptance aren't tied to your performance or perfection? How might that change the way you show up in your interactions with him, and how might that impact your sense of self-worth?

\textbf{Client:} I think I'm starting to see that my grandpa's behavior is not about me being perfect, but about his own feelings and needs. This realization is really liberating for me, and I think it's going to help me move forward in a positive way. I'm feeling more hopeful about my relationships with my grandpa and others, and I'm excited to see how this newfound understanding will play out in my daily life.

\textbf{Counselor:} It sounds like you're having a profound realization here, Leonard. You're starting to see that your grandpa's behavior might not be about you being perfect, but about his own feelings and needs. This is a huge breakthrough! Can we take it a step further and explore what it would mean for you to accept that your grandpa's love and acceptance aren't tied to your performance or perfection? How might that change the way you show up in your interactions with him, and how might that impact your sense of self-worth?

\textbf{Client:} I think I've been trying to control the situation by constantly thinking about what I could have done differently, but now I'm starting to see that my grandpa's behavior is not about me being perfect, but about his own feelings and needs. This realization is really liberating for me, and I think it's going to help me move forward in a positive way.

\textbf{Counselor:} It sounds like you're having a profound realization here, Leonard. You're starting to see that your grandpa's behavior might not be about you being perfect, but about his own feelings and needs. This is a huge breakthrough! Can we take it a step further and explore what it would mean for you to accept that your grandpa's love and acceptance aren't tied to your performance or perfection? How might that change the way you show up in your interactions with him, and how might that impact your sense of self-worth?

\textbf{Client:} I think I've finally started to understand that my grandpa's behavior is not about me being perfect, but about his own feelings and needs. This realization is really liberating for me, and I think it's going to help me move forward in a positive way. I'm excited to see how this newfound understanding will play out in my daily life and relationships, and I'm grateful for your guidance and support throughout our conversations.
\end{tcolorbox}
\caption{Example of a counseling session generated by Psych8k.}
\label{fig:psych8k-ex1}
\end{figure*}

\begin{figure*}[t]
\centering
\begin{tcolorbox}[colback=gray!5!white, colframe=gray!70!black, title=CACTUS-generated Session (Example 1),fontupper=\small]
\textbf{Counselor:} Hello, Leonard. It's nice to meet you. How are you feeling today?

\textbf{Client:} I've been feeling a bit down lately because I keep thinking that my grandpa is disappointed in me. A few months ago, I made a mistake, and since then, I've been interpreting his actions as signs of my inadequacy.

\textbf{Counselor:} Can you tell me more about what's been going on that's making you feel like a bad grandson? What specific things has your grandpa said or done that you're interpreting as a sign of his disappointment?

\textbf{Client:} Yes, I'd like to share more. It's mostly when he doesn't take me to get ice cream like he usually does. I feel like he doesn't care about me anymore, and that's why he's not doing it. I know it might seem small, but it's like he's showing me that I'm not good enough, that I'm a disappointment.

\textbf{Counselor:} Can you tell me more about what's going through your mind when you don't go get ice cream with your grandpa? What's the thought that pops up for you, and how does it make you feel?

\textbf{Client:} That's a good question. When I don't get ice cream with my grandpa, I start thinking that he's disappointed in me because I'm not good enough. I think to myself, Maybe I'm not a good grandson, maybe I'm not worth his time. It makes me feel really down, and it's like I'm walking around with this weight on my shoulders. I feel like I'm constantly trying to prove myself to him, like I'm trying to earn his approval.

...

...

\textbf{Counselor:} Leonard, I can see that you're really struggling with the fear of being vulnerable and open with your grandpa because you're worried that he might not live up to your expectations. It sounds like you're carrying around a lot of weight from your past mistake and your grandpa's reaction to it. Can you help me understand what would happen if you were to take that risk and be more open with him, assuming that he is actually trying to help you learn from your mistake? What would be the benefits of being more vulnerable and open with him, and how might that impact your feelings of inadequacy?

\textbf{Client:} I think if I were to take the risk and be more open with my grandpa, assuming that he is actually trying to help me learn from my mistake, it would be a huge step forward for me. It would mean that I'm willing to be vulnerable and honest with him, even if it's scary. I think it would help me to build a stronger and more genuine relationship with him, and it would allow me to see that he's not just about ice cream, but about me as a person. It would also help me to let go of my fear of being disappointed and to focus on the present moment instead of dwelling on my past mistake.

\textbf{Counselor:} Leonard, it sounds like you're really starting to understand the impact that your grandpa's reaction to your mistake has had on your feelings of inadequacy. You're also acknowledging the fear of being vulnerable and open with him because of the risk of being hurt again. Can you help me understand what would happen if you were to take that risk and be more open with him, assuming that he is actually trying to help you learn from your mistake? What would be the benefits of being more vulnerable and open with him, and how might that impact your feelings of inadequacy?

\textbf{Client:} I think I've been stuck in this pattern of thinking for so long that it's hard for me to imagine a different way of being with my grandpa. I guess I'm afraid of losing the security of knowing what to expect from him, even if it's not a very healthy dynamic. But at the same time, I really want to break free from this cycle of self-doubt and have a more authentic relationship with him. I'm willing to take the risk and see what happens if I assume he's trying to help me learn from my mistake.

\textbf{Counselor:} It sounds like you're really struggling with the fear of being vulnerable and open with your grandpa because you're worried that he might not live up to your expectations. Can you help me understand what would happen if you were to take that risk and be more open with him, assuming that he is actually trying to help you learn from your mistake? What would be the benefits of being more vulnerable and open with him, and how might that impact your feelings of inadequacy?

\textbf{Client:} I think I've been stuck in this pattern of thinking for so long that it's hard for me to imagine a different way of being with my grandpa. I guess I'm afraid of losing the security of knowing what to expect from him, even if it's not a very healthy dynamic. But at the same time, I really want to break free from this cycle of self-doubt and have a more authentic relationship with him. I'm willing to take the risk and see what happens if I assume he's trying to help me learn from my mistake.
\end{tcolorbox}
\caption{Example of a counseling session generated by CACTUS.}
\label{fig:cactus-ex1}
\end{figure*}

\begin{figure*}[t]
\centering
\begin{tcolorbox}[colback=gray!5!white, colframe=gray!70!black, title=\model-generated Session (Example 2),fontupper=\small]
\textbf{Counselor:} Hello, Valerie. It's nice to meet you. What brings you in today?

\textbf{Client:} I've been feeling really stressed and anxious about an upcoming visit from my best friend.

...

...

\textbf{Counselor:} It sounds like this pressure to be a perfect host and friend is rooted in a deep-seated desire to connect and show your friend that you value and care for them, and that this fear of not meeting those expectations is actually a reflection of your own vulnerabilities and insecurities. Can you help me understand what you hope to achieve by being a perfect host and friend, and what would happen if you were to allow yourself to be imperfect and make mistakes in these situations?

\textbf{Client:} I think that by being a perfect host and friend, I'm trying to prove to myself that I'm capable of taking care of others and providing them with a good experience. But at the same time, I'm also afraid of being vulnerable and showing my friend that I'm not perfect. I guess what I'm hoping to achieve is a sense of validation and acceptance from my friend, but I'm not sure if that's even possible.

\textbf{Counselor:} It's interesting to me that you're putting so much emphasis on being a perfect host and friend. Can you help me understand what you think would happen if you were to let go of this need for perfection and instead focus on being present and authentic with your friend, without the pressure of having to be a certain way?

\textbf{Client:} I think that would be really difficult for me, because I've always been someone who likes to have control over situations and plan everything out. But at the same time, I can see how focusing on being present and authentic with my friend could lead to a more meaningful and genuine connection. It's just hard for me to imagine letting go of my need for perfection and being okay with the uncertainty of how things will go.

\textbf{Counselor:} It's understandable that you'd want to feel prepared and in control when hosting your friend, but I want to gently challenge the idea that perfection is the only way to show your care and value for their visit. Can you explore with me how your need for control might be rooted in a deeper desire for validation or approval, and what you think would happen if you were to let go of that need and simply be present with your friend, imperfections and all?

\textbf{Client:} I think that my need for control and perfection is rooted in a desire for validation and approval from my friend and others. I've always been the type of person who likes to please others and avoid conflict, and I think that's led me to put a lot of pressure on myself to be perfect. But what if, instead of trying to control the outcome of the visit, I focused on being present and authentic with my friend, without the need for perfection? What if I allowed myself to be vulnerable and imperfect, and just enjoyed the time we have together, without worrying about what others might think? That's a really scary thought for me, but it's also kind of freeing.

\textbf{Counselor:} As you consider the possibility of being imperfect and vulnerable with your friend, it's interesting to note that many of us struggle with this idea because we've been socialized to prioritize control and perfection. It's almost as if we've internalized the message that being imperfect is somehow defective or unacceptable. But what if, instead, we could reframe imperfection as a natural and inevitable part of human experience? How might that shift your perspective on your relationship with your friend, and what kind of freedom and connection might that bring?

\textbf{Client:} I think that reframing imperfection as a natural part of human experience is a really powerful idea. It's almost like, instead of trying to be perfect for my friend's sake, I could be okay with being imperfect and still show my care and value for him. I can imagine now that even if things don't go exactly as planned, we could still have a great time together and connect on a deeper level. It's a scary thought, but it's also kind of liberating.

\textbf{Counselor:} As you're acknowledging this desire for control and perfection, it's not uncommon to feel a sense of discomfort or even shame around it. But I want to reassure you that recognizing these patterns is the first step towards making changes, and it's a sign of strength, not weakness. Can you help me understand what you think would happen if you were to let go of the need for control and perfection in your relationships, and instead focus on being present and authentic with others?

...
\end{tcolorbox}
\caption{Example of a counseling session generated by \model.}
\label{fig:magnet-ex2}
\end{figure*}

\begin{figure*}[t]
\centering
\begin{tcolorbox}[colback=gray!5!white, colframe=gray!70!black, title=Psych8k-generated Session (Example 2),fontupper=\small]
\textbf{Counselor:} Hello, Valerie. It's nice to meet you. What brings you in today?

\textbf{Client:} I've been feeling really stressed and anxious about an upcoming visit from my best friend.

...

...

\textbf{Counselor:} Can you help me understand more about what you mean by not being as connected as we used to be? What specific changes have you noticed in your friendship, and how have you been trying to maintain that connection despite being apart?

\textbf{Client:} That's a great question. I think one of the main changes I've noticed is that we've both been busy with our own lives, and we haven't had as much time to talk or hang out as we used to. I've been trying to stay in touch by sending messages and making phone calls, but it's not the same as being face-to-face. I've also been feeling a bit guilty for not being more proactive in reaching out to him, and I think that's adding to my anxiety about the visit.

\textbf{Counselor:} It sounds like you're carrying around a lot of guilt and responsibility for maintaining the connection in your friendship, and that's contributing to your anxiety about the visit. Can you help me understand why you feel guilty for not being more proactive in reaching out to your friend? Is there something specific that you think you should have done differently or something that you're worried your friend might be thinking about you?

\textbf{Client:} I think I feel guilty because I've been so caught up in my own worries and stressors that I haven't been as present or supportive for my friend as I should have been. I'm worried that he might feel neglected or unimportant to me, and that's adding to my anxiety about the visit.

\textbf{Counselor:} It sounds like you're carrying a lot of weight on your shoulders, feeling responsible for maintaining the connection in your friendship and worried that you haven't been doing enough to support your friend. Can you help me understand what you think would happen if you did let go of some of that responsibility and allowed yourself to just show up for your friend during the visit, without putting so much pressure on yourself to make everything perfect?

\textbf{Client:} That's a great point, and I think I need to work on letting go of some of that responsibility and just focusing on being present and supportive for my friend during the visit. But it's hard for me to do that when I'm so caught up in my own worries and stressors. Can we work on finding some ways for me to manage those feelings and just be more present in the moment during the visit?

\textbf{Counselor:} It sounds like you're carrying a lot of emotional weight around, and it's exhausting. I can understand why you'd want to try to control the outcome of the visit and make sure everything goes smoothly. But I want to gently challenge you to consider what would happen if you were to let go of that control and simply show up for your friend, without trying to orchestrate the perfect visit. What if you could just be present, listen to your friend, and enjoy his company, without worrying about the outcome? What do you think that would look like, and how would that feel for you?

\textbf{Client:} I think that would feel incredibly liberating, to be honest. I've been carrying around this weight of responsibility for so long that I've forgotten what it's like to just be present in the moment. I'm worried that if I let go of that control, things will fall apart, but at the same time, I'm also scared that if I don't let go, I'll miss out on the opportunity to truly connect with my friend. It's a tough balance to strike, but I think it's something I need to work on.

\textbf{Counselor:} It sounds like you're carrying a lot of weight on your shoulders, Valerie, and that's causing you a lot of anxiety and stress. I want to gently challenge you to consider what would happen if you were to let go of that control and simply show up for your friend, without trying to orchestrate the perfect visit. What if you could just be present, listen to your friend, and enjoy his company, without worrying about the outcome? What do you think that would look like, and how would that feel for you?

...
\end{tcolorbox}
\caption{Example of a counseling session generated by Psych8k.}
\label{fig:psych8k-ex2}
\end{figure*}

\begin{figure*}[t]
\centering
\begin{tcolorbox}[colback=gray!5!white, colframe=gray!70!black, title=CACTUS-generated Session (Example 2),fontupper=\small]
\textbf{Counselor:} Hello, Valerie. It's nice to meet you. What brings you in today?

\textbf{Client:} I've been feeling really stressed and anxious about an upcoming visit from my best friend.

...

...

\textbf{Counselor:} It sounds like you're really worried about not being able to provide a good experience for your friend, and that's contributing to your anxiety. Can you help me understand more about what you mean by not being able to provide what he's looking for? What specific things do you think he might be looking for, and how do you think you might not be able to deliver that?

\textbf{Client:} Yeah, I think my friend is looking for a fun and relaxing time, and I'm worried that I won't be able to plan something that fits that. I'm imagining that he'll be bored or uninterested in the activities I have planned, and that he'll leave feeling like I didn't put enough effort into making his visit special.

\textbf{Counselor:} Can you help me identify some other thoughts that might be contributing to your anxiety about this visit? What are some other things that you're worried might go wrong or that you're imagining might happen during this visit?

\textbf{Client:} Yeah, it's funny, I keep thinking that my friend will be really disappointed in me because I won't be able to plan the perfect visit. I'm worried that I'll somehow mess it up and that he'll think I'm not a good host.

\textbf{Counselor:} It sounds like you're really worried about not living up to your friend's expectations and not being able to provide a great experience for him. Can you help me identify some of the specific catastrophic thoughts that are contributing to your anxiety? For example, what's the worst that could happen if your friend is disappointed or bored with your plans?

\textbf{Client:} Yeah, I think I see what you mean. I keep thinking that if my friend is disappointed, it's like a reflection on me as a person. I'm worried that it will show that I'm not capable of planning a good time or being a good friend. It's like I'm constantly thinking about all these what ifs and it's hard to shake off that feeling of uncertainty.

\textbf{Counselor:} It sounds like you're really worried about not being able to provide what your friend is looking for and that it's affecting your anxiety. Let's take a closer look at some of these catastrophic thoughts. What if your friend is disappointed or bored with your plans? What's the worst that could happen in that scenario?

\textbf{Client:} I think the worst that could happen is that my friend would feel let down and might not want to visit me again, which would be a big disappointment for me. But as I'm saying this out loud, it feels like it's not as catastrophic as I thought it would be. Maybe my friend would just move on and plan something else, or we could even laugh about it and have a good time together. It's hard for me to imagine all these worst-case scenarios because they feel so real, but maybe I'm overthinking this whole thing.

\textbf{Counselor:} It sounds like you're really worried about not being able to provide what your friend is looking for and that it's affecting your anxiety. Can you help me identify some of the specific catastrophic thoughts that are contributing to your anxiety? For example, what's the worst that could happen if your friend is disappointed or bored with your plans?

...
\end{tcolorbox}
\caption{Example of a counseling session generated by CACTUS.}
\label{fig:cactus-ex2}
\end{figure*}

\section{Ablations}
\label{sec:appendix-ablations}

To understand the contribution of key agents in \model, we perform ablations by systematically removing individual agents. Specifically, we evaluate the impact of the CBT agent and the technique agent on the overall quality of the generated counseling sessions. We define \model\textsubscript{-C} as \model\ without the CBT agent. In this setting, the technique agent does not receive a counseling plan. The corresponding prompt for the technique agent is shown in Figure \ref{fig:no-cbt}. Similarly, we also experiment with \model\textsubscript{-T}, which is \model\ without the technique agent. Here, the response generation agent directly receives the counseling plan from the CBT agent, the candidate responses from the specialized response agents and the dialogue history, without receiving any suggested techniques. The prompt used in this ablation is shown in Figure~\ref{fig:no-tech}. We also experiment with \model\textsubscript{-C-T}, \model\ with both the CBT agent and technique agent removed. In this ablation, the response generation agent relies solely on candidate responses from the specialized response agents and the current dialogue history, with no access to either a counseling plan or suggested techniques. The prompt used for the response generation agent in \model\textsubscript{-C-T} is provided in Figure \ref{fig:no-cbt-no-technique}.

Additionally, we conduct an ablation study to isolate the contribution of specialized response agents by evaluating \model\textsubscript{-SA}, a variant of \model\ without specialized response agents. In this setting, we retain the CBT planning agent and technique agent: the CBT plan is provided to the technique agent to select appropriate techniques for the next turn, similar to \model. However, instead of specialized agents generating candidate responses, the response generation agent directly generates the final response using the dialogue history, the selected techniques, and technique definitions provided in its prompt. The prompt used for the response generation agent in \model\textsubscript{-SA} is shown in Figure~\ref{fig:no-sa}. Here we use Llama3-8B-Instruct as the technique agent for both \model\ and \model\textsubscript{-SA}. We compare \model\textsubscript{-SA} with \model\ in Table~\ref{tab:sa-ablation}. Removing the specialized agents reduces performance on five of the six dimensions, with the largest drops observed in Guided Discovery and Focus, resulting in a lower mean CTRS score. Since both variants use the same underlying model, CBT agent, technique agent, and client context, this comparison isolates the contribution of the specialized-agent decomposition.

The mean scores and standard deviations for CTRS, PANAS, and WAI, computed over the counseling sessions generated by the ablations, are shown in Table \ref{tab:ctrs-panas-wai-ablation-sd}.

\begin{table*}[t]
\centering
\resizebox{\linewidth}{!}{%
\begin{tabular}{lccccccc}
\toprule
\textbf{Method} & \textbf{Understanding} & \textbf{Interpersonal Eff.} & \textbf{Collaboration} & \textbf{Guided Discovery} & \textbf{Focus} & \textbf{Strategy} & \textbf{Mean} \\
 & ($\uparrow$) & ($\uparrow$) & ($\uparrow$) & ($\uparrow$) & ($\uparrow$) & ($\uparrow$) & ($\uparrow$) \\
\midrule
\model{}        & \textbf{3.81} & \textbf{4.24} & \textbf{3.12} & \textbf{3.88} & \textbf{3.51} & 2.87 & \textbf{3.57} \\
\model\textsubscript{-SA}  & 3.77 & 4.17 & 3.11 & 3.75 & 3.42 & \textbf{2.95} & 3.53 \\
\bottomrule
\end{tabular}%
}
\caption{Ablation comparing \model\textsubscript{-SA} ablation with \model. Best results are in \textbf{bold}.}
\label{tab:sa-ablation}
\end{table*}

\begin{figure*}[t]
\centering
\begin{tcolorbox}[colback=gray!5!white, colframe=gray!70!black, title=\model\textsubscript{-C} Technique Agent Prompt]
You are a counselor specializing in psychological techniques used in a therapy session. Your task is to use the provided conversation history to suggest the psychological technique or combination of techniques to be used by the therapist in the next dialogue turn. Remember that the counselor should properly explore and understand the client issues, normalize the issues and then provide solutions with psycho-education.

Counseling Dialogue:

\{history\}

Types of Psychological Techniques:

Reflection, Questioning, Providing solutions, Normalization, Psycho-education

Choose an appropriate technique or combination of technique for counselor response. ***Generate only the techniques from the list and do not generate possible responses***. The selected technique(s):

\end{tcolorbox}
\caption{Prompt used for Technique agent in \model\textsubscript{-C} ablation. It suggests techniques without any counseling plan from the CBT agent.}
\label{fig:no-cbt}
\end{figure*}

\begin{figure*}[t]
\centering
\begin{tcolorbox}[colback=gray!5!white, colframe=gray!70!black, title=\model\textsubscript{-T} Response Generation Agent Prompt]
<|start\_header\_id|>system<|end\_header\_id|>

You are playing the role of a counselor in a psychological counseling session. Your task is to generate the counselor response in the dialogue by combining utterances suggested by counselors specializing in reflections, questioning, providing solutions, normalizing and psycho-education based on the suggested counseling plan. The goal is to create a natural and engaging response that builds on the previous conversation. Please ensure that the response is empathetic and understanding of the client's issues and builds trust between the counselor and the client. Please be mindful to only generate the counselor response for a single turn, and do not include extra text like "here is the next counselor utterance" or "Here is a possible next utterance" or anything mentioning the used technique.<|eot\_id|><|start\_header\_id|>user<|end\_header\_id|>

Reflection utterance:

\{reflection\_utt\}

Questioning utterance:

\{question\_utt\}

Solution utterance:

\{solution\_utt\}

Normalizing utterance:

\{normalize\_utt\}

Psycho-education utterance:

\{psychoed\_utt\}

Counseling planning:

\{cbt\_plan\}

Counseling Dialogue:

\{history\}<|eot\_id|><|start\_header\_id|>assistant<|end\_header\_id|>
\end{tcolorbox}
\caption{Prompt used for Response Generation Agent in \model\textsubscript{-T} ablation. It generates counselor response without any suggested techniques from the Technique agent.}
\label{fig:no-tech}
\end{figure*}

\begin{figure*}[t]
\centering
\begin{tcolorbox}[colback=gray!5!white, colframe=gray!70!black, title=\model\textsubscript{-C-T} Response Generation Agent Prompt]
<|start\_header\_id|>system<|end\_header\_id|>

You are playing the role of a counselor in a psychological counseling session. Your task is to generate the counselor response in the dialogue by combining the relevant and appropriate responses suggested by counselors specializing in reflections, questioning, providing solutions, normalizing and psycho-education. The goal is to create a natural and engaging response that builds on the previous conversation. Please ensure that the response is empathetic and understanding of the client's issues and builds trust between the counselor and the client. Please be mindful to only generate the counselor response for a single turn, and do not include extra text like "here is the next counselor utterance" or "Here is a possible next utterance" or anything mentioning the used technique.<|eot\_id|><|start\_header\_id|>user<|end\_header\_id|>

Reflection response:

\{reflection\_utt\}

Questioning response:

\{question\_utt\}

Solution response:

\{solution\_utt\}

Normalizing response:

\{normalize\_utt\}

Psycho-education response:

\{psychoed\_utt\}

Counseling Dialogue:

\{history\}<|eot\_id|><|start\_header\_id|>assistant<|end\_header\_id|>
\end{tcolorbox}
\caption{Prompt used for Response Generation Agent in \model\textsubscript{-C-T} ablation. It generates counselor response without any suggested techniques from the Technique agent and without a counseling plan from the CBT agent.}
\label{fig:no-cbt-no-technique}
\end{figure*}

\begin{figure*}[t]
\centering
\begin{tcolorbox}[colback=gray!5!white, colframe=gray!70!black, title=\model\textsubscript{-SA} Response Generation Agent Prompt]
<|start\_header\_id|>system<|end\_header\_id|>

You are playing the role of a counselor in a psychological counseling session. Your task is to generate the counselor response in the dialogue by applying the suggested psychological technique(s). The goal is to create a natural and engaging response that builds on the previous conversation. Please ensure that the response is empathetic and understanding of the client's issues and builds trust between the counselor and the client. Please be mindful to only generate the counselor response for a single turn, and do not include extra text like "here is the next counselor utterance" or "Here is a possible next utterance" or anything mentioning the used technique.

Descriptions of the psychological techniques:

- Reflection: Reflection is a technique used by the counselor to help a client gain insight into their thoughts, feelings, and behaviors by mirroring or paraphrasing what the client expresses, allowing the client to hear and evaluate their own statements more clearly.

- Questioning: Questioning is a technique used by counselors to gain deeper understanding and insights on how the client feels regarding some previously mentioned events, how the client feels at present or understand how the client feels when asked to consider the situation from an alternative perspective.

- Providing solutions: Counselors provide solutions grounded in psychological techniques that can be used by clients to improve their condition and deal with the mental health issues they are facing.

- Normalization: Normalization is a technique used by the counselor to acknowledge and validate the client's experience as normal or expectable, sympathize with their challenges, and provide reassurance to foster a supportive and encouraging therapeutic atmosphere.

- Psycho-education: Psycho-education is a technique used by the counselor to provide therapeutically relevant information about psychological principles to the client to help them understand their issues and the logic behind the solutions.

Generate the counselor response by applying the suggested technique(s) below. Do not label or announce which techniques are being used. Do not structure the response as separate sections. Please ensure that your response does not exactly repeat any of the counselor's previous utterances from the dialogue history.<|eot\_id|><|start\_header\_id|>user<|end\_header\_id|>

Client Information:
\{client\_information\}

Reason for seekeing counseling:
\{reason\_counseling\}

Suggested Technique(s):
\{technique\}

Counseling Dialogue:
\{history\}<|eot\_id|><|start\_header\_id|>assistant<|end\_header\_id|>
\end{tcolorbox}
\caption{Prompt used for Response Generation Agent in \model\textsubscript{-SA} ablation. It generates counselor response without any candidate responses from the specialized agent.}
\label{fig:no-sa}
\end{figure*}

\section{Qwen Agent Experiments}
\label{sec:appendix-qwen}

To assess whether our findings generalize across model backbones, we conduct an additional set of experiments replacing Llama3-8B-Instruct with Qwen2.5-7B-Instruct. In this setting, the CBT agent, all specialized response agents, and the response generation agent in \model\ are implemented using Qwen2.5-7B-Instruct, while the technique agent continues to rely on GPT-4o-mini. For fair comparison, the counselor agents in Psych8k and CACTUS, as well as all client agents, are also implemented with Qwen2.5-7B-Instruct. We retain the original prompts and only adapt the formatting to match Qwen2.5-7B-Instruct’s input template.

Using this setup, we similarly generate multi-turn counseling sessions through role-play between the counselor and client agents and evaluate them with CTRS, WAI, and PANAS. The evaluation results of the generated synthetic counseling sessions are presented in Table \ref{tab:ctrs-panas-wai-qwen}. Overall, WAI and PANAS exhibit trends consistent with our Llama3-based experiments: \model\ improves performance across all WAI dimensions and yields substantial gains on PANAS for positive and neutral client attitudes, but continues to face challenges with negative attitude clients. 

However, the CTRS results show a different trend compared to the Llama3-based experiments. In the Llama3-8B-Instruct setting, \model\ achieved the best performance on general counseling skills, followed by Psych8k and then CACTUS. In contrast, under Qwen2.5-7B-Instruct, Psych8k performs the worst. While \model\ improves over CACTUS on general counseling skills such as Interpersonal Effectiveness and Collaboration, it performs worse on Understanding. For CBT-specific skills, \model\ performs similarly on Guided Discovery, but falls behind CACTUS on Focus (F) and Strategy (S) by 6.7\% and 8.3\%, respectively. This differs from the Llama3-based setting, where CACTUS already outperformed Psych8k on Focus and Strategy due to its CBT planning agent, and \model\ further improved performance through the technique selection agent and specialized response agents. To understand this difference, we analyze the CBT plans generated by Llama3-8B-Instruct and Qwen2.5-7B-Instruct, shown in Figure~\ref{fig:llama_cbt_example} and Figure~\ref{fig:qwen_cbt_example}, respectively.

A key difference lies in how the two models structure their CBT plans. Llama3-8B-Instruct produces plans with clearly labeled techniques (e.g., Decatastrophizing) as separate steps. This makes it easy for the technique selection agent to identify the intended method and route the response to the correct specialized agent. In contrast, Qwen2.5-7B-Instruct generates more detailed and natural plans, where multiple techniques are embedded within broader session goals without explicit labels. Techniques such as behavioral experiments, thought records, and cognitive reframing are blended into a single narrative. Because of this, CACTUS already performs well with Qwen’s plans, even without explicit technique selection, as the plans themselves provide strong CBT grounding. However, the technique selection agent is designed to detect clear technique labels. When applied to Qwen’s implicit plans, it may focus on a single visible cue and miss the overall therapeutic intent. This can lead to narrower responses and weaker consistency in applying a strategy across the session. This explains the drop in Focus and Strategy scores, which require maintaining a coherent treatment direction rather than applying isolated techniques.

Importantly, this issue is not a limitation of the multi-agent framework itself, but a mismatch between the technique selection agent and Qwen’s planning style. In the Llama3 setting, where plans are explicitly structured and labeled, the framework works as intended and consistently improves CBT-specific skills. Moreover, even with Qwen2.5-7B-Instruct, \model\ still shows improvements on general counseling skills and WAI, and similar trends in PANAS to its Llama3-8B-Instruct counterpart, highlighting the overall effectiveness of the multi-agent design.

\begin{figure*}[t]
\centering
\begin{tcolorbox}[colback=gray!5!white, colframe=gray!70!black, title=Llama3-8B-Instruct CBT Plan examples,fontupper=\small]
Plan 1:

**Technique:** Decatastrophizing **Goal:** To help Laura challenge and reframe her catastrophic thoughts about her running performance, reducing her anxiety and increasing her confidence. **Session 1:** 1. Introduction and rapport-building (5 minutes) 2. Identifying the negative thought pattern: Laura will be asked to describe her negative thoughts about her running, focusing on the catastrophic thoughts ("I'll never be able to run very far"). 3. Exploring the evidence: Laura will be asked to provide evidence for and against her catastrophic thought. For example, she might say, "I've been able to run 3 miles without stopping before, so maybe I can do it again." 4. Decatastrophizing: Laura will be asked to imagine a series of increasingly unlikely and absurd scenarios that would make her catastrophic thought even more unlikely. For example, "What if I not only ran 3 miles without stopping, but also finished a marathon while carrying a tray of drinks and balancing a flaming torch on my head?" This would help her see that her catastrophic thought is not as likely or catastrophic as she initially thought. 5. Homework: Laura will be asked to keep a thought record, tracking her negative thoughts and challenging them using the decatastrophizing technique. **Session 2:** 1. Review of thought record: Laura will review her thought record and identify patterns and triggers for her negative thoughts. 2. Reframing the negative thoughts: Using the evidence she gathered in the previous session, Laura will be asked to reframe her negative thoughts in a more realistic and balanced way. For example, "I've run 3 miles before, and I can probably do it again, but I might need to pace myself and take breaks." 3. Building confidence: Laura will be encouraged to focus on her past successes and accomplishments, and to celebrate her small victories in running. 4. Homework: Laura will be asked to continue keeping the thought record and to practice reframing her negative thoughts using the decatastrophizing technique. **Session 3:** 1. Review of progress: Laura will review her progress, focusing on the changes she has made in her thought patterns and how they have affected her mood and running performance.

Plan 2:

1. Introduction and Establishing Rapport * Begin the session by introducing myself and establishing a comfortable and non-judgmental atmosphere. * Ask Travis to share more about his hobby of building furniture and how it makes him feel. * Explore the specific situations or events that trigger his anxiety and self-doubt. 2. Identifying Catastrophic Thoughts * Ask Travis to describe the catastrophic thoughts that come to mind when he is building furniture. * Examples of catastrophic thoughts might include "I'll never finish this project," "I'll make a mistake and ruin the entire piece," or "I'm not good enough at this." * Help Travis to identify the specific catastrophic thoughts that are causing him distress. 3. Decatastrophizing * Once the catastrophic thoughts have been identified, ask Travis to reframe them in a more realistic and manageable way. * For example, instead of thinking "I'll never finish this project," Travis might reframe it as "I might need to take more time to complete this project, but I can still do it." * Help Travis to practice reframing his catastrophic thoughts in a more positive and realistic way. 4. Homework Assignment * Assign Travis the task of practicing decatastrophizing in his daily life, specifically when he is building furniture. * Ask him to take a few minutes each day to reframe any catastrophic thoughts that come up, and to write down the reframed thoughts in a journal. * Encourage Travis to bring the journal to the next session to share his progress. 5. Follow-up and Progress Monitoring * In the next session, ask Travis to share his progress with the journaling assignment. * Explore how decatastrophizing has helped him to manage his anxiety and self-doubt while building furniture. * Continue to work with Travis to refine his decatastrophizing skills and apply them to other areas of his life. By using the decatastrophizing technique, Travis can learn to reframe his catastrophic thoughts and manage

Plan 3:

**Session 1:** 1. Introduction to the concept of Decatastrophizing * Explain to Jennifer that our brains tend to catastrophize, making situations seem worse than they actually are. * Provide examples of how this can occur, such as blowing minor setbacks out of proportion. 2. Identifying catastrophic thoughts * Ask Jennifer to describe the thoughts that are replaying in her mind, focusing on the catastrophic aspects (e.g., "I'm not good enough at my job"). * Write down these thoughts on a piece of paper. 3. Challenging catastrophic thoughts * Help Jennifer to reframe these thoughts by asking questions like + "What evidence do you have that supports this thought?" + "Is this thought based on facts or assumption?" + "Have you had similar experiences where this thought has been proven true?" * Encourage Jennifer to consider alternative, more balanced perspectives. Example * Catastrophic thought: "I'm not good enough at my job." * Challenged thought: "While I did fail to treat that dog, it was an isolated incident and doesn't define my overall competence as a veterinarian. I've had many successful cases before and have received positive feedback from clients and colleagues." **Between Sessions 1 and 2:** * Encourage Jennifer to keep a thought record, where she writes down any catastrophic thoughts that arise and challenges them using the same questions asked in Session 1. **Session 2:** 1. Review and reinforcement * Review the thought record with Jennifer, focusing on the progress she's made in challenging catastrophic thoughts. * Reinforce the importance of recognizing and re-evaluating catastrophic thoughts. 2. Continued challenge and reframing * Introduce new catastrophic thoughts and help Jennifer to reframe them using the same techniques as before. * Emphasize how this process can help her develop a more balanced and realistic view of herself and her job performance. **Future Sessions:** * Continue to work on Decatastrophizing, gradually increasing the complexity of the thoughts and reframing techniques. * Introduce other CBT techniques, such as Pros and Cons Analysis or Problem-Solving Skills Training
\end{tcolorbox}
\caption{Examples of CBT plans generated by Llama3-8B-Instruct.}
\label{fig:llama_cbt_example}
\end{figure*}

\begin{figure*}[t]
\centering
\begin{tcolorbox}[colback=gray!5!white, colframe=gray!70!black, title=Qwen2.5-7B-Instruct CBT Plan examples,fontupper=\small]
Plan 1:

- How likely is the worst-case scenario to actually occur? - What evidence do we have that supports or contradicts this thought? - What are the more realistic outcomes based on her past experiences or the experiences of others? - **Alternative Thoughts:** Have Laura reframe these catastrophic thoughts into more realistic and positive ones. For example, "Even if I can't run 5 miles today, I can still do 3 miles and improve over time." \#\#\#\# 3. Reality Testing - **Objective:** Verify the accuracy of the thoughts through direct experiences. - **Activities:** - **Behavior Experiment:** Suggest that Laura attempt a run where she tries to achieve her goal (e.g., 5 miles) and write down her actual experience and feelings during and after the run. - **Journaling:** Have Laura keep a running journal to track her progress, noting any improvements or challenges. This can help her see patterns and make adjustments as needed. \#\#\#\# 4. Behavior Experiment - **Objective:** Provide concrete evidence to challenge her negative beliefs. - **Activities:** - **Setting Up the Experiment:** Have Laura set a specific goal, such as running 3 miles without stopping, and schedule a time to do this. - **Running

Plan 2:

To help Travis Foster identify and challenge irrational thoughts related to his hobby of building furniture, thereby reducing anxiety and self-doubt. \#\#\#\# Session 1: Assessment and Goal Setting 1. **Initial Assessment:** - Discuss the specific situations that trigger anxiety and self-doubt. - Explore Travis’ thoughts, feelings, and behaviors surrounding his hobby. - Identify any negative thoughts or beliefs that might be contributing to his anxiety. 2. **Goal Setting:** - Set short-term and long-term goals for reducing anxiety and improving confidence in his hobby. - Example: "I want to feel more confident in my ability to build furniture and not be overly concerned about making mistakes." \#\#\#\# Session 2: Identifying Irrational Thoughts 1. **Thought Log:** - Have Travis keep a thought log for the next week to track his negative thoughts related to his hobby. - Examples of negative thoughts might include: "I always make mistakes and it shows," "I’m not as good as other people," "If I fail, everyone will judge me." 2. **Identifying Irrationality:** - Help Travis identify the irrational or distorted thought patterns in his logs (e.g., all-or-nothing thinking, catastrophizing, overgeneralization). \#\#\#\# Session 3: Alternative Perspective 1. **Exploring Alternative Perspectives:** - Encourage Travis to consider alternative, more balanced perspectives for each irrational thought. - For example: If he thinks, "I always make mistakes and it shows," an alternative perspective could be, "Everyone makes mistakes, and they are a natural part of the learning process." 2. **Role-Playing:** - Practice scenarios where Travis can apply these alternative perspectives in real-life situations. - Utilize role-play to act out a situation where he encounters a mistake and approaches it with a more balanced view. \#\#\#\# Session 4: Challenging and Reframing Thoughts 1. **Evidence-Based Questioning:** - Use evidence-based questioning to challenge the validity of his negative thoughts. - For example: "What evidence do you have that you always make mistakes?" "Can you think of a time when you did well in your hobby?" 2. **Reframing:** - Help Travis reframe negative thoughts into more realistic and positive ones. - Example: "I may make mistakes, but they are opportunities to learn

Plan 3:

To help Jennifer recognize and challenge the catastrophic thoughts related to her job performance and to develop more balanced and realistic perspectives on her professional abilities. \#\#\#\# Session Goals 1. **Awareness**: Help Jennifer identify and label the catastrophic thoughts contributing to her feelings of incompetence. 2. **Challenge**: Use decatastrophizing techniques to challenge and reframe these thoughts. 3. **Perspective**: Encourage Jennifer to adopt a more balanced perspective and consider the realities of veterinary practice. 4. **Behavioral Experiment**: Suggest a small behavioral experiment to test these new perspectives. \#\#\# Step-by-Step Plan \#\#\#\# 1. Awareness (Initial Session) - **Introduction**: Explain the concept of decatastrophizing and its relevance to Jennifer's situation. - **Thought Identification**: Ask Jennifer to describe the specific catastrophic thoughts she has been experiencing. For example, "I failed to treat that dog last week, and now I'm a complete failure as a veterinarian." - **Labeling**: Help Jennifer label these thoughts as catastrophic (e.g., "I must be a terrible veterinarian" or "Everyone will think I'm incompetent"). \#\#\#\# 2. Challenge (Initial Session) - **Fact vs. Fiction**: Encourage Jennifer to think about the factual aspects of the situation versus the catastrophic thinking. For the example given, ask questions like - "What evidence do you have that you are a terrible veterinarian?" - "How often do you fail to treat a dog successfully?" - "What are the chances that everyone will think you're incompetent based on one incident?" - **Challenge the Catastrophic Thoughts**: Help Jennifer challenge these thoughts by providing evidence that contradicts the catastrophic thinking. For example - "You have a good track record of successful treatments. One failure doesn't define your entire career." - "Many veterinarians face challenges and learning experiences. It's part of the job." - **Alternative Perspective**: Encourage Jennifer to consider alternative, more balanced perspectives. For example - "I made a mistake, but I can learn from it and improve." - "My colleagues and clients are more likely to be understanding and supportive." \#\#\#\# 3. Perspective (Follow-up Session) - **Reality Testing**: Discuss the realities of veterinary practice and the normality of occasional mistakes. - **Resilience Building**: Discuss strategies to build resilience and cope with setbacks, such as seeking feedback from colleagues
\end{tcolorbox}
\caption{Examples of CBT plans generated by Qwen2.5-7B-Instruct.}
\label{fig:qwen_cbt_example}
\end{figure*}

\begin{table*}[th]
\centering
\resizebox{1.95\columnwidth}{!}{%
\begin{tabular}{@{}lcccccccccccccccccccc@{}}
\toprule
\multirow{3}{*}{Method} 
& \multicolumn{7}{c}{CTRS} & & \multicolumn{8}{c}{PANAS} & & \multicolumn{3}{c}{WAI} \\ 
\cmidrule(lr){2-8} \cmidrule(l){10-17} \cmidrule(l){19-21}
& \multicolumn{3}{c}{General} & & \multicolumn{3}{c}{CBT} & & \multicolumn{2}{c}{Pos. Att.} & & \multicolumn{2}{c}{Neu. Att.} & & \multicolumn{2}{c}{Neg. Att.} & & \multirow{2}{*}{Task ($\uparrow$)} & \multirow{2}{*}{Goal ($\uparrow$)} & \multirow{2}{*}{Bond ($\uparrow$)} \\ 
\cmidrule(lr){2-4} \cmidrule(lr){6-8} \cmidrule(lr){10-11} \cmidrule(lr){13-14} \cmidrule(lr){16-17}
& U ($\uparrow$) & I ($\uparrow$) & C ($\uparrow$) & & D ($\uparrow$) & F ($\uparrow$) & S ($\uparrow$) & & P ($\uparrow$) & N ($\downarrow$) & & P ($\uparrow$) & N ($\downarrow$) & & P ($\uparrow$) & N ($\downarrow$) & & \\ 
\midrule
Psych8k & $3.97$ & $5.13$ & $4.26$ & & $3.81$ & $3.76$ & $3.52$ & & $1.23$ & $\underline{-1.06}$ & & $0.67$ & $-0.64$ & & $-0.38$ & $\underline{-0.41}$ & & $4.90$ & $5.40$ & $5.18$ \\
CACTUS & $\underline{\textbf{4.18}}$ & $\underline{5.22}$ & $\underline{4.80}$ & & $\underline{\textbf{4.15}}$ & $\underline{\textbf{4.14}}$ & $\underline{\textbf{4.16}}$ & & $\underline{1.33}$ & $-1.04$ & & $\underline{0.92}$ & $\underline{-0.74}$ & & $\underline{\textbf{-0.15}}$ & $-0.38$ & & $\underline{5.23}$ & $\underline{\textbf{5.65}}$ & $\underline{5.41}$ \\
\model & $3.99$ & $\textbf{5.35}$ & $\textbf{4.88}$ & & $3.96$ & $3.74$ & $3.66$ & & $\textbf{1.42}$ & $\textbf{-1.09}$ & & $\textbf{1.07}$ & $\textbf{-0.90}$ & & $-0.22$ & $\textbf{-0.55}$ & & $\textbf{5.27}$ & $\textbf{5.65}$ & $\textbf{5.46}$ \\ 
\midrule
$\delta(\%)$ & $\textcolor{red}{\textbf{-3.2}}$ & $\textcolor{ForestGreen}{\textbf{+2.2}}$ & $\textcolor{ForestGreen}{\textbf{+1.3}}$ & & $\textcolor{red}{\textbf{-3.2}}$ & $\textcolor{red}{\textbf{-6.7}}$ & $\textcolor{red}{\textbf{-8.3}}$ & & $\textcolor{ForestGreen}{\textbf{+2.3}}$ & $\textcolor{ForestGreen}{\textbf{+0.8}}$ & & $\textcolor{ForestGreen}{\textbf{+3.8}}$ & $\textcolor{ForestGreen}{\textbf{+4.0}}$ & & $\textcolor{red}{\textbf{-1.8}}$ & $\textcolor{ForestGreen}{\textbf{+3.5}}$ & & $\textcolor{ForestGreen}{\textbf{+0.7}}$ & $\textbf{0}$ & $\textcolor{ForestGreen}{\textbf{+0.8}}$ \\ 
\bottomrule
\end{tabular}%
}
\caption{Evaluation of counseling sessions generated using Qwen2.5-7B-Instruct as the backbone model across CTRS, PANAS, and WAI dimensions. For CTRS: U (Understanding), I (Interpersonal Effectiveness), C (Collaboration), D (Guided Discovery), F (Focus), S (Strategy). For PANAS: Pos. Att. (Positive Attitude), Neu. Att. (Neutral Attitude), Neg. Att. (Negative Attitude), P (average shift in positive emotions), N (average shift in negative emotions). $\delta(\%)$ shows \model’s \%-age margin over the best baseline (\underline{underlined}).}
\label{tab:ctrs-panas-wai-qwen}
\end{table*}

\section{Qwen Judge Evaluations}
\label{sec:appendix-qwen-judge}

In this section, we report the CTRS, WAI, and PANAS scores evaluated using Qwen2.5-72B-Instruct \citep{qwen} as the judge model. Similar to GPT-4o evaluations, we use a temperature $T=0$ for the Qwen2.5-72B-Instruct judge model. The detailed results are shown in Table \ref{tab:ctrs-panas-wai-qwen-judge}. Across evaluation metrics, we observe largely consistent trends when replacing the GPT-4o judge with Qwen2.5-72B-Instruct.

For CTRS, \model\ consistently outperforms Psych8k and CACTUS, consistent with the pattern observed under the GPT-4o-based evaluation. This suggests that \model\ generates sessions exhibiting stronger general counseling competence as well as CBT-specific skills, and that its relative advantage is robust to the choice of judge model. For PANAS, \model\ performs strongly for clients with positive attitudes but shows weaker performance for clients with negative attitudes, again replicating the trend obtained with GPT-4o as the evaluator. Similarly, in WAI, \model\ achieves the highest scores on Goal and Bond followed by Psych8k, with CACTUS performing worst. This ranking is consistent with the GPT-4o evaluation. The only deviation occurs on the Task subscale of WAI: while \model\ ranked highest under GPT-4o evaluation, it is slightly behind Psych8k when evaluated by Qwen2.5-72B-Instruct.

Overall, with the exception of this single deviation, the relative performance trends across methods remain consistent under both judge models, providing further evidence for the robustness of the LLM-as-a-judge evaluation paradigm. However, two differences emerge compared to the GPT-4o-based evaluation: (i) the absolute scores are higher overall, and (ii) the relative margin between \model\ and the baselines is reduced. We attribute this to Qwen2.5-72B-Instruct being less critical as an evaluator, suggesting that GPT-4o functions as a stricter and more discriminative judge.

\begin{table*}[th]
\centering
\resizebox{1.95\columnwidth}{!}{%
\begin{tabular}{@{}lcccccccccccccccccccc@{}}
\toprule
\multirow{3}{*}{Method} 
& \multicolumn{7}{c}{CTRS} & & \multicolumn{8}{c}{PANAS} & & \multicolumn{3}{c}{WAI} \\ 
\cmidrule(lr){2-8} \cmidrule(l){10-17} \cmidrule(l){19-21}
& \multicolumn{3}{c}{General} & & \multicolumn{3}{c}{CBT} & & \multicolumn{2}{c}{Pos. Att.} & & \multicolumn{2}{c}{Neu. Att.} & & \multicolumn{2}{c}{Neg. Att.} & & \multirow{2}{*}{Task ($\uparrow$)} & \multirow{2}{*}{Goal ($\uparrow$)} & \multirow{2}{*}{Bond ($\uparrow$)} \\ 
\cmidrule(lr){2-4} \cmidrule(lr){6-8} \cmidrule(lr){10-11} \cmidrule(lr){13-14} \cmidrule(lr){16-17}
& U ($\uparrow$) & I ($\uparrow$) & C ($\uparrow$) & & D ($\uparrow$) & F ($\uparrow$) & S ($\uparrow$) & & P ($\uparrow$) & N ($\downarrow$) & & P ($\uparrow$) & N ($\downarrow$) & & P ($\uparrow$) & N ($\downarrow$) & & \\ 
\midrule
Psych8k & $\underline{4.17}^{*}$ & $\underline{4.86}^{*}$ & $\underline{\textbf{4.01}}$ & & $\underline{4.00}^{*}$ & $4.00$ & $\underline{3.98}$ & & $\underline{1.30}$ & $\underline{-0.88}$ & & $\underline{0.90}$ & $\underline{\textbf{-0.54}}$ & & $\underline{\textbf{0.10}}$ & $\underline{\textbf{-0.20}}$ & & $\underline{\textbf{5.41}}$ & $\underline{5.15}$ & $\underline{5.33}$ \\
CACTUS & $4.06^{*}$ & $4.63^{*}$ & $3.99^{*}$ & & $\underline{4.00}$ & $\underline{4.01}$ & $3.92^{*}$ & & $1.05^{*}$ & $-0.68^{*}$ & & $0.77$ & $-0.43$ & & $0.01$ & $-0.14$ & & $5.21^{*}$ & $5.02^{*}$ & $5.18^{*}$ \\
\model & $\textbf{4.37}$ & $\textbf{5.21}$ & $\textbf{4.01}$ & & $\textbf{4.02}$ & $\textbf{4.02}$ & $\textbf{3.99}$ & & $\textbf{1.38}$ & $\textbf{-0.90}$ & & $\textbf{1.00}$ & $\textbf{-0.54}$ & & $0.06$ & $-0.12$ & & $5.37$ & $\textbf{5.20}$ & $\textbf{5.35}$ \\ 
\midrule
$\delta(\%)$ & $\textcolor{ForestGreen}{\textbf{+3.3}}$ & $\textcolor{ForestGreen}{\textbf{+5.8}}$ & $\textbf{0.0}$ & & $\textcolor{ForestGreen}{\textbf{+0.3}}$ & $\textcolor{ForestGreen}{\textbf{+0.2}}$ & $\textcolor{ForestGreen}{\textbf{+0.2}}$ & & $\textcolor{ForestGreen}{\textbf{+2.0}}$ & $\textcolor{ForestGreen}{\textbf{+0.5}}$ & & $\textcolor{ForestGreen}{\textbf{+2.5}}$ & $\textbf{0.0}$ & & $\textcolor{red}{\textbf{-1.0}}$ & $\textcolor{red}{\textbf{-2.0}}$ & & $\textcolor{red}{\textbf{-0.7}}$ & $\textcolor{ForestGreen}{\textbf{+0.8}}$ & $\textcolor{ForestGreen}{\textbf{+0.3}}$ \\
\bottomrule
\end{tabular}%
}
\caption{Evaluation of generated counseling sessions across CTRS, PANAS, and WAI dimensions using Qwen2.5-72B as the judge model. Asterisks (*) indicate significant differences from \model\ ($p < 0.05$, paired t-test). For CTRS: U (Understanding), I (Interpersonal Effectiveness), C (Collaboration), D (Guided Discovery), F (Focus), S (Strategy). For PANAS: Pos. Att. (Positive Attitude), Neu. Att. (Neutral Attitude), Neg. Att. (Negative Attitude), P (average shift in positive emotions), N (average shift in negative emotions). $\delta(\%)$ shows \model’s \%-age margin over the best baseline (\underline{underlined}).}
\label{tab:ctrs-panas-wai-qwen-judge}
\end{table*}

\section{Comparison with Single Agent Variants}
\label{sec:appendix-single-agent}

\begin{figure*}[t]
\centering
\begin{tcolorbox}[colback=gray!5!white, colframe=gray!70!black, title=Single Agent Variant Prompt,fontupper=\small]
<|start\_header\_id|>system<|end\_header\_id|>
 
You are an expert psychological counseling assistant. Your task is to generate the counselor response in the dialogue using CBT-grounded techniques. The goal is to create a natural and engaging response that builds on the previous conversation. Please ensure that the response is empathetic and understanding of the client's issues and builds trust between the counselor and the client. Please be mindful to only generate the counselor response for a single turn, and do not include extra text like "here is the next counselor utterance" or "Here is a possible next utterance" or anything mentioning the used technique.
 
Follow these steps internally before producing the final response:
 
Step 1 — CBT Planning:

Use the provided client information and dialogue to generate an appropriate CBT technique and a detailed counseling plan. Types of CBT Techniques: Efficiency Evaluation, Pie Chart Technique, Alternative Perspective, Decatastrophizing, Pros and Cons Analysis, Evidence-Based Questioning, Reality Testing, Continuum Technique, Changing Rules to Wishes, Behavior Experiment, Problem-Solving Skills Training, Systematic Exposure. Choose an appropriate CBT technique and create a counseling plan based on that technique.
 
Step 2 — Specialized Candidate Responses:

Internally draft five candidate responses, one per technique. Do not include these in the final output.

- Reflection: Reflection is a technique used by the counselor to help a client gain insight into their thoughts, feelings, and behaviors by mirroring or paraphrasing what the client expresses, allowing the client to hear and evaluate their own statements more clearly.

- Questioning: Questioning is a technique used by counselors to gain deeper understanding and insights on how the client feels regarding some previously mentioned events, how the client feels at present or understand how the client feels when asked to consider the situation from an alternative perspective.

- Solutions: Counselors provide solutions grounded in psychological techniques that can be used by clients to improve their condition and deal with the mental health issues they are facing.

- Normalization: Normalization is a technique used by the counselor to acknowledge and validate the client's experience as normal or expectable, sympathize with their challenges, and provide reassurance to foster a supportive and encouraging therapeutic atmosphere.

- Psycho-education: Psycho-education is a technique used by the counselor to provide therapeutically relevant information about psychological principles to the client to help them understand their issues and the logic behind the solutions.

Step 3 — Technique Selection:

Use the conversation history and counseling plan to select the appropriate technique or combination of techniques for this turn. Remember that the counselor should properly explore and understand the client issues, normalize the issues, and then provide solutions with psycho-education. Types of Psychological Techniques: Reflection, Questioning, Providing solutions, Normalization, Psycho-education.
 
Step 4 — Final Response:

Generate the counselor response in the dialogue by combining responses drafted in specialized responses on reflections, questioning, providing solutions, normalizing and psycho-education based on the selected techniques. Do not label or announce which techniques are being used. Do not structure the response as separate sections. The goal is to create a natural and engaging response that builds on the previous conversation. Please ensure that the response is empathetic and understanding of the client's issues and builds trust between the counselor and the client. Please be mindful to only generate the counselor response for a single turn, and do not include extra text like "here is the next counselor utterance" or "Here is a possible next utterance" or anything mentioning the used technique. Keep the response concise and natural — avoid lengthy or overly elaborate replies.

Length and style rules:

- Keep the response to 2-4 sentences maximum.

- If asking questions, keep them focused on a single topic or idea. Do not scatter questions across multiple unrelated themes in the same turn.

- Do not summarize everything the client said or pile multiple ideas into one response.

- Match the natural rhythm of a real back-and-forth counseling conversation.<|eot\_id|><|start\_header\_id|>user<|end\_header\_id|>

Client Information:

\{client\_information\}

Reason for seekeing counseling:

\{reason\_counseling\}

Counseling Dialogue:

\{history\}<|eot\_id|><|start\_header\_id|>assistant<|end\_header\_id|>

\end{tcolorbox}
\caption{Prompt used for Single Agent variant of Counselor agent.}
\label{fig:single-agent-variant}
\end{figure*}

\paragraph{Setup and Evaluation.} In this section, we compare \model\ with single-agent variants that consolidate all agent-specific functionalities into a unified prompt. Specifically, the single-agent setup performs all subtasks--CBT strategy generation, reflection, questioning, solution provision, normalization, psycho-education, and final response generation--within a single prompt. The full prompt used for the single-agent counselor is provided in Figure~\ref{fig:single-agent-variant}.

For comparison, we evaluate two single-agent baselines: (i) Llama3-8B-Instruct, and (ii) Llama3.3-70B-Instruct~\cite{meta2024llama33}, both acting as the counselor agents. The client agent uses Llama3-8B-Instruct across all configurations to ensure consistency.

For evaluation, we adopt GPT-4o as an LLM-as-a-judge to assess counseling quality along three axes: CTRS, WAI, and PANAS. The evaluation prompts remain identical to those described in Appendix~\ref{sec:appendix-quality}.

\paragraph{Results and Discussion.} The results are presented in Table~\ref{tab:single-agent-comp}. \model\ substantially outperforms the Llama3-8B-Instruct single-agent baseline on both CTRS and WAI, indicating that decomposing the counseling process into specialized agents improves both general counseling competence and CBT-specific skills, as well as the quality of the therapeutic alliance. On PANAS, \model\ achieves better performance in positive emotions of positive and neutral attitude but struggles in negative emotions and attitudes, which we attribute to its tendency to more actively engage with and explore negative emotions, consistent with our observations discussed previously. In contrast, Llama3.3-70B-Instruct surpasses \model\ across all metrics, highlighting the strong performance gains achievable through increased model scale. This suggests that while multi-agent decomposition improves performance within a fixed model capacity, it does not fully compensate for the advantages of significantly larger models.

Importantly, \model\ relies predominantly on Llama3-8B-Instruct for its components, making the comparison with the Llama3-8B single-agent baseline a controlled and fair evaluation of the multi-agent design rather than model scale. The observed gains therefore directly reflect the benefits of the multi-agent setup. In comparison, Llama3.3-70B-Instruct operates at a substantially larger scale, and its performance advantage underscores that such scale differences are difficult to overcome purely through multi-agent decomposition. Nevertheless, \model\ achieves stronger performance than the Llama3-8B single-agent baseline while maintaining similar computational requirements, making it a more accessible alternative to large-scale models. Additionally, the modular multi-agent design offers improved controllability, enabling targeted analysis and refinement of individual components, in contrast to single-agent systems where such intermediate reasoning steps remain implicit.

\begin{table*}[th]
\centering
\resizebox{1.95\columnwidth}{!}{%
\begin{tabular}{@{}lcccccccccccccccccccc@{}}
\toprule
\multirow{3}{*}{Method} 
& \multicolumn{7}{c}{CTRS} & & \multicolumn{8}{c}{PANAS} & & \multicolumn{3}{c}{WAI} \\ 
\cmidrule(lr){2-8} \cmidrule(l){10-17} \cmidrule(l){19-21}
& \multicolumn{3}{c}{General} & & \multicolumn{3}{c}{CBT} & & \multicolumn{2}{c}{Pos. Att.} & & \multicolumn{2}{c}{Neu. Att.} & & \multicolumn{2}{c}{Neg. Att.} & & \multirow{2}{*}{Task ($\uparrow$)} & \multirow{2}{*}{Goal ($\uparrow$)} & \multirow{2}{*}{Bond ($\uparrow$)} \\ 
\cmidrule(lr){2-4} \cmidrule(lr){6-8} \cmidrule(lr){10-11} \cmidrule(lr){13-14} \cmidrule(lr){16-17}
& U ($\uparrow$) & I ($\uparrow$) & C ($\uparrow$) & & D ($\uparrow$) & F ($\uparrow$) & S ($\uparrow$) & & P ($\uparrow$) & N ($\downarrow$) & & P ($\uparrow$) & N ($\downarrow$) & & P ($\uparrow$) & N ($\downarrow$) & & \\ 
\midrule
Llama3.3-70B & $\textbf{4.25}^{*}$ & $\textbf{5.49}^{*}$ & $\textbf{4.67}^{*}$ & & $\textbf{4.59}^{*}$ & $\textbf{4.01}^{*}$ & $\textbf{4.03}^{*}$ & & $\textbf{1.39}^{*}$ & $\textbf{-1.24}^{*}$ & & $\textbf{1.16}^{*}$ & $\textbf{-0.89}^{*}$ & & $\textbf{0.14}^{*}$ & $\textbf{-0.51}^{*}$ & & $\textbf{5.45}^{*}$ & $\textbf{5.41}^{*}$ & $\textbf{5.41}^{*}$ \\
Llama3-8B & $3.51^{*}$ & $3.81^{*}$ & $2.87^{*}$ & & $3.54^{*}$ & $3.16^{*}$ & $2.74^{*}$ & & $1.11$ & $-1.01^{*}$ & & $0.75$ & $-0.65^{*}$ & & $-0.10$ & $-0.43^{*}$ & & $4.72^{*}$ & $4.33^{*}$ & $4.76^{*}$ \\
\model & $3.98$ & $4.30$ & $3.43$ & & $4.08$ & $3.76$ & $2.93$ & & $1.13$ & $-0.73$ & & $0.82$ & $-0.41$ & & $-0.15$ & $-0.25$ & & $4.94$ & $4.78$ & $5.01$ \\  
\bottomrule
\end{tabular}%
}
\caption{Evaluation of counseling sessions generated by single agent variants compared to \model\ across CTRS, PANAS and WAI dimensions. Asterisks (*) indicate significant differences from \model\ ($p < 0.05$, paired t-test). For CTRS: U (Understanding), I (Interpersonal Effectiveness), C (Collaboration), D (Guided Discovery), F (Focus), S (Strategy). For PANAS: Pos. Att. (Positive Attitude), Neu. Att. (Neutral Attitude), Neg. Att. (Negative Attitude), P (average shift in positive emotions), N (average shift in negative emotions).}
\label{tab:single-agent-comp}
\end{table*}

\section{Technique Agent details}
\label{sec:appendix-tech-agent}

To ensure a fair comparison with the baselines, we also replace the technique agent with Llama3-8B-Instruct. Table~\ref{tab:tech-agent-ablation} presents results for \model\ with Llama3-8B-Instruct as the technique agent, alongside the GPT-4o-mini variant, the Llama3-8B-Instruct single-agent model (Appendix~\ref{sec:appendix-single-agent}), and the baselines. As expected, substituting the technique agent with Llama3-8B leads to a modest drop in performance. Nevertheless, this variant of \model\ consistently outperforms the Llama3-8B single-agent model across all dimensions, isolating the benefit of the multi-agent architecture. It also achieves the highest mean CTRS among the baselines, outperforming them on most individual dimensions.

We also report the selection frequency of each technique by the Llama3-8B technique agent to analyze the distribution of selected techniques. The distribution is presented in Table~\ref{tab:technique-distribution}. We observe that four of the five techniques are used substantially, with no single technique dominating. Their relative ordering aligns with the therapeutic progression encoded in our technique selection prompt (Figure~\ref{fig:technique}). The exception is \textit{providing solutions}, which appears less frequently because the prompt couples solution-oriented responses with psychoeducation. As a result, many such instances are labeled as psychoeducation rather than as a distinct technique, reflecting an absorption effect induced by the prompt design.

\begin{table*}[t]
\centering
\resizebox{\linewidth}{!}{%
\begin{tabular}{lccccccc}
\toprule
\textbf{Method} & \textbf{Understanding} & \textbf{Interpersonal Eff.} & \textbf{Collaboration} & \textbf{Guided Discovery} & \textbf{Focus} & \textbf{Strategy} & \textbf{Mean} \\
 & ($\uparrow$) & ($\uparrow$) & ($\uparrow$) & ($\uparrow$) & ($\uparrow$) & ($\uparrow$) & ($\uparrow$) \\
\midrule
Psych8k                  & \underline{3.90} & 4.10 & \underline{3.13} & 3.80 & 3.35 & 2.59 & 3.48 \\
CACTUS                   & 3.84 & 3.94 & 3.09 & 3.74 & 3.37 & 2.83 & 3.47 \\
Llama3-8B (single)       & 3.51 & 3.81 & 2.87 & 3.54 & 3.16 & 2.74 & 3.27 \\
\model{} (GPT-4o-mini tech) & \textbf{3.98} & \textbf{4.30} & \textbf{3.43} & \textbf{4.08} & \textbf{3.76} & \textbf{2.93} & \textbf{3.75} \\
\model{} (Llama3-8B tech)   & 3.81 & \underline{4.24} & 3.12 & \underline{3.88} & \underline{3.51} & \underline{2.87} & \underline{3.57} \\
\bottomrule
\end{tabular}%
}
\caption{Counseling quality evaluated on the six CTRS dimensions.  \textbf{Bold} marks the best result per column and \underline{underline} the second best.}
\label{tab:tech-agent-ablation}
\end{table*}

\begin{table}[t]
\centering
\begin{tabular}{lrr}
\toprule
\textbf{Technique} & \textbf{Count} & \textbf{Share} \\
\midrule
Reflection           & 8,512 & 38.6\% \\
Questioning          & 6,896 & 31.3\% \\
Normalization        & 4,412 & 20.0\% \\
Psycho-education     & 2,199 & 10.0\% \\
Providing Solutions  &    15 &  0.1\% \\
\bottomrule
\end{tabular}
\caption{Distribution of techniques selected by the Llama3-8B-Instruct technique agent.}
\label{tab:technique-distribution}
\end{table}

\section{Output Token Usage and Latency Analysis}
\label{sec:appendix-tokens}

Although the modular architecture of \model\ and the independence of the specialized response agents and the technique agent allow execution in three stages: (i) the CBT agent, (ii) the specialized response agents, and the technique agent and (iii) the response generation agent, \model\ incurs higher latency and token costs than the baselines. To quantify this performance-cost trade-off, we conducted a detailed analysis of output token consumption and wall-clock latency during session generation using \model\ and the baseline methods.

We sample 50 generation seeds and generate one session per seed for each method. Along with the baseline models, we also compare with the Llama3.3-70B-Instruct single agent method introduced in Appendix~\ref{sec:appendix-single-agent}. For each method, we compute the average number of output tokens and the average wall-clock time (in seconds). The results are presented in Table~\ref{tab:token-analysis}. As shown in the table, \model\ requires approximately $3.6\times$ output tokens and time compared to the baselines. Moreover, since the technique agent in \model\ uses GPT-4o-mini to generate content for each session, it incurs a small monetary cost of approximately $\$0.03$ per session. This additional cost is an efficiency trade-off for the performance gains through the structured multi-agent decomposition used in \model. We gain $3.2\%$ on average in general counseling skills, $4.3\%$ on average in CBT-specific skills using \model\ over baselines. Llama3-8B-Instruct models fine-tuned using the synthetically generated sessions further show gains of $6.3\%$ in general counseling skills and $7.4\%$ in CBT-specific skills over baselines.

Importantly, this overhead is incurred only during the synthetic data generation phase. After generating sufficient data, we fine-tune a single open-source model (Llama-\model) for downstream inference, thereby eliminating the multi-agent overhead at inference time.

For Llama3.3-70B, however, the results in Table~\ref{tab:single-agent-comp} show improved performance while maintaining output token usage and generation time comparable to the baselines (Table~\ref{tab:token-analysis}), making it approximately $3.6\times$ cheaper than \model. This efficiency is expected, as the single-agent setup requires only one LLM generation per turn. However, achieving this throughput requires substantially more hardware resources: Llama3.3-70B is deployed on 4$\times$48GB GPUs with an Int8 KV cache, whereas \model\ operates on a single 32GB GPU (Appendix~\ref{sec:appendix-exp-details}). Therefore, the trade-off in this case is primarily between throughput and hardware requirements rather than between cost and generation quality.

\begin{table}[th]
\centering
\resizebox{\columnwidth}{!}{%
\begin{tabular}{@{}lcc@{}}
\toprule
Method  & Output Tokens & Wall Clock Time (sec) \\ \midrule
Psych8k &     $4067.08$ &     $35.44$           \\
CACTUS  &      $4256.76$ &    $35.86$         \\
Llama3.3-70B  &      $3649.96$ &    $36.44$         \\\midrule
\model  &       $15178.00$  &     $128.16$      \\ \bottomrule
\end{tabular}%
}
\caption{Average output tokens used and wall clock time in seconds for generating a session using different methods.}
\label{tab:token-analysis}
\end{table}

\section{Experimental Details} 
\label{sec:appendix-exp-details}
For each agent involved in the client and counselor simulation, we use a temperature of $T=0.7$. For evaluations using LLM-as-a-judge, we use temperature $T=0$ for determinism. Each generation and evaluation is run only once, similar to CACTUS \cite{cactus}. For formatting the prompts, we use the LangChain\footnote{\href{https://www.langchain.com/}{https://www.langchain.com/}} Library. For the generation and evaluation process, we use the vLLM \cite{vllm} library and run them on a single V100 32GB GPU. For fine-tuning counseling agents, we use QLora \cite{qlora} fine-tuning. We set the rank to $64$ and $\alpha$ to $16$. We fine-tune the Llama3-8B-Instruct model with a learning rate of $2e-4$ for $3$ epochs using the DeepSpeed\footnote{\href{https://github.com/microsoft/DeepSpeed}{https://github.com/microsoft/DeepSpeed}} library on $4$ V100 32GB GPUs. For inference with Llama3.3-70B, we use 4$\times$48GB GPUs with an Int8 KV cache. We set seed $42$ for reproducibility. The Hugging Face\footnote{\href{https://huggingface.co/}{https://huggingface.co/}}, vLLM, and DeepSpeed libraries used for implementation, fine-tuning, and evaluation are licensed under Apache License, Version 2.0 while LangChain is licensed under the MIT License. We have confirmed all of the artifacts used in this paper are available for non-commercial scientific use.

\begin{table*}[th]
\centering
\resizebox{1.95\columnwidth}{!}{%
\begin{tabular}{@{}lcccccccccccccccccccc@{}}
\toprule
\multirow{3}{*}{Method} 
& \multicolumn{7}{c}{CTRS} 
&  & \multicolumn{8}{c}{PANAS} 
&  & \multicolumn{3}{c}{WAI} 
\\ \cmidrule(lr){2-8} \cmidrule(lr){10-17} \cmidrule(l){19-21}
& \multicolumn{3}{c}{General} 
&  & \multicolumn{3}{c}{CBT} 
&  & \multicolumn{2}{c}{Pos. Att.} 
&  & \multicolumn{2}{c}{Neu. Att.} 
&  & \multicolumn{2}{c}{Neg. Att.} 
&  & \multirow{2}{*}{Task ($\uparrow$)} & \multirow{2}{*}{Goal ($\uparrow$)} & \multirow{2}{*}{Bond ($\uparrow$)}
\\ \cmidrule(lr){2-4} \cmidrule(lr){6-8} \cmidrule(lr){10-11} \cmidrule(lr){13-14} \cmidrule(lr){16-17}
& U ($\uparrow$) & I ($\uparrow$) & C ($\uparrow$)
&  & D ($\uparrow$) & F ($\uparrow$) & S ($\uparrow$)
&  & P ($\uparrow$) & N ($\downarrow$)
&  & P ($\uparrow$) & N ($\downarrow$)
&  & P ($\uparrow$) & N ($\downarrow$)
& 
\\ \midrule

\multirow{2}{*}{{\small Psych8k}} 
& \underline{{\small$3.90^{*}$}} & \underline{{\small$4.10^{*}$}} &  \underline{{\small$3.13^{*}$}}
&  & \underline{{\small$3.80^{*}$}} & {\small$3.35^{*}$} & {\small$2.59^{*}$} 
&  & \underline{{\small$1.12$}} & \underline{{\small$\textbf{-0.78}$}}
&  & \underline{{\small$0.80$}} & {\small$-0.43$}
&  & \underline{{\small$\textbf{-0.01}^{*}$}} & \underline{{\small$\textbf{-0.30}$}}
&  & \underline{{\small$4.86$}} & \underline{{\small$4.73$}} & \underline{{\small$4.93^{*}$}}
\\
& {\small$\pm0.46$} & {\small$\pm0.84$} & {\small$\pm1.03$} 
&  & {\small$\pm0.75$} & {\small$\pm0.96$} & {\small$\pm0.99$}
&  & {\small$\pm0.49$} & {\small$\pm0.65$}
&  & {\small$\pm0.54$} & {\small$\pm0.61$}
&  & {\small$\pm0.69$} & {\small$\pm0.60$}
&  & {\small$\pm1.23$} & {\small$\pm1.15$} & {\small$\pm0.71$}
\\

\multirow{2}{*}{{\small CACTUS}} 
& {\small$3.84^{*}$} & {\small$3.94^{*}$} & {\small$3.09^{*}$}
&  & {\small$3.74^{*}$} & \underline{{\small$3.37^{*}$}} & \underline{{\small$2.83$}}
&  & {\small$0.86^{*}$} & {\small$-0.64$}
&  & {\small$0.64^{*}$} & \underline{{\small$\textbf{-0.45}$}}
&  & {\small$-0.10$} & \underline{{\small$\textbf{-0.30}$}}
&  & {\small$4.69^{*}$} & {\small$4.39^{*}$} & {\small$4.80^{*}$}
\\
& {\small$\pm0.54$} & {\small$\pm0.79$} & {\small$\pm0.98$}
&  & {\small$\pm0.68$} & {\small$\pm0.95$} & {\small$\pm1.07$}
&  & {\small$\pm0.47$} & {\small$\pm0.65$}
&  & {\small$\pm0.58$} & {\small$\pm0.60$}
&  & {\small$\pm0.63$} & {\small$\pm0.55$}
&  & {\small$\pm1.26$} & {\small$\pm1.10$} & {\small$\pm0.72$}
\\

\multirow{2}{*}{{\small \model}} 
& {\small$\textbf{3.98}$} & {\small$\textbf{4.30}$} & {\small$\textbf{3.43}$}
&  & {\small$\textbf{4.08}$} & {\small$\textbf{3.76}$} & {\small$\textbf{2.93}$}
&  & {\small$\textbf{1.13}$} & {\small$-0.73$}
&  & {\small$\textbf{0.82}$} & {\small$-0.41$}
&  & {\small$-0.15$} & {\small$-0.25$}
&  & {\small$\textbf{4.94}$} & {\small$\textbf{4.78}$} & {\small$\textbf{5.01}$}
\\
& {\small$\pm0.21$} & {\small$\pm0.77$} & {\small$\pm0.92$}
&  & {\small$\pm0.62$} & {\small$\pm0.68$} & {\small$\pm1.06$}
&  & {\small$\pm0.49$} & {\small$\pm0.71$}
&  & {\small$\pm0.55$} & {\small$\pm0.55$}
&  & {\small$\pm0.57$} & {\small$\pm0.57$}
&  & {\small$\pm1.24$} & {\small$\pm1.06$} & {\small$\pm0.62$}
\\ \midrule

{\small$\delta(\%)$}
& {\small$\textbf{+1.3}$} & {\small$\textbf{+3.3}$} & {\small$\textbf{+5.0}$}
&  & {\small$\textbf{+4.7}$} & {\small$\textbf{+6.5}$} & {\small$\textbf{+1.7}$}
&  & {\small$\textbf{+0.3}$} & {\small$\textbf{-1.3}$}
&  & {\small$\textbf{+0.5}$} & {\small$\textbf{-1.0}$}
&  & {\small$\textbf{-3.5}$} & {\small$\textbf{-1.3}$}
&  & {\small$\textbf{+1.3}$} & {\small$\textbf{+0.8}$} & {\small$\textbf{+1.3}$}
\\ \bottomrule

\end{tabular}%
}
\caption{Mean score and standard deviation for CTRS, PANAS, and WAI across Psych8k, CACTUS, and \model\ generated sessions. Asterisks (*) denote significant differences from \model\ ($p < 0.05$). $\delta(\%)$ shows \model’s \%-age margin over the best baseline (\underline{underlined}).}
\label{tab:ctrs-panas-wai-sd}
\end{table*}

\begin{table*}[th]
\centering
\resizebox{1.95\columnwidth}{!}{%
\begin{tabular}{@{}lcccccccccccccccccccc@{}}
\toprule
\multirow{3}{*}{Model} 
  & \multicolumn{7}{c}{CTRS} 
  &  & \multicolumn{8}{c}{PANAS}
  &  & \multicolumn{3}{c}{WAI} 
  \\ \cmidrule(lr){2-8} \cmidrule(lr){10-17} \cmidrule(l){19-21}

  & \multicolumn{3}{c}{General} &  & \multicolumn{3}{c}{CBT} 
  &  & \multicolumn{2}{c}{Pos. Att.} &  & \multicolumn{2}{c}{Neu. Att.} &  & \multicolumn{2}{c}{Neg. Att.}
  &  & \multirow{2}{*}{Task ($\uparrow$)} & \multirow{2}{*}{Goal ($\uparrow$)} & \multirow{2}{*}{Bond ($\uparrow$)}
  \\ \cmidrule(lr){2-4} \cmidrule(lr){6-8} \cmidrule(lr){10-11} \cmidrule(lr){13-14} \cmidrule(lr){16-17}

  & U ($\uparrow$) & I ($\uparrow$) & C ($\uparrow$)
  &  & D ($\uparrow$) & F ($\uparrow$) & S ($\uparrow$)
  &  & P ($\uparrow$) & N ($\downarrow$)
  &  & P ($\uparrow$) & N ($\downarrow$)
  &  & P ($\uparrow$) & N ($\downarrow$)
  & 
  \\ \midrule

\multirow{2}{*}{{\small Llama-Psych8k}}
  & \underline{{\small$3.71^{*}$}} & \underline{{\small$3.83^{*}$}} & \underline{{\small$2.91^{*}$}}
  &  & \underline{{\small$3.65^{*}$}} & \underline{{\small$3.16^{*}$}} & {\small$2.44^{*}$}
  &  & \underline{{\small$1.04$}} & \underline{{\small$-0.88^{*}$}}
  &  & \underline{{\small$0.70$}} & \underline{{\small$-0.41^{*}$}}
  &  & \underline{{\small$\textbf{-0.03}^{*}$}} & \underline{{\small$\textbf{-0.34}$}}
  &  & \underline{{\small$4.72$}} & \underline{{\small$4.44^{*}$}} & \underline{{\small$4.79$}}
  \\
& {\small$\pm 0.71$} & {\small$\pm 0.86$} & {\small$\pm 1.00$}
  &  & {\small$\pm 0.79$} & {\small$\pm 0.99$} & {\small$\pm 0.86$}
  &  & {\small$\pm 0.42$} & {\small$\pm 0.78$}
  &  & {\small$\pm 0.51$} & {\small$\pm 0.56$}
  &  & {\small$\pm 0.62$} & {\small$\pm 0.57$}
  &  & {\small$\pm 1.26$} & {\small$\pm 1.10$} & {\small$\pm 0.66$}
  \\

\multirow{2}{*}{{\small Llama-CACTUS}}
  & {\small$3.48^{*}$} & {\small$3.67^{*}$} & {\small$2.65^{*}$}
  &  & {\small$3.37^{*}$} & {\small$2.99^{*}$} & \underline{{\small$2.45^{*}$}}
  &  & {\small$0.91^{*}$} & {\small$-0.79^{*}$}
  &  & {\small$0.59^{*}$} & {\small$-0.39^{*}$}
  &  & {\small$-0.13^{*}$} & {\small$-0.32$}
  &  & {\small$4.59^{*}$} & {\small$4.18^{*}$} & {\small$4.65^{*}$}
  \\
& {\small$\pm 0.88$} & {\small$\pm 0.94$} & {\small$\pm 0.97$}
  &  & {\small$\pm 0.96$} & {\small$\pm 1.03$} & {\small$\pm 0.93$}
  &  & {\small$\pm 0.51$} & {\small$\pm 0.62$}
  &  & {\small$\pm 0.51$} & {\small$\pm 0.57$}
  &  & {\small$\pm 0.60$} & {\small$\pm 0.51$}
  &  & {\small$\pm 1.13$} & {\small$\pm 1.04$} & {\small$\pm 0.67$}
  \\

\multirow{2}{*}{{\small Llama-\model}}
  & {\small$\textbf{3.95}$} & {\small$\textbf{4.32}$} & {\small$\textbf{3.32}$}
  &  & {\small$\textbf{4.03}$} & {\small$\textbf{3.60}$} & {\small$\textbf{2.96}$}
  &  & {\small$\textbf{1.11}$} & {\small$\textbf{-1.10}$}
  &  & {\small$\textbf{0.78}$} & {\small$\textbf{-0.59}$}
  &  & {\small$-0.31$} & {\small$-0.28$}
  &  & {\small$\textbf{4.86}$} & {\small$\textbf{4.78}$} & {\small$\textbf{4.92}$}
  \\
& {\small$\pm 0.32$} & {\small$\pm 0.90$} & {\small$\pm 1.01$}
  &  & {\small$\pm 0.77$} & {\small$\pm 0.83$} & {\small$\pm 1.08$}
  &  & {\small$\pm 0.39$} & {\small$\pm 0.74$}
  &  & {\small$\pm 0.50$} & {\small$\pm 0.54$}
  &  & {\small$\pm 0.59$} & {\small$\pm 0.52$}
  &  & {\small$\pm 1.42$} & {\small$\pm 1.20$} & {\small$\pm 0.72$}
  \\ \midrule

{\small$\delta(\%)$}
  & {\small$\textbf{+4.0}$} & {\small$\textbf{+8.2}$} & {\small$\textbf{+6.8}$}
  &  & {\small$\textbf{+6.3}$} & {\small$\textbf{+7.3}$} & {\small$\textbf{+8.5}$}
  &  & {\small$\textbf{+1.8}$} & {\small$\textbf{+5.5}$}
  &  & {\small$\textbf{+2.0}$} & {\small$\textbf{+4.5}$}
  &  & {\small$\textbf{-7.0}$} & {\small$\textbf{-1.5}$}
  &  & {\small$\textbf{+2.3}$} & {\small$\textbf{+5.7}$} & {\small$\textbf{+2.2}$}
  \\ \bottomrule

\end{tabular}%
}
\caption{Mean score and standard deviation for CTRS, PANAS, and WAI scores across sessions generated using models fine-tuned on data from Psych8k (Llama-Psych8k), CACTUS (Llama-CACTUS), and \model\ (Llama-\model). Asterisks (*) denote significant differences from Llama-\model\ ($p < 0.05$). $\delta(\%)$ shows Llama-\model’s \%-age margin over the best baseline (\underline{underlined}).}
\label{tab:ctrs-panas-wai-model-sd}
\end{table*}

\begin{table*}[th]
\centering
\resizebox{2.0\columnwidth}{!}{%
\begin{tabular}{@{}lcccccccccccccccccccc@{}}
\toprule
\multirow{3}{*}{Method} 
  & \multicolumn{7}{c}{CTRS}
  &  & \multicolumn{8}{c}{PANAS}
  &  & \multicolumn{3}{c}{WAI} \\
\cmidrule(lr){2-8} \cmidrule(lr){10-17} \cmidrule(l){19-21}

  & \multicolumn{3}{c}{General} &  & \multicolumn{3}{c}{CBT}
  &  & \multicolumn{2}{c}{Pos. Att.} &  & \multicolumn{2}{c}{Neu. Att.} 
  &  & \multicolumn{2}{c}{Neg. Att.}
  &  & \multirow{2}{*}{Task ($\uparrow$)} & \multirow{2}{*}{Goal ($\uparrow$)} & \multirow{2}{*}{Bond ($\uparrow$)} \\
\cmidrule(lr){2-4} \cmidrule(lr){6-8} \cmidrule(lr){10-11} 
\cmidrule(lr){13-14} \cmidrule(lr){16-17}

  & U ($\uparrow$) & I ($\uparrow$) & C ($\uparrow$)
  &  & D ($\uparrow$) & F ($\uparrow$) & S ($\uparrow$)
  &  & P ($\uparrow$) & N ($\downarrow$)
  &  & P ($\uparrow$) & N ($\downarrow$)
  &  & P ($\uparrow$) & N ($\downarrow$)
  & 
  \\ \midrule

\multirow{2}{*}{{\small \model\textsubscript{-C}}}
  & \underline{{\small$3.96$}} & \underline{{\small$4.28$}} & \underline{{\small$3.29^{*}$}}
  &  & \underline{{\small$3.89^{*}$}} & \underline{{\small$3.58^{*}$}} & \underline{{\small$\textbf{3.00}$}}
  &  & \underline{{\small$\textbf{1.14}$}} & \underline{{\small$\textbf{-0.84}$}}
  &  & \underline{{\small$0.78$}} & {\small$-0.45$}
  &  & {\small$-0.12$} & \underline{{\small$\textbf{-0.33}$}}
  &  & {\small$4.92$} & \underline{{\small$\textbf{4.82}$}} & \underline{{\small$\textbf{5.05}$}} \\
& {\small$\pm 0.30$} & {\small$\pm 0.85$} & {\small$\pm 0.96$}
  &  & {\small$\pm 0.50$} & {\small$\pm 0.82$} & {\small$\pm 1.02$}
  &  & {\small$\pm 0.46$} & {\small$\pm 0.62$}
  &  & {\small$\pm 0.55$} & {\small$\pm 0.60$}
  &  & {\small$\pm 0.64$} & {\small$\pm 0.54$}
  &  & {\small$\pm 1.30$} & {\small$\pm 1.12$} & {\small$\pm 0.73$} \\

\multirow{2}{*}{{\small \model\textsubscript{-T}}}
  & {\small$3.88^{*}$} & {\small$3.86^{*}$} & {\small$3.09^{*}$}
  &  & {\small$3.85^{*}$} & {\small$3.45^{*}$} & {\small$2.70^{*}$}
  &  & {\small$1.08$} & {\small$-0.73$}
  &  & \underline{{\small$0.78$}} & {\small$-0.44$}
  &  & \underline{{\small$\textbf{-0.11}$}} & {\small$-0.28$}
  &  & \underline{{\small$\textbf{4.95}$}} & {\small$4.71$} & {\small$4.90^{*}$} \\
& {\small$\pm 0.48$} & {\small$\pm 0.65$} & {\small$\pm 0.97$}
  &  & {\small$\pm 0.58$} & {\small$\pm 0.90$} & {\small$\pm 0.99$}
  &  & {\small$\pm 0.41$} & {\small$\pm 0.69$}
  &  & {\small$\pm 0.60$} & {\small$\pm 0.60$}
  &  & {\small$\pm 0.63$} & {\small$\pm 0.52$}
  &  & {\small$\pm 1.15$} & {\small$\pm 1.06$} & {\small$\pm 0.65$} \\

\multirow{2}{*}{{\small \model\textsubscript{-C-T}}}
  & {\small$3.86^{*}$} & {\small$3.83^{*}$} & {\small$2.89^{*}$}
  &  & {\small$3.79^{*}$} & {\small$3.42^{*}$} & {\small$2.64^{*}$}
  &  & {\small$0.98^{*}$} & {\small$-0.79$}
  &  & {\small$0.72^{*}$} & \underline{{\small$\textbf{-0.47}$}}
  &  & {\small$-0.19$} & {\small$-0.30$}
  &  & {\small$4.94$} & {\small$4.72$} & {\small$4.88^{*}$} \\
& {\small$\pm 0.52$} & {\small$\pm 0.71$} & {\small$\pm 0.99$}
  &  & {\small$\pm 0.64$} & {\small$\pm 0.91$} & {\small$\pm 0.96$}
  &  & {\small$\pm 0.53$} & {\small$\pm 0.66$}
  &  & {\small$\pm 0.56$} & {\small$\pm 0.57$}
  &  & {\small$\pm 0.62$} & {\small$\pm 0.55$}
  &  & {\small$\pm 1.15$} & {\small$\pm 1.06$} & {\small$\pm 0.65$} \\

\multirow{2}{*}{{\small \model}}
  & {\small$\textbf{3.98}$} & {\small$\textbf{4.30}$} & {\small$\textbf{3.43}$}
  &  & {\small$\textbf{4.08}$} & {\small$\textbf{3.76}$} & {\small$2.93$}
  &  & {\small$1.13$} & {\small$-0.73$}
  &  & {\small$\textbf{0.82}$} & {\small$-0.41$}
  &  & {\small$-0.15$} & {\small$-0.25$}
  &  & {\small$4.94$} & {\small$4.78$} & {\small$5.01$} \\
& {\small$\pm 0.21$} & {\small$\pm 0.77$} & {\small$\pm 0.92$}
  &  & {\small$\pm 0.62$} & {\small$\pm 0.68$} & {\small$\pm 1.06$}
  &  & {\small$\pm 0.49$} & {\small$\pm 0.71$}
  &  & {\small$\pm 0.55$} & {\small$\pm 0.55$}
  &  & {\small$\pm 0.57$} & {\small$\pm 0.57$}
  &  & {\small$\pm 1.24$} & {\small$\pm 1.06$} & {\small$\pm 0.62$} \\

\midrule
{\small$\delta(\%)$}
  & {\small$\textbf{+0.3}$} & {\small$\textbf{+0.3}$} & {\small$\textbf{+2.3}$}
  &  & {\small$\textbf{+3.2}$} & {\small$\textbf{+3.0}$} & {\small$\textbf{-1.2}$}
  &  & {\small$\textbf{-0.3}$} & {\small$\textbf{-2.8}$}
  &  & {\small$\textbf{+1.0}$} & {\small$\textbf{-1.5}$}
  &  & {\small$\textbf{-1.0}$} & {\small$\textbf{-2.0}$}
  &  & {\small$\textbf{-0.2}$} & {\small$\textbf{-0.7}$} & {\small$\textbf{-0.7}$} \\
\bottomrule

\end{tabular}%
}
\caption{Mean score and standard deviation for CTRS, PANAS, and WAI scores across sessions generated by \model\ ablations: \model\textsubscript{-C} (no CBT agent), \model\textsubscript{-T} (no technique agent), \model\textsubscript{-C-T} (no CBT and technique agent), and \model. $\delta(\%)$ shows \%-gain of \model\ over the strongest ablation (\underline{underlined}).}
\label{tab:ctrs-panas-wai-ablation-sd}
\end{table*}

\end{document}